\documentclass[preprint,review,10pt]{elsarticle}
\usepackage[left=25mm,right=25mm,top=25mm,bottom=25mm,paper=a4paper]{geometry}

\usepackage{amssymb}
\usepackage{amsmath}

\usepackage{float}
\usepackage{graphicx}
\usepackage{array}
\usepackage{booktabs}
\usepackage{subcaption}
\usepackage{multirow}
\usepackage{tabularray}
\usepackage{arydshln} 
\usepackage{amsfonts}

\usepackage{adjustbox}

\usepackage{color, colortbl}
\usepackage{xcolor}
\usepackage{caption}

\usepackage[ruled, linesnumbered, noend]{algorithm2e}
\SetAlgoNlRelativeSize{-1}
\SetAlCapSkip{0.5em}

\usepackage{algpseudocode}
\usepackage{tikz} 
\usepackage{pifont}
\newcommand{\cmark}{\ding{51}}
\newcommand{\xmark}{\ding{55}}

\journal{Pattern Recognition}

\begin{document}

\begin{frontmatter}

\title{PSMamba: Progressive Self-supervised Vision Mamba for \\ Plant Disease Recognition}

\author[label1,label3]{Abdullah Al Mamun}
\author[label2,label3]{Miaohua Zhang}
\author[label3]{David Ahmedt-Aristizabal}
\author[label3]{\\Zeeshan Hayder}
\author[label1]{Mohammad Awrangjeb}

\affiliation[label1]{organization={School of Information and Communication Technology},
            addressline={Griffith University},
            city={Nathan},
            postcode={4111},
            state={Queensland},
            country={ Australia.}}
\affiliation[label2]{organization={School of Engineering and Built Environment},
            addressline={Griffith University},
            city={Nathan},
            postcode={4111},
            state={Queensland},
            country={ Australia.}}
\affiliation[label3]{organization={Imaging and Computer Vision Group},
            addressline={Data61, CSIRO},
            city={Black Mountain},
            postcode={2601},
            state={Canberra},
            country={Australia}}

\begin{abstract}

Self-supervised Learning (SSL) has become a powerful paradigm for representation learning without manual annotations. However, most existing frameworks focus on global alignment and struggle to capture the hierarchical, multi-scale lesion patterns characteristic of plant disease imagery. To address this gap, we propose \textbf{PSMamba}, a progressive self-supervised framework that integrates the efficient sequence modelling of Vision Mamba (VM) with a dual-student hierarchical distillation strategy. Unlike conventional single teacher-student designs, PSMamba employs a shared global teacher and two specialised students: one processes mid-scale views to capture lesion distributions and vein structures, while the other focuses on local views to capture fine-grained cues such as texture irregularities and early-stage lesions. 
This multi-granular supervision facilitates the joint learning of contextual and detailed representations, with consistency losses ensuring coherent cross-scale alignment. 
Experiments on three benchmark datasets show that PSMamba consistently outperforms state-of-the-art SSL methods, delivering superior accuracy and robustness in both domain-shifted and fine-grained scenarios.  
\end{abstract}




\end{frontmatter}


\section{Introduction}
\label{sec:intro}

Plant diseases cause up to 40\% yield losses and over USD 220 billion in annual damage \cite{al2024plant}. Automated detection is crucial, but traditional expert diagnosis is slow and subjective \cite{barbedo2019plant}. 
While supervised deep learning models \cite{kumar2023soybean,abid2024bangladeshi} offer scalable alternatives, they depend on costly expert annotations and cannot feasibly cover the diversity of crop diseases. The diversity of crop diseases and variability of symptoms further make exhaustive labelling impractical, underscoring the need for self-supervised approaches that learn robust, transferable features without heavy labelling requirements \cite{Zhao2023CLA}.

\begin{figure}[t]
\centering
\includegraphics[width=0.6\linewidth]{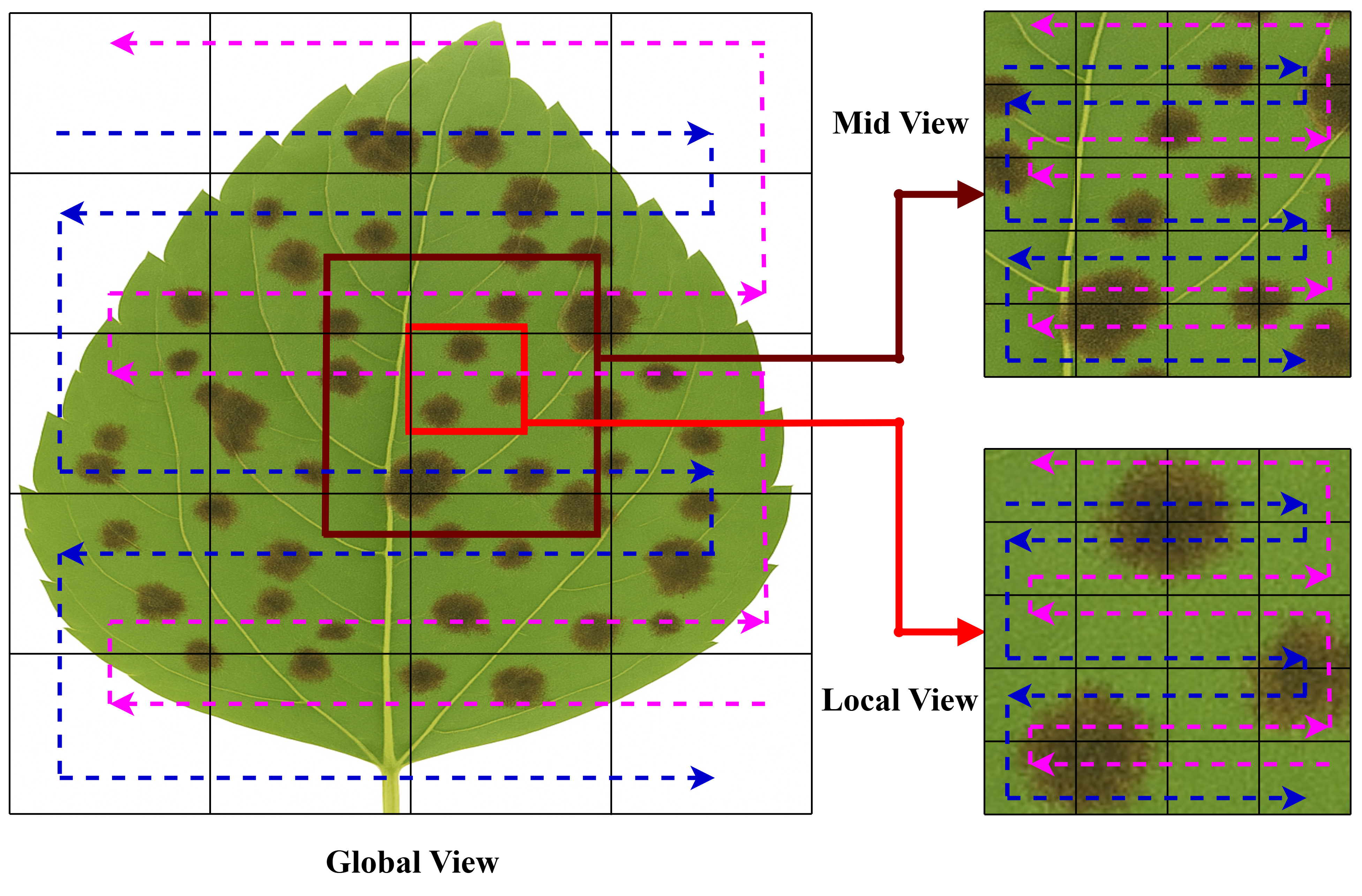}
\caption{Multi-granular views and notation. The Global view explains where lesions are likely to appear; the Mid view captures how lesions interact with nearby veins; the Local view represents micro textures.}
\label{fig:concept}
\end{figure}

Recent advances in SSL have produced a wide range of pretext tasks for representation learning without annotations. Early context-based tasks, such as rotation prediction \cite{gidaris2018unsupervised}, and image colourisation \cite{zhang2016colorful}, demonstrated the feasibility of exploiting spatial cues but offered limited adaptability. Contrastive learning later became dominant, with methods such as SimCLR \cite{chen2020simple}, MoCo \cite{he2020momentum}, and ConMamba \cite{mamun2025conmamba} achieving strong transfer performance. Non-contrastive frameworks, including BYOL \cite{grill2020bootstrap} and SimSiam \cite{chen2021exploring}, revealed that negative pairs are not essential for learning discriminative features. Clustering-based approaches, such as SwAV \cite{caron2020unsupervised}, capture semantic groupings, while redundancy reduction designs, like VICReg \cite{bardes2021vicreg}, encourage decorrelated representations. Reconstruction-based approaches, especially masked image modelling (BEiT \cite{bao2021beit}, MAE \cite{he2022masked}), gained prominence by predicting missing patches, while the hybrid framework iBOT \cite{zhou2021ibot} integrates contrastive, clustering, and generative principles.

Despite these advances, most SSL methods emphasise global feature alignment and overlook the hierarchical structure of plant disease symptoms. As shown in Figure \ref{fig:concept}, disease cues appear at multiple scales: subtle necrotic spots at the local level, vein-related lesion propagation at the mid level, and broader discolouration patterns at the global level \cite{li2024hierarchical}. Capturing this hierarchy is essential for early detection, where fine-grained cues are easily overshadowed by dominant global patterns. 
Multiview SSL frameworks such as SwAV \cite{caron2020unsupervised}, iBOT \cite{zhou2021ibot}, MViT \cite{fan2021multiscale}, and DINO \cite{caron2021emerging} introduce scale diversity through varied crops, yet typically embed all crops into a single representation space, limiting semantic alignment. Shared positional embeddings suppress resolution-specific cues \cite{dosovitskiy2020image,caron2021emerging}, while per-scale embeddings \cite{chen2021crossvit} preserve scale-dependent information but lack mechanisms for explicit alignment, leaving cross-resolution coherence unresolved.

Backbone design further influences representation quality. Convolutional Neural Networks (CNNs) capture local texture but struggle with long-range dependencies. Vision Transformers \cite{dosovitskiy2020image} model global context effectively but incur quadratic complexity, making them less practical for high-resolution agricultural imagery. State Space Models (SSMs), particularly VM \cite{zhu2024vision}, offer an efficient alternative by replacing full attention with structured recurrence. This yields linear time processing while retaining the ability to capture both short and long-range interactions, making Mamba \cite{gu2023mamba} well-suited for modelling the spatial progression of plant disease symptoms. Although ConMamba \cite{mamun2025conmamba} integrates VM \cite{zhu2024vision} into a contrastive SSL framework and achieves competitive performance, its design inherits several limitations from the contrastive formulation. Because the method aligns only two global augmented views, the model is primarily driven by coarse structural cues and has limited access to the mid and local granularities where critical disease details arise. As illustrated in Figure \ref{fig:concept}, disease symptoms span diverse spatial scales, and without granular supervision, the model often overlooks early-stage signals and fails to capture consistent relationships between local anomalies and global disease patterns. 

To address these limitations, we propose Progressive Self-Supervised Vision Mamba (PSMamba) for plant disease detection, which integrates an SSL framework that integrates a bidirectional VM encoder with a dual student hierarchical distillation strategy. PSMamba processes each image at three granularities (Figure \ref{fig:concept}): a Global view capturing broad lesion distribution and spatial layout, a Mid view modelling vein-aligned lesion interactions, and a Local view focusing on fine-grained necrotic and texture level anomalies essential for early detection.

To maintain coherence across resolutions, PSMamba employs a dual student architecture in which two student networks specialise in Mid and Local granularities and are guided by a shared Global teacher through cross-scale consistency. To bridge differences in patch resolution, we introduce Multi-scale Positional Alignment (MPA), a geometry-aware adaptation module that aligns tokens across granularities and ensures comparability within the shared feature space. The bidirectional VM encoder further enhances efficiency and expressivity by capturing both localised lesion cues and long-range dependencies with linear time complexity. Through progressive distillation from the Global teacher, PSMamba explicitly aligns hierarchical features across scales and overcomes the semantic gap inherent in single-representation SSL.

In summary, our main contributions are as follows: 
\begin{itemize}
    \item[i)]  We introduce PSMamba, a dual-student hierarchical distillation State Space Model (SSM) for multi-granular feature learning. The Mid student captures lesion distribution and vein structure, while the Local student detects fine-grained details such as texture irregularities.
    \item[ii)]  We propose Multi-scale Positional Alignment (MPA), a geometry-aware, scale-specific adaptation module that stabilises cross-scale supervision by ensuring token-level comparability across resolutions.
    \item[iii)]  We integrate the bidirectional Vision Mamba encoder within the PSMamba framework, demonstrating how its efficient long-range modelling complements multi-granular feature extraction, enabling robust disease recognition under a lightweight, modular backbone.
\end{itemize}


\section{Related Works}
\label{sec:formatting}

\subsection{Distillation Strategies} 

SSL has become central for learning representations without large-scale annotations, with self-distillation driving many state-of-the-art methods in both vision and multimodal domains~\cite{chen2020simple}. 
A seminal work, SimCLR \cite{chen2020simple}, aligned embeddings of two augmented views using a contrastive loss. Building on this idea,  extensions such as 
MoCo \cite{he2020momentum} incorporated a momentum encoder for dynamic negative sampling, while BYOL \cite{grill2020bootstrap} showed that meaningful representations could be learned without negatives using an EMA-updated teacher. Beyond contrastive setups, SwAV \cite{caron2020unsupervised} employed clustering-based prototypes, and BINGO \cite{xu2021bag} enforced consistency both within and across grouped instances. A parallel line of research has explored self-supervised model compression and student–teacher knowledge transfer, focusing on distilling rich representations from large SSL models into compact, computationally efficient student networks. CompRess~\cite{abbasi2020compress} demonstrated that compressed students can retain the semantic structure of teacher embeddings by aligning nearest-neighbour relationships in feature space. ISD~\cite{tejankar2021isd} preserved instance similarity distributions during distillation, while SimReg~\cite{navaneet2022simreg} introduced regularised distribution matching to stabilise the student’s training trajectory. SEED~\cite{fang2021seed} showed that enforcing consistent prediction variance improves generalisation, and auxiliary-branch distillation approaches~\cite{dadashzadeh2022auxiliary} leveraged side networks to refine student representations. Collectively, these methods confirm that smaller models can inherit the discriminative capacity of larger SSL teachers, provided that feature alignment is kept stable across augmentations. The DINO \cite{caron2021emerging} advanced teacher–student learning toward local-to-global supervision by allowing students to process small and intermediate crops while the teacher processes a global view. This multi-crop strategy encouraged emergent part discovery and improved downstream transferability. However, even in these frameworks, all crops are projected into a unified embedding space without mechanisms that explicitly align features across different spatial scales. As a result, fine-grained cues present in local views may not correspond consistently to the global structural context, and cross-scale relationships remain weakly constrained. This gap motivates a design that not only processes multiple views but also aligns and supervises them coherently across resolutions.

\subsection{Positional Embedding Techniques}
Vision Transformers (ViTs) treat image patches as token sequences without inherent spatial structure, making positional embeddings necessary. Early approaches used fixed sinusoidal embeddings from the original Transformer~\cite{vaswani2017attention}, but their rigidity limited performance on scale-variant tasks. 
Learned absolute embeddings, introduced in ViT \cite{dosovitskiy2020image}, improved flexibility by assigning trainable vectors to patch positions, yet they generalise poorly to varying input sizes and require interpolation that can degrade performance \cite{touvron2021training,bao2021beit}. This formulation has also been widely adopted in various domain-specific ViT derivatives such as Mix-ViT \cite{yu2023mix} and IEM-ViT \cite{zhang2023information}, which add trainable position vectors directly to patch tokens. While effective within fixed input resolutions, these absolute embeddings remain sensitive to scale changes and offer no strategy to ensure consistency across multi-crop SSL settings.

A second line of research distinguishes between shared and scale-specific embeddings. Shared embeddings use a single learnable table across all input views, as in DINO \cite{caron2021emerging} and iBOT \cite{zhou2021ibot}. This reduces parameters and stabilises optimisation but suppresses scale-specific cues by forcing different views into a common reference frame, weakening semantic alignment between Global and Local views. Scale-specific embeddings instead allocate distinct tables for each resolution branch, preserving resolution-dependent features. CrossViT \cite{chen2021crossvit}, for example, applies separate embeddings for small- and large-patch tokens, while MViT \cite{fan2021multiscale} learns embeddings independently across hierarchical stages. More recently, MAE \cite{he2022masked} and SimMIM \cite{xie2022simmim} employed resolution-aware embeddings in encoder–decoder frameworks, demonstrating improved reconstruction and recognition stability under variable input sizes.

Despite these advances, aligning semantic information across multiview SSL frameworks remains an open challenge. Shared embeddings risk over-compressing positional differences, while scale-specific embeddings isolate resolution and prevent explicit cross-scale consistency. This motivates alignment strategies that preserve scale-specific cues while enforcing comparability across resolutions.

\subsection{Vision Mamba and State-Space Models}

SSMs have recently emerged as efficient sequence learners, offering a middle ground between CNNs and transformers. CNNs excel at local feature extraction but struggle with long-range dependencies, while transformers capture global context at quadratic cost in self-attention \cite{dosovitskiy2020image}. 
SSMs address this by reformulating sequence modelling through latent state transitions with linear complexity, enabling efficient modelling of long sequences \cite{gu2021efficiently}. 
The breakthrough S4 model \cite{gu2021efficiently} leveraged HiPPO compression \cite{gu2020hippo} and low-rank parameterisation for stability, while Mamba \cite{gu2023mamba} further improved efficiency with selective mechanism and hardware-aware scanning, achieving linear-time modelling without sacrificing long-range capacity.

Mamba has shown versatility across domains such as speech \cite{masuyama2024mamba}, audio \cite{erol2024audio}, motion \cite{zhang2024motion}, and time-series modelling \cite{zhou2024bit,li2024harmamba}, consistently outperforming CNNs and Transformers in efficiency and scalability.
In speech and audio, MADEON \cite{masuyama2024mamba} applied it to decoder-only ASR with bidirectional speech prefixing, while AudioMamba \cite{erol2024audio} introduced bidirectional designs for audio deepfake detection and classification. In motion and biomedical sensing, Motion Mamba \cite{zhang2024motion} addressed long-sequence human motion generation, and BiT-MamSleep \cite{zhou2024bit}, and HARMamba \cite{li2024harmamba} applied bidirectional Mamba to EEG sleep staging and wearable activity recognition. For time series, Bi-Mamba+ \cite{liang2024bi} and SSD-TS \cite{gao2025ssd} enhanced forecasting and imputation, while BEST-STD and XLSR-Mamba \cite{singh2025best} advanced spoken term detection and spoofing detection. Bidirectional VM is also employed as a backbone in plant disease detection, where it has been shown to outperform CNN and Transformer architectures by effectively modelling structured dependencies \cite{mamun2025conmamba}. However, its supervision is limited to two global augmented views, restricting the capture of hierarchical symptom patterns that emerge at intermediate and local scales. Therefore, it lacks strategies for cross-scale alignment and multi-granular supervision.

These limitations, together with VM’s linear complexity and bidirectional context, motivate the development of PSMamba, which extends VM from single-scale contrastive alignment to a multi-granular, alignment-aware, dual-student distillation framework capable of jointly learning fine-grained, vein-level, and global lesion cues within a unified representation space.

\section{Method}


\begin{figure*}[t!]
    \centering
    \includegraphics[width=\linewidth]{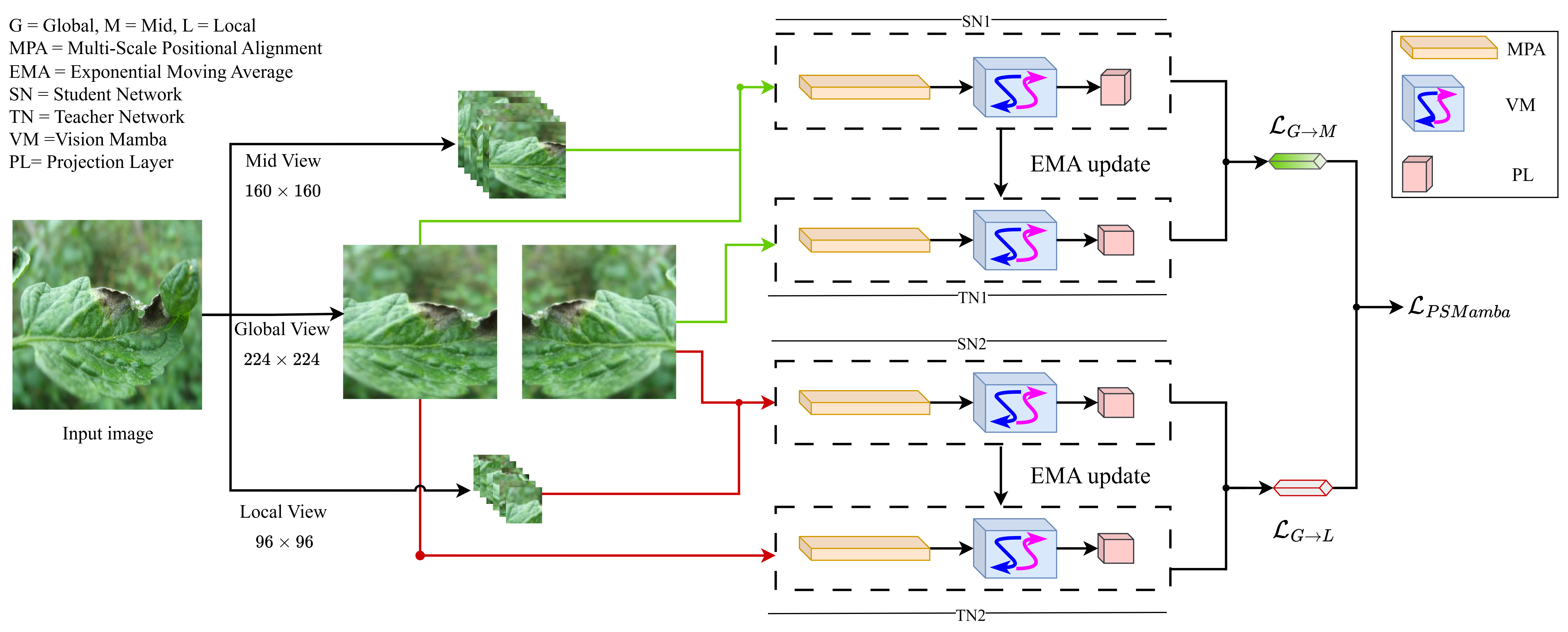}
    \caption{Overview of the proposed PSMamba framework. 
    An input image is augmented into three spatial views: Global, Mid, and Local. 
    Mid and Local tokens are aligned to the Global token grid through the Multi-scale Positional Alignment (MPA) module. 
    All views are encoded by the bidirectional Vision Mamba (VM), which captures both fine-grained lesion details and long-range dependencies. 
    On these aligned representations, the Exponential Moving Average (EMA) teacher supervises Mid and Local students within their visible regions, while a symmetric agreement loss enforces consistency over overlapping areas.}
    \label{fig:MMDP}
    \vspace{-4mm}
\end{figure*}

\subsection{Problem Formulation and Overview of PSMamba}
We consider self-supervised representation learning for plant disease imagery, where each training sample is a high-resolution leaf image $x$ containing lesion cues distributed across multiple spatial scales. Traditional multi-granular SSL approach \cite{caron2021emerging} typically employs two crop resolutions, $224\times224$ for the Global view and six $96\times96$ for the Local view, corresponding to spatial size of $2^5\times7$ and $2^5\times3$, respectively. This two-level hierarchy omits the intermediate scale at which vein topology and mid-size lesion structures become prominent. To address this gap, PSMamba introduces a Mid view by selecting the intermediate size $2^5\times5$ ($160\times160$), which lies naturally between the Global and Local resolutions and provides a principled middle granularity for capturing mid-scale disease patterns. Instead of relying on image-level labels, we learn a feature encoder $f_\theta$ by enforcing representation consistency across these three multi-granular augmented views of the same image.

For each input image $x$, we generate a set of augmented crops at three spatial scales: 
two \emph{Global} views $\phi_G(x)\!\in\!\mathbb{R}^{224\times224}$ capturing whole-leaf structure, 
six \emph{Mid} views $\phi_M(x)\!\in\!\mathbb{R}^{160\times160}$ focusing on vein-level and mid-scale lesion patterns, 
and six \emph{Local} views $\phi_L(x)\!\in\!\mathbb{R}^{96\times96}$ capturing fine-grained textures and lesion boundaries. 
These multi-scale and multi-view crops form the set 
$\{\phi_s^{(k)}(x)\}_{s\in\{G,M,L\},\,k}$, with overlapping but non-identical spatial support. For clarity of exposition, the formulation below uses a single representative view per scale.

Learning from these multi-scale crops presents two fundamental challenges in plant disease imagery:
(i) discriminative evidence is distributed across scales, from whole-leaf layout and vein-level relations to micro-texture patterns; and 
(ii) cross-scale supervision is ill-posed unless predictions from different views can be aligned to a common token index. 
Together, these challenges make hierarchical SSL non-trivial, as features extracted at different resolutions cannot be directly compared without an alignment mechanism.

Our objective is therefore to learn view-consistent token representations on a shared Global grid $\Lambda_G$ while preserving scale-specific information. 
Formally, for each view $s\in\{G,M,L\}$ we obtain a token sequence $Z_s = f_\theta(\phi_s(x)) \in \mathbb{R}^{N_s \times d}$ and seek a mapping
\begin{equation}
\widetilde Z_s = \mathcal{A}_s\!\big(Z_s, \Lambda_s \to \Lambda_G\big) \in \mathbb{R}^{N_G \times d},
\end{equation}
such that (i) $\widetilde Z_s$ lies on the Global token grid $\Lambda_G$ and (ii) tokens corresponding to the same physical region of the leaf are aligned across views. 
Self-distillation then encourages the MPA-aligned representations $\{\widetilde Z_M,\widetilde Z_L\}$ to match a Global teacher $Z_G^{\text{teacher}}$ on their visible regions.  

Therefore, PSMamba integrates three key components to address the above challenges and fulfill the objective of learning view-consistent representations on a shared Global grid:
(i) a bidirectional VM backbone (Section \ref{sec:mamba}) that provides linear-time sequence modelling while capturing both fine-grained lesion cues and long-range structural dependencies;
(ii) an MPA module (Section \ref{sec:mpa}) that maps Mid and Local tokens into the Global coordinate frame via geometry-aware resampling; and
(iii) an MPA-aware dual-student distillation strategy (Section \ref{sec:distill}) in which two specialised students (Mid and Local) are supervised by a Global EMA teacher on the aligned Global grid, with an optional symmetric student–student agreement term over overlapping regions. 
Figure \ref{fig:MMDP} summarises how these components interact within the PSMamba pipeline to resolve cross-scale inconsistencies, and the following sections describe each module in detail.

\subsection{Bidirectional Vision Mamba}
\label{sec:mamba}

For each view $s\!\in\!\{G,M,L\}$ (Global, Mid, Local), the input $\phi_s(x)$ is partitioned into $N_s$ patch tokens on a 2D grid 
$\Lambda_s=\{v_j\}_{j=1}^{N_s}$ and embedded as $u_j^{(0)}\!\in\!\mathbb{R}^d$.
A fixed, invertible rasterisation $\pi:\Lambda_s\!\to\!\{1,\ldots,N_s\}$ assigns a 1D scan index $t=\pi(v_j)$
for sequence modelling, while the inverse $\pi^{-1}$ preserves the spatial coordinates required by the MPA module.

Mamba is derived from continuous-time SSMs, whose latent dynamics are followed by Equation \ref{eq:ssm_cont} with the output of Equation \ref{eq:ssm_output}. 
\begin{equation}
\frac{d h(t)}{dt} = A h(t) + B u(t).
\label{eq:ssm_cont}
\end{equation}

\begin{equation}
y(t) = C h(t).
\label{eq:ssm_output}
\end{equation}
To adapt these models for deep learning, the continuous-time parameters are converted into a discrete-time recurrence using Zero-Order Hold (ZOH) discretization with timestep $\Delta$, yielding
\begin{equation}
\bar A = \exp(\Delta A), \qquad
\bar B = (\Delta A)^{-1}\big(\exp(\Delta A) - I\big)\,\Delta B.
\label{eq:zoh}
\end{equation}
This produces the discrete update of Equation \ref{eq:ssm_disc}, which forms the basis of the Mamba layer, and Equation \ref{eq:ssm} corresponds to this discretised formulation.

\begin{equation}
h_{t+1} = \bar A h_t + \bar B u_t, \qquad
z_t = C h_t.
\label{eq:ssm_disc}
\end{equation}

Each Mamba layer, therefore, implements an input–dependent discrete SSM scan.  
Denoting $u_t$ as the layer input at position $t$ and $z_t$ as the output, the update becomes
\begin{equation}
\begin{aligned}
x_{t+1} &= A_\ell(g_s,u_t)\,x_t + B_\ell(g_s,u_t)\,u_t,\\
z_t &= C_\ell(g_s,u_t)\,x_t,
\end{aligned}
\label{eq:ssm}
\end{equation}
where $x_t$ is the hidden state and $g_s$ is a scale-specific code that modulates the layer.
Residual connections and normalisation are applied following standard Mamba practice.

To capture non-causal spatial dependencies without quadratic attention, each layer performs both forward and backward scans over the sequence:
\begin{equation}
\begin{aligned}
\overrightarrow{Z}_t &= \mathrm{SSM}_\ell^{\rightarrow}(u_t),\\
\overleftarrow{Z}_t &= \mathrm{SSM}_\ell^{\leftarrow}(u_t),\\
Z_t &= W_f \overrightarrow{Z}_t + W_b \overleftarrow{Z}_t,
\end{aligned}
\label{eq:bi}
\end{equation}
where $W_f$ and $W_b$ are learnable fusion weights.
Stacking $L$ layers yields view features
\[
Z_s = \mathrm{Mamba}_\theta(\phi_s(x)) \in \mathbb{R}^{N_s \times d}.
\]

Plant disease symptoms exhibit both subtle local cues and long-range patterns that follow vein structures and spread irregularly across the lamina. Bidirectional SSM scanning allows PSMamba to aggregate information from both ‘past’ and ‘future’ in the raster order without quadratic attention, which is important for capturing elongated lesions and vein-aligned patterns over long spatial ranges.

\subsection{Multi-scale Positional Alignment (MPA)}
\label{sec:mpa}

Mid and Local views provide complementary evidence to the Global view, but their tokens reside on different coordinate grids. 
MPA is a lightweight alignment module that (i) maps view tokens into the Global coordinate frame and (ii) enforces token-level consistency so that teacher$\to$student and student$\leftrightarrow$student losses are well-posed.

Given a view $s \in \{G, M, L\}$, the augmentation operator $\phi_s(\cdot)$ induces a token grid
$\Lambda_s = \{v_j\}_{j=1}^{N_s}$, while the Global
view has $\Lambda_G = \{u_i\}_{i=1}^{N_G}$. 
A VM encoder $f_\theta$ produces token features
\begin{equation}
\begin{aligned}
&Z_s \;=\; f_{\theta}\!\big(\phi_s(x)\big) \in \mathbb{R}^{N_s \times d},\\
&Z_G^{\text{teacher}} \;=\; f_{\bar{\theta}}\!\big(\phi_G(x)\big),
\end{aligned}
\label{eq:mpa-feats}
\end{equation}

for student views ($s \in \{M,L\}$) and the teacher ($G$), respectively. Let $M_s \in \{0,1\}^{N_G}$ denote the visibility mask marking Global tokens covered by view $s$.

Each view token $v_j$ is mapped to Global coordinates by the known view-to-global geometry $T_s : \Lambda_s \to \mathbb{R}^2$. 
MPA adds a small learnable offset $\delta_s(v_j)$ to correct residual mis-registration:
\begin{equation}
\widehat T_s(v_j) = T_s(v_j) + \delta_s(v_j),
\label{eq:mpa-landing}
\end{equation}
where $T_s(v_j)$ is the geometric landing, and $\delta_s(v_j) \in \mathbb{R}^2$ provides a fine-scale, learnable warp.

To project view tokens onto $\Lambda_G$, we discretise $\widehat T_s(v_j)$ onto nearby Global tokens using a sparse, mass-preserving resampler $\Pi^{\mathrm{MPA}}_s$:
\begin{equation}
\begin{aligned}
&(\Pi^{\mathrm{MPA}}_s)_{ij} \;\propto\; k\!\big(u_i - \widehat T_s(v_j)\big),\\
&i \;\in\; \mathcal{N}\!\big(\widehat T_s(v_j)\big), \sum\nolimits_i (\Pi^{\mathrm{MPA}}_s)_{ij} \;=\; 1,
\end{aligned}
\label{eq:mpa-kernel}
\end{equation}
where $k(\cdot)$ is a nonnegative bilinear kernel and $\mathcal{N}(\cdot)$ denotes the $2{\times}2$ neighbourhood. Column normalisation preserves constant feature fields and average energy.

Student features are then transported to the Global grid as
\begin{equation}
\widetilde Z_s = \Pi^{\mathrm{MPA}}_s\, Z_s
\in \mathbb{R}^{N_G \times d},
\label{eq:mpa-transport}
\end{equation}
resampling tokens as the refined positions $\{\widehat T_s(v_j)\}_j$. Losses are computed only on visible positions with $M_s(i)=1$.

Scale changes can still alter appearance after geometric alignment. To address this, MPA introduces a view-specific positional adapter $P^{(s)}:\Lambda_s \to \mathbb{R}^d$, which adjusts features in a scale-aware manner before transport:
\begin{equation}
\widehat Z_s(j,\cdot) = Z_s(j,\cdot) + P^{(s)}(v_j),
\widetilde Z_s = \Pi^{\mathrm{MPA}}_s\, \widehat Z_s.
\label{eq:adapter-def}
\end{equation}
Because transport is linear, Equation \eqref{eq:adapter-def} implies
$\widetilde Z_s = \Pi^{\mathrm{MPA}}_s Z_s + \Pi^{\mathrm{MPA}}_s P^{(s)}$, 
so the adapter contributes a well-defined, token-level correction in the Global frame. 
No auxiliary loss is required; it is learned end-to-end through the distillation objective.

In the proposed PSMamba framework, we do not align Mid and Local views directly, as this introduces optimisation instability arising from their partial spatial overlap and inherently different receptive fields. Mid views also contain fewer pixels than the original image, causing loss of fine-grained information when forcing alignment with Local views. Moreover, such pairwise alignment substantially increases computational cost, since each Local crop would need to be processed alongside six Mid crops. Instead, we use the Global view as a shared anchor, which provides consistent geometry, stable gradients, and an effective means of coupling Mid and Local representations through the teacher and the symmetric student agreement mechanism.

\subsection{MPA-aware Dual-Student Distillation}
\label{sec:distill}

Given MPA-aligned features $\widetilde Z_s \in \mathbb{R}^{N_G \times d}$ for $s \in \{M,L\}$ from
Equation \eqref{eq:mpa-transport} (or Equation \eqref{eq:adapter-def} when the adapter is used),
we maintain a Global EMA teacher $f_{\bar\theta}$ and two student networks (Mid and Local).
Both students share encoder parameters $\theta$, with scale-specific conditioning as in the backbone.

\subsubsection{Distillation on Visible Positions.}
After MPA alignment (Section \ref{sec:mpa}), the teacher and each student predict distributions on the same Global grid $\Lambda_G$. Since Mid and Local views cover only subsets of $\Lambda_G$, we restrict supervision to visible regions using their masks $M_s \in \{0,1\}^{N_G}$. This avoids noisy gradients from unobserved tokens, with the loss normalised by $|M_s|_1$ to remain view-size invariant. Thus, the teacher acts as a calibrated target only where the student has direct evidence.

\begin{algorithm}[t]
\caption{Training pipeline of PSMamba with MPA and Dual-Student Distillation.}
\label{alg:psmamba}
\begin{algorithmic}[1]
\State \textbf{View generation and geometry:} sample $\phi_G, \phi_M, \phi_L$; compute $T_M, T_L$ and masks $M_M, M_L$.
\State \textbf{Encode:} $Z_G^{\text{Teacher}} \!\leftarrow f_{\bar\theta}(\phi_G)$ (stop-grad), 
$Z_s \!\leftarrow f_{\theta}(\phi_s)$ for $s \in \{M,L\}$.
\State \textbf{MPA:} predict offsets $\delta_s(v_j)$; build bilinear resampler around 
$\widehat T_s = T_s + \delta_s$; transport $\widetilde Z_s \!\leftarrow \Pi^{\text{MPA}}_s Z_s$ 
(or Equation \ref{eq:adapter-def} when using the adapter).
\State \textbf{Head and distributions:} $\ell_G \!\leftarrow H Z_G^{\text{Teacher}}$, 
$\ell_s \!\leftarrow H \widetilde Z_s$; 
$q \!\leftarrow \mathrm{softmax}((\ell_G - c_t)/\tau_T)$, 
$p_s \!\leftarrow \mathrm{softmax}(\ell_s/\tau_s)$.
\State \textbf{Losses (masked, size-normalised):} 
$\mathcal{L}_{G \to s} \!\leftarrow\! 
\alpha_s \, \|M_s\|_1^{-1} \! \sum_i M_s(i)\,\mathcal{D}(q(i)\|p_s(i))$ for $s \in \{M,L\}$; 
$\mathcal{L}_{\text{PSMamba}} \!\leftarrow \mathcal{L}_{G \to M} + \mathcal{L}_{G \to L}$.
\State \textbf{Update:} minimise $\mathcal{L}_{\text{PSMamba}}$ w.r.t. $(\theta,H)$; update centre $c_t$; 
EMA update $\bar\theta \!\leftarrow\! m\bar\theta + (1-m)\theta$.
\end{algorithmic}
\end{algorithm}

\subsubsection{Distillation on the Global Grid.}
A linear head $H \in \mathbb{R}^{K \times d}$ maps token features to logits: 

\begin{equation}
\begin{aligned}
&\ell_G \;=\; H \, Z_G^{\text{Teacher}},\\
&\ell_s \;=\; H \, \widetilde Z_s,\\
& s \in \{M,L\},
\end{aligned}
\label{eq:distill-logits}
\end{equation}

Following DINO \cite{caron2021emerging}, we apply centering and temperature scaling:
\begin{equation}
\begin{aligned}
&q \;=\; \mathrm{softmax}\!\left(\frac{\ell_G - c_t}{\tau_T}\right),\\
&p_s \;=\; \mathrm{softmax}\!\left(\frac{\ell_s}{\tau_s}\right),
\end{aligned}
\label{eq:distill-dists}
\end{equation}

where $c_t$ is an EMA of $\ell_G$. 
Because  $\widetilde Z_s$ lies on $\Lambda_G$, the teacher distribution $q(i)$ and each student
distribution $p_s(i)$ correspond to the same Global token $u_i$. 
Centring $c_t$ stabilises the teacher distribution, while the teacher temperature $\tau_T$ is annealed from soft to sharp, and the student temperature $\tau_s$ is fixed or gently annealed.

\subsubsection{Consistency on Visible Tokens}
Mid and Local views only cover subsets of $\Lambda_G$. 
To avoid label noise, supervision is restricted to visible tokens $M_s$, normalised by their count.
The masked, size-normalised teacher$\rightarrow$student loss is defined as
\begin{equation}
\mathcal{L}_{T \to S}
= \sum_{s \in \{M,L\}} \alpha_s \frac{1}{\|M_s\|_1}
\sum_{i=1}^{N_G} M_s(i)\,
\mathrm{KL}\!\left(q(i)\,\middle\|\,p_s(i)\right)
\label{eq:t2s}
\end{equation}

where $M_s \in \{0,1\}^{N_G}$ is the visibility mask of view $s$ and $\|M_s\|_1 = \sum_i M_s(i)$. For each view $s \in \{M,L\}$, we compute the average token-wise divergence between the teacher distribution
$q(i)$ and the student distribution $p_s(i)$ only on visible tokens. The weights $\alpha_s$ follow a progressive schedule, emphasising Local detail early in training and Mid-scale relations later.

\subsubsection{Objective and Optimisation.}
The total distillation loss combines Global$\to$Mid and Global$\to$Local objectives:
\begin{equation}
\mathcal{L}_{\text{PSMamba}}
=
\lambda_1 \, \mathcal{L}_{G \to M}
+
\lambda_2 \, \mathcal{L}_{G \to L},
\label{eq:g2v-total}
\end{equation}
where $\lambda_1$ and $\lambda_2$ are progressive schedule weights that emphasise Local cues in the early stages of training and then shift toward Mid-scale structure later. Here, $\mathcal{L}_{G \to M}$ and $\mathcal{L}_{G \to L}$ correspond to the masked teacher$\to$student loss (Equation \ref{eq:t2s}) restricted to the Mid and Local views, respectively.

The EMA teacher is updated by momentum:
\begin{equation}
\begin{aligned}
&\bar\theta \;\leftarrow\; m\,\bar\theta + (1-m)\,\theta,\\
&m \in [0.996,\,0.999].
\end{aligned}
\label{eq:ema}
\end{equation}

In practice, the curriculum weights $\lambda_1$ and $\lambda_2$ control the relative focus of Local vs. Mid supervision over time. The teacher temperature $\tau_T(t)$ is annealed from soft to sharp, while the student temperature $\tau_s$ remains fixed or is gently annealed.

This progressive distillation strategy mirrors the diagnostic process in plant pathology, where early detection relies on fine-grained cues before integrating broader structural patterns for robust recognition. The overall training procedure, including multi-granular view generation, MPA alignment, and dual-student supervision, is summarised step by step in Algorithm \ref{alg:psmamba}.


\section{Results and Discussion}

This section evaluates how PSMamba performs across controlled, real-field, and fine-grained plant disease scenarios and examines the specific contributions of its multi-granular design. We first describe the experimental setup and dataset characteristics, then compare PSMamba with state-of-the-art baselines. Targeted ablations isolate the roles of the Global, Mid, and Local branches and the MPA module. Finally, qualitative analyses show that PSMamba produces more coherent, disease-focused attention and more discriminative embedding structures than competing methods.

\subsection{Experimental Setup}
To validate the effectiveness of the proposed PSMamba framework, we conducted extensive experiments using a high-performance GPU server equipped with 10$\times$ NVIDIA GeForce RTX 2080 Ti GPUs (11\, GB each). All models were implemented in PyTorch 2.1.1 and leveraged the Mamba SSM v2.2.3 library for efficient state-space modelling. Multi-GPU training was enabled via DistributedDataParallel (DDP) with NCCL as the backend, using CUDA 11.8 and Python 3.10.13 on Ubuntu 20.04.

\begin{table}[!t]
\centering
\scriptsize
\caption{Performance of different self-supervised methods on the \textbf{PlantVillage} dataset. The best-performing method is marked in bold.}
\label{tab:pv}
\begin{tabular}{lcccccc}
\hline
\textbf{Methods} & \textbf{Year} & \textbf{Backbone} & \textbf{Accuracy} & \textbf{Precision} & \textbf{Recall} & \textbf{F1-score} \\
\hline
SimCLR \cite{chen2020simple}         & 2020 & CNN         & 88.67 & 87.94 & 88.11 & 88.02 \\
Clustering \cite{monowar2022self}    & 2022 & CNN         & 88.93 & --    & --    & --    \\
CIKICS \cite{fang2021self}           & 2021 & CNN         & 89.14 & 89.42 & 83.73 & 79.62 \\
MoCo \cite{he2020momentum}         & 2021 & CNN         & 89.23 & 88.37 & 88.66 & 88.48 \\
BYOL \cite{grill2020bootstrap}       & 2020 & CNN         & 90.06 & 89.33 & 89.47 & 89.38 \\
MaskCOV \cite{yu2021maskcov}         & 2023 & CNN         & 92.03 & 91.34 & 91.59 & 91.43 \\
DeiT \cite{touvron2021training}      & 2021 & Transformer & 92.17 & 91.52 & 91.87 & 91.69 \\
TransFG \cite{he2022transfg}         & 2021 & Transformer & 92.83 & 92.07 & 92.31 & 92.18 \\
MAE \cite{he2022masked}              & 2022 & Transformer & 92.94 & 92.27 & 92.43 & 92.34 \\
ViT \cite{dosovitskiy2020image} & 2024 & Transformer & 93.06 & 92.31 & 92.62 & 92.41 \\
DINO \cite{caron2021emerging}        & 2021 & Transformer & 93.12 & 92.47 & 92.83 & 92.64 \\
ASCL \cite{chen2021exploring}        & 2025 & CNN         & 93.50 & --    & --    & --    \\
Mix-ViT \cite{yu2023mix}             & 2024 & Transformer & 97.81 & 97.22 & 97.04 & 97.11 \\
IEM-ViT \cite{zhang2023information}  & 2024 & Transformer & 98.04 & 97.41 & 97.22 & 97.29 \\
CAE \cite{bedi2021plant}             & 2021 & CNN         & 98.41 & 96.21 & 97.11 & 96.65 \\
ConMamba \cite{mamun2025conmamba}    & 2025 & Mamba       & 98.63 & 96.74 & 97.59 & 97.38 \\
\textbf{PSMamba}                     & --   & Mamba       & \textbf{98.87} & \textbf{97.43} & \textbf{97.72} & \textbf{97.69} \\
\hline
\end{tabular}
\end{table}

\subsection{Evaluation Metrics}

Model performance was assessed using four widely adopted metrics in visual recognition: accuracy, precision, recall, and F1-score. Since real-world plant disease datasets often exhibit varying levels of class imbalance, we additionally report macro-F1 and minority-F1 to capture behaviour under uneven label distribution. Beyond quantitative evaluation, we employ t-SNE visualisation to analyse the separability of learned representations and Grad-CAM heatmaps to examine whether the model attends to disease-relevant regions. Together, these quantitative and qualitative metrics provide a comprehensive evaluation of PSMamba across balanced, imbalanced, and fine-grained scenarios.

\subsection{Datasets}
We evaluate PSMamba on three publicly available plant disease benchmarks that together span controlled laboratory imagery, complex in-field conditions, and fine-grained classification scenarios. The first of these is \textbf{PlantVillage} \cite{hughes2015open}, which represents the most controlled setting, with more than 54,000 images captured under uniform backgrounds and consistent lighting. Its 38 disease categories across 14 crop species provide broad class diversity while minimising environmental noise, making it suitable for analysing in-domain representational quality. In contrast, \textbf{PlantDoc} \cite{singh2020plantdoc} introduces substantial real-world variability. It's 2,598 images span 28 disease classes but differ from PlantVillage through natural illumination, varied viewpoints, occlusions, and cluttered field backgrounds. These characteristics introduce a domain shift and moderate class imbalance, allowing us to examine robustness under uncontrolled acquisition conditions. Complementing both datasets, the \textbf{Citrus} \cite{rauf2019citrus} dataset focuses on fine-grained discrimination. It contains 609 high-resolution field images across 5 visually similar disease categories, where lesions are subtle and localised. The limited dataset size further stresses the model’s ability to extract discriminative features when supervision is scarce and inter-class differences are minimal.

Overall, the combination of PlantVillage, PlantDoc, and Citrus offers a balanced range of evaluation conditions, from clean laboratory images to highly variable field environments and subtle fine-grained categories. This diversity ensures a robust assessment of PSMamba, allowing us to examine how its multi-scale design adapts to increasing visual complexity and real-world variability.

\begin{table}[!t]
\centering
\scriptsize
\caption{Performance of different self-supervised methods on the \textbf{PlantDoc} dataset. The best-performing method is marked in bold.}
\label{tab:pd}
\begin{tabular}{lcccccc}
\hline
\textbf{Methods} & \textbf{Year} & \textbf{Backbone} & \textbf{Accuracy} & \textbf{Precision} & \textbf{Recall} & \textbf{F1-score} \\
\hline
Unified \cite{huan2025unified}               & 2025 & CNN          & 77.82 & 80.00 & 78.24 & 77.48 \\
SimCLR \cite{chen2020simple}                 & 2020 & CNN          & 81.43 & 80.67 & 81.04 & 80.85 \\
MoCo \cite{he2020momentum}                 & 2021 & CNN          & 82.13 & 81.63 & 81.81 & 81.72 \\
DeiT \cite{touvron2021training}              & 2021 & Transformer  & 82.94 & 82.31 & 82.59 & 82.44 \\
TransFG \cite{he2022transfg}                 & 2021 & Transformer  & 83.24 & 82.70 & 82.92 & 82.81 \\
BYOL \cite{grill2020bootstrap}               & 2020 & CNN          & 83.53 & 82.93 & 83.13 & 83.03 \\
SimSiam \cite{chen2021exploring}             & 2021 & CNN          & 84.76 & 83.92 & 84.33 & 84.12 \\
SwAV \cite{caron2020unsupervised}            & 2020 & CNN          & 84.95 & 84.21 & 83.94 & 84.07 \\
MaskCOV \cite{yu2021maskcov}                & 2023 & CNN          & 86.03 & 85.39 & 85.22 & 85.30 \\
ContraL \cite{chung2024addressing}           & 2024 & CNN          & 87.46 & 86.73 & 86.27 & 86.48 \\
MAE \cite{he2022masked}                      & 2022 & Transformer  & 88.14 & 87.53 & 87.72 & 87.62 \\
DINO \cite{caron2021emerging}                & 2021 & Transformer  & 88.23 & 87.72 & 87.91 & 87.81 \\
ViT \cite{dosovitskiy2020image}       & 2024 & Transformer  & 88.33 & 87.61 & 87.89 & 87.73 \\
Mix-ViT \cite{yu2023mix}                     & 2024 & Transformer  & 92.63 & 92.04 & 91.82 & 91.92 \\
IEM-ViT \cite{zhang2023information}          & 2024 & Transformer  & 93.17 & 92.49 & 92.31 & 92.40 \\
ConMamba \cite{mamun2025conmamba}            & 2025 & Mamba        & 94.29 & 93.88 & 93.87 & 93.97 \\
\textbf{PSMamba}                              & --   & Mamba        & \textbf{95.64} & \textbf{94.23} & \textbf{94.27} & \textbf{94.11} \\
\hline
\end{tabular}
\end{table}

\subsection{Experimental results}

Tables \ref{tab:pv}--\ref{tab:citrus} present a comprehensive comparison of PSMamba with the state-of-the-art self-supervised and state-space plant disease recognition methods across the PlantVillage, PlantDoc, and Citrus benchmarks. Across all datasets, PSMamba consistently achieves the highest performance on all evaluation metrics, demonstrating strong in-domain representational quality, robustness under real-field domain shift, and fine-grained discriminative capability. On average across three benchmarks, PSMamba improves accuracy by +0.24\%, +1.35\%, and +0.88\% over ConMamba on PlantVillage, PlantDoc, and Citrus, respectively, with similarly consistent gains across the other evaluation measures. Together, these results establish PSMamba as a unified SSL framework that generalises effectively across diverse visual conditions and disease manifestations.

\begin{table}[!t]
\centering
\scriptsize
\caption{Performance of different self-supervised methods on the \textbf{Citrus} dataset. The best-performing method is marked in bold.}
\label{tab:citrus}
\begin{tabular}{lcccccc}
\hline
\textbf{Methods} & \textbf{Year} & \textbf{Backbone} & \textbf{Accuracy} & \textbf{Precision} & \textbf{Recall} & \textbf{F1-score} \\
\hline
CIKICS \cite{fang2021self}                & 2021 & CNN          & 64.13 & 52.13 & 42.19 & 46.58 \\
SimCLR \cite{chen2020simple}              & 2020 & CNN          & 86.14 & 85.63 & 85.82 & 85.72 \\
MoCo \cite{he2020momentum}              & 2021 & CNN          & 87.03 & 86.41 & 86.71 & 86.52 \\
BYOL \cite{grill2020bootstrap}            & 2020 & CNN          & 87.24 & 86.69 & 86.91 & 86.80 \\
SimSiam \cite{chen2021exploring}          & 2021 & CNN          & 87.53 & 86.72 & 86.41 & 86.55 \\
DeiT \cite{touvron2021training}           & 2021 & Transformer  & 89.03 & 88.59 & 88.17 & 88.39 \\
MAE \cite{he2022masked}                   & 2022 & Transformer  & 89.13 & 88.62 & 88.33 & 88.44 \\
Clustering \cite{monowar2022self}         & 2022 & CNN          & 89.33 & 88.69 & 88.85 & 88.77 \\
DINO \cite{caron2021emerging}             & 2021 & Transformer  & 89.44 & 88.93 & 88.71 & 88.82 \\
TransFG \cite{he2022transfg}              & 2021 & Transformer  & 89.53 & 88.93 & 89.09 & 89.01 \\
MaskCOV \cite{yu2021maskcov}              & 2023 & CNN          & 89.63 & 89.14 & 88.74 & 88.92 \\
ViT \cite{dosovitskiy2020image}    & 2024 & Transformer  & 89.73 & 89.22 & 88.97 & 89.10 \\
Mix-ViT \cite{yu2023mix}                  & 2024 & Transformer  & 90.27 & 89.61 & 89.41 & 89.50 \\
IEM-ViT \cite{zhang2023information}       & 2024 & Transformer  & 90.63 & 89.97 & 89.59 & 89.77 \\
ConMamba \cite{mamun2025conmamba}         & 2025 & Mamba        & 91.38 & 91.74 & 90.51 & 91.66 \\
\textbf{PSMamba}                           & --   & Mamba        & \textbf{92.26} & \textbf{91.81} & \textbf{91.42} & \textbf{91.71} \\
\hline
\end{tabular}
\end{table}

\paragraph{Results on PlantVillage dataset (Table \ref{tab:pv})}
The first set of results on the PlantVillage dataset highlights PSMamba’s representational strength under controlled imaging conditions.  
PSMamba achieves an accuracy of 98.87\%, outperforming all competing models, including the transformer-based Mix-ViT \cite{yu2023mix}, IEM-ViT \cite{zhang2023information}, and CAE \cite{bedi2021plant}, as well as the state-space model ConMamba \cite{mamun2025conmamba}. Because PlantVillage contains clean backgrounds and consistent illumination, its classes are primarily distinguished by lesion morphology and colour–texture patterns rather than environmental noise. PSMamba’s hierarchical multi-granular design, which integrates global structure, mid-scale vein and lesion interactions, and fine-grained texture cues, enables a more comprehensive modelling of these variations. Compared with ConMamba \cite{mamun2025conmamba}, PSMamba achieves an improvement of 0.24\% in accuracy and 0.31\% in F1-score, showing that multi-scale distillation combined with bidirectional state-space encoding provides complementary lesion cues that single-scale models \cite{mamun2025conmamba,yu2021maskcov,bedi2021plant} do not capture. While transformer-based models such as DeiT \cite{touvron2021training}, TransFG \cite{he2022transfg}, MAE \cite{he2022masked}, and DINO \cite{caron2021emerging} deliver strong performance, their accuracy tends to fall in the 92--93\% range, reflecting the difficulty of jointly capturing global and local cues and the computational burden associated with high-resolution patch processing. In contrast, PSMamba’s linear-time bidirectional recurrence establishes an efficient and lesion-aware inductive bias suitable for controlled conditions.

\paragraph{Results on PlantDoc dataset (Table \ref{tab:pd})}
The results on the PlantDoc dataset demonstrate PSMamba’s ability to generalise effectively under complex, real-field conditions. 
PSMamba achieves an accuracy of 95.64\%, surpassing all prior methods. It provides a gain of 1.35\% over ConMamba \cite{mamun2025conmamba} and improvements ranging from 7--15\% over earlier self-supervised methods, including SimCLR \cite{chen2020simple}, MoCo \cite{he2020momentum}, and BYOL \cite{grill2020bootstrap}, among others. 
These significant gains highlight the importance of the MPA module and the dual-student distillation mechanism, both of which align tokens across resolutions and maintain coherent supervision across scale-specific evidence regions. Traditional self-supervised models often struggle in PlantDoc because background artefacts or high-contrast distractors in the scene mislead their attention mechanisms. In contrast, the bidirectional state-space recurrence in PSMamba encourages smooth spatial propagation of contextual information, allowing the model to focus on disease-relevant leaf regions even when symptoms are subtle or partially occluded. PSMamba therefore experiences far less degradation under domain shift than transformer-based models that perform well on controlled datasets but drop significantly on real-field imagery.

\begin{figure}[!t]
\centering
\begin{tikzpicture}
  \node[inner sep=0pt] (img) {
    \begin{tabular}{ccc ccc ccc}
        \rotatebox{90}{\scriptsize ~~~~Original} &  
        \includegraphics[width=0.12\textwidth]{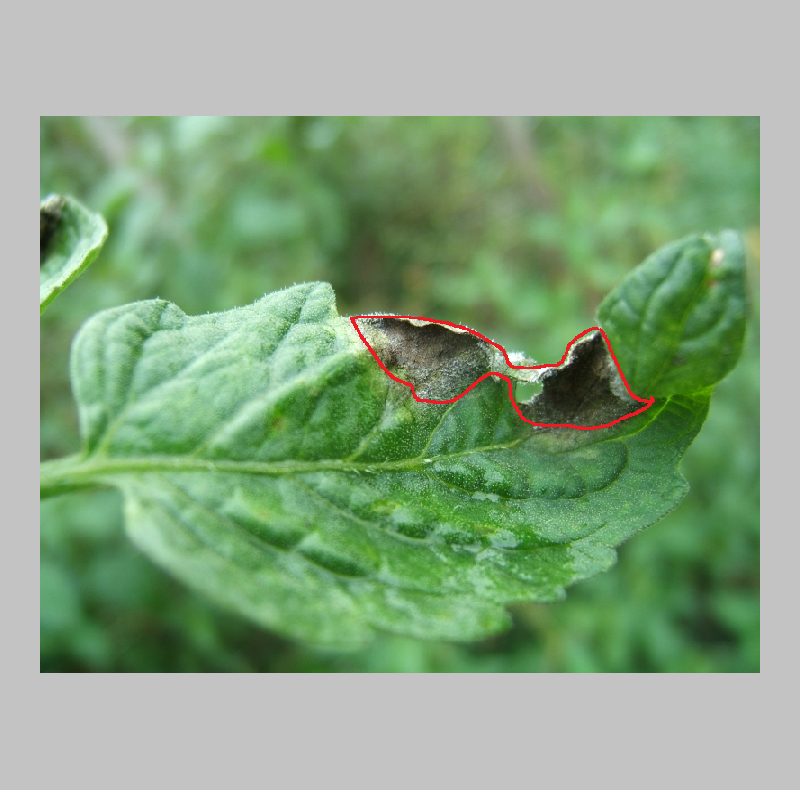} &
        \includegraphics[width=0.12\textwidth]{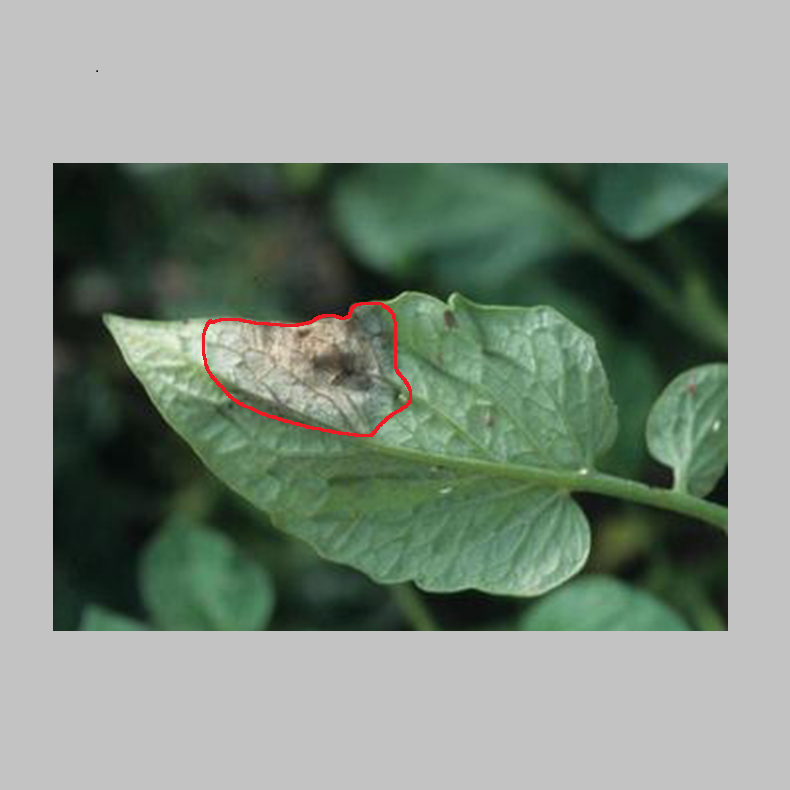} &
        \includegraphics[width=0.12\textwidth]{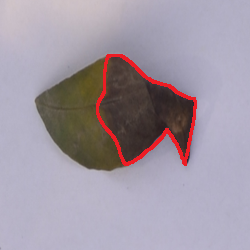} &
        \includegraphics[width=0.12\textwidth]{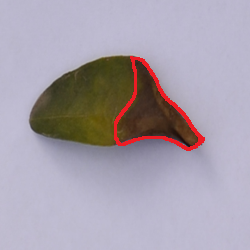} &
        \includegraphics[width=0.12\textwidth]{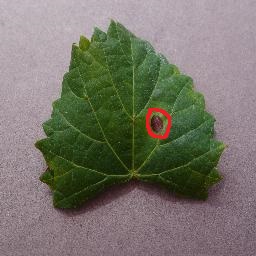} &
        \includegraphics[width=0.12\textwidth]{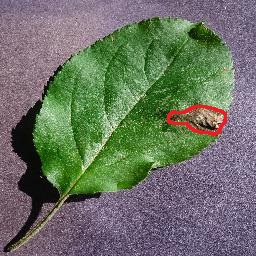} \\ 

        \rotatebox{90}{\scriptsize ConMamba \cite{mamun2025conmamba}} &  
        \includegraphics[width=0.12\textwidth]{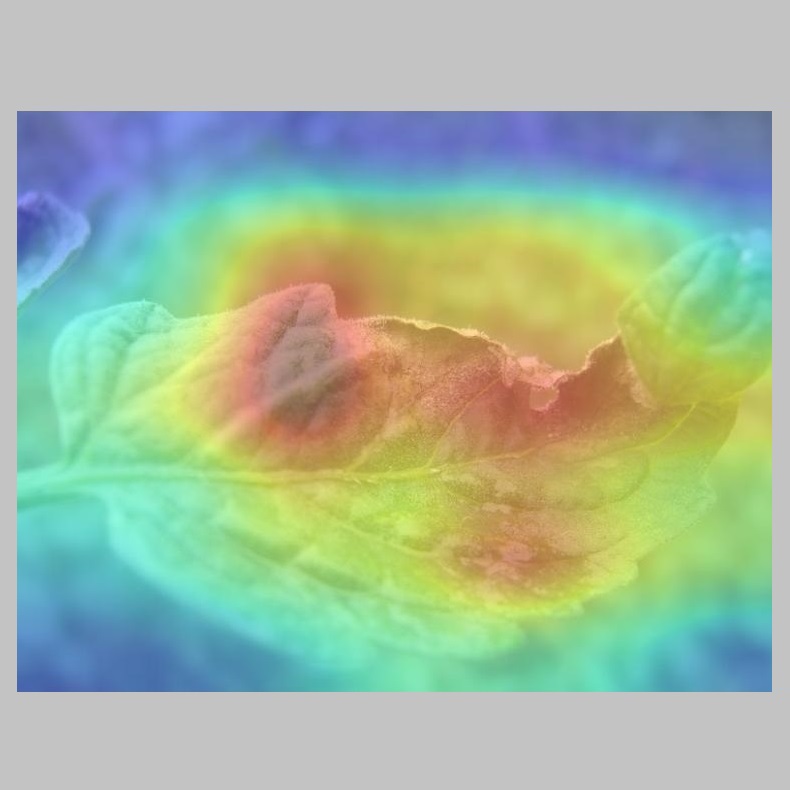} &
        \includegraphics[width=0.12\textwidth]{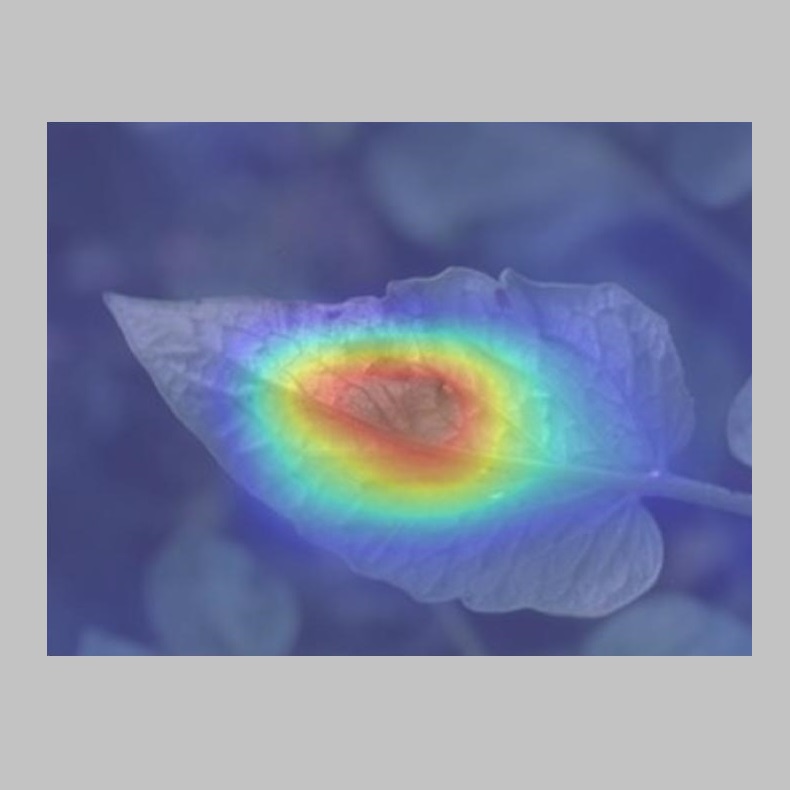} &
        \includegraphics[width=0.12\textwidth]{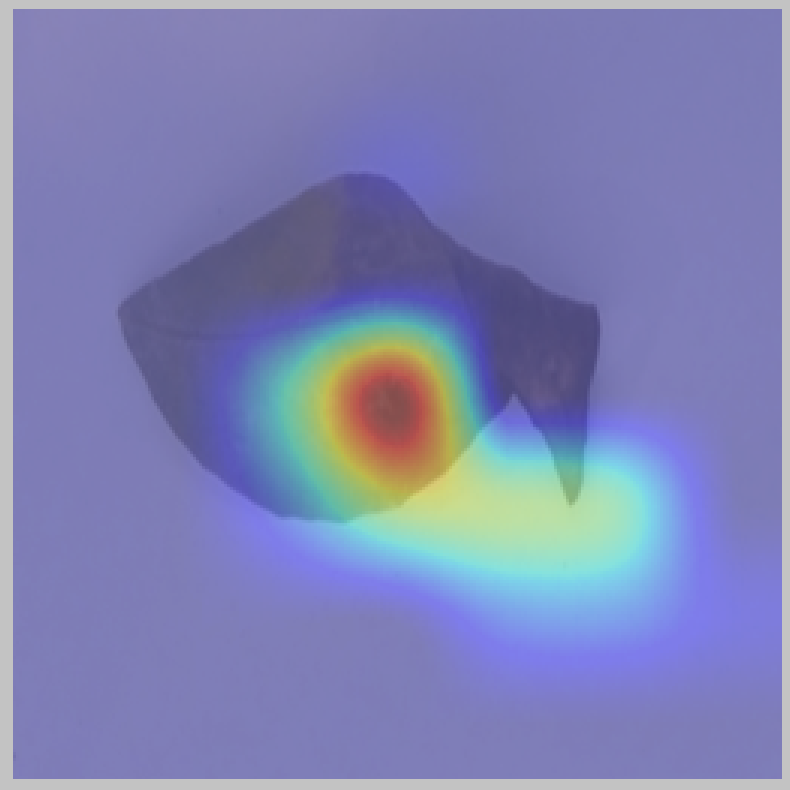} &
        \includegraphics[width=0.12\textwidth]{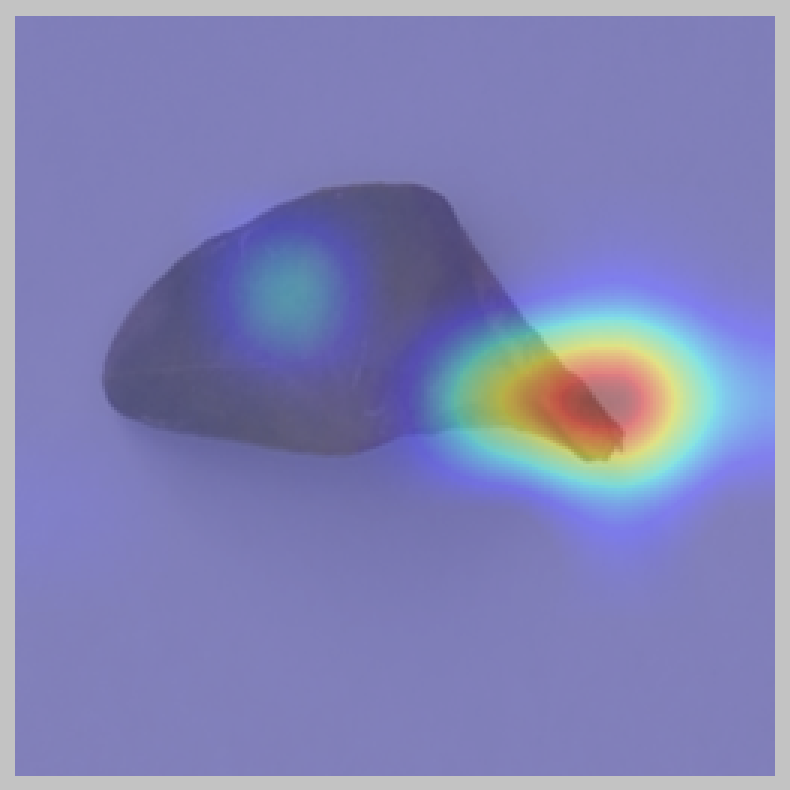} &
        \includegraphics[width=0.12\textwidth]{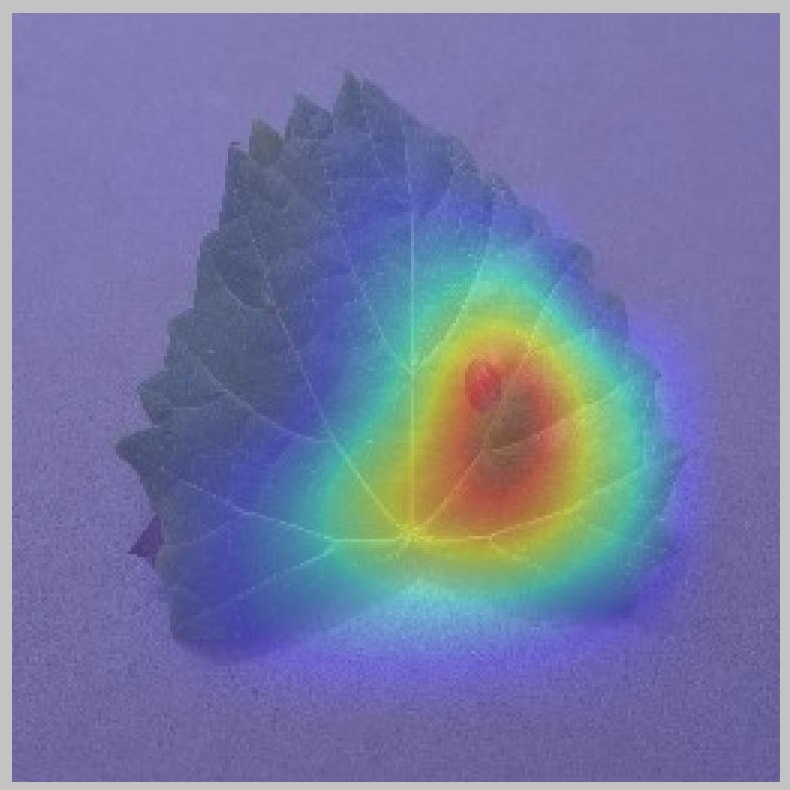} &
        \includegraphics[width=0.12\textwidth]{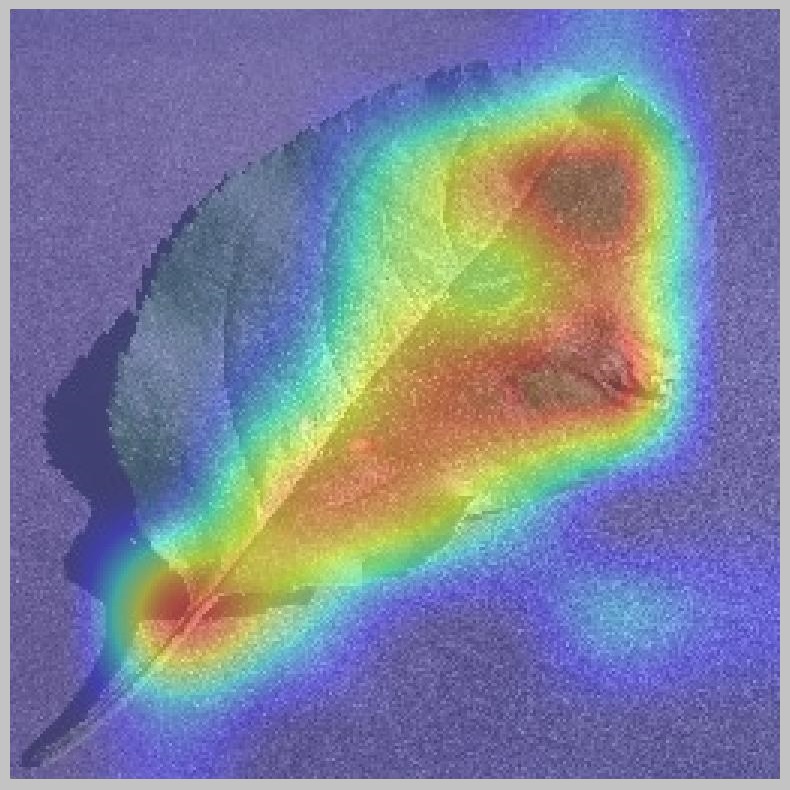} \\ 

        \rotatebox{90}{\scriptsize ~~~PSMamba} &  
        \includegraphics[width=0.12\textwidth]{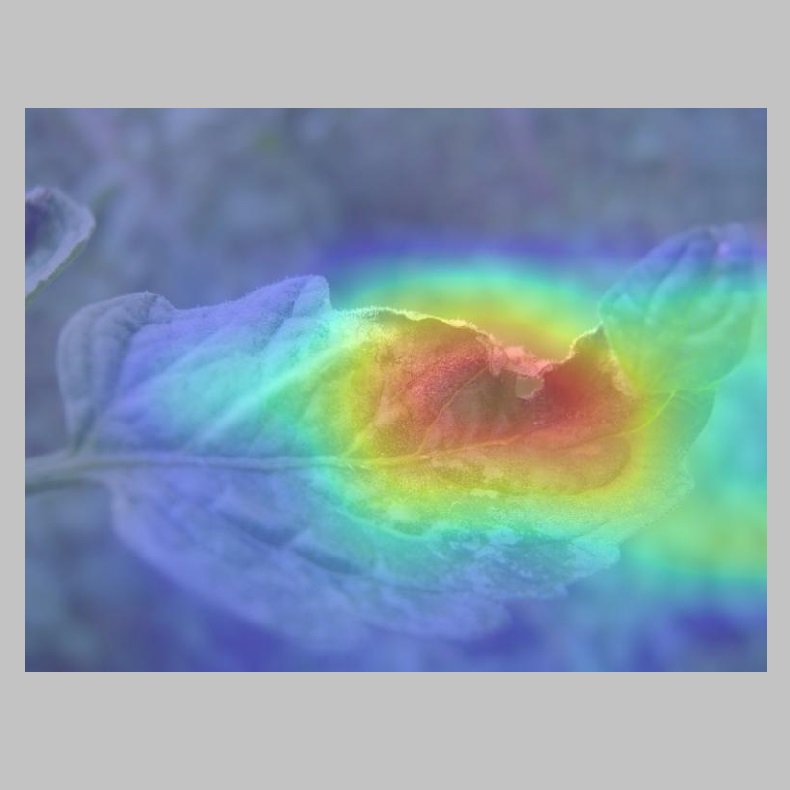} &
        \includegraphics[width=0.12\textwidth]{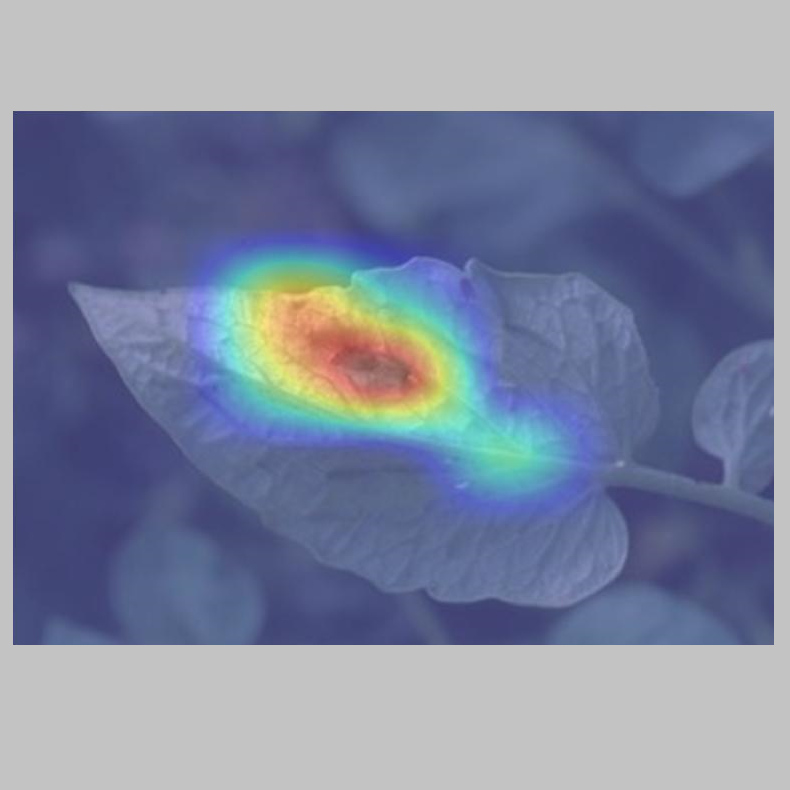} & 
        \includegraphics[width=0.12\textwidth]{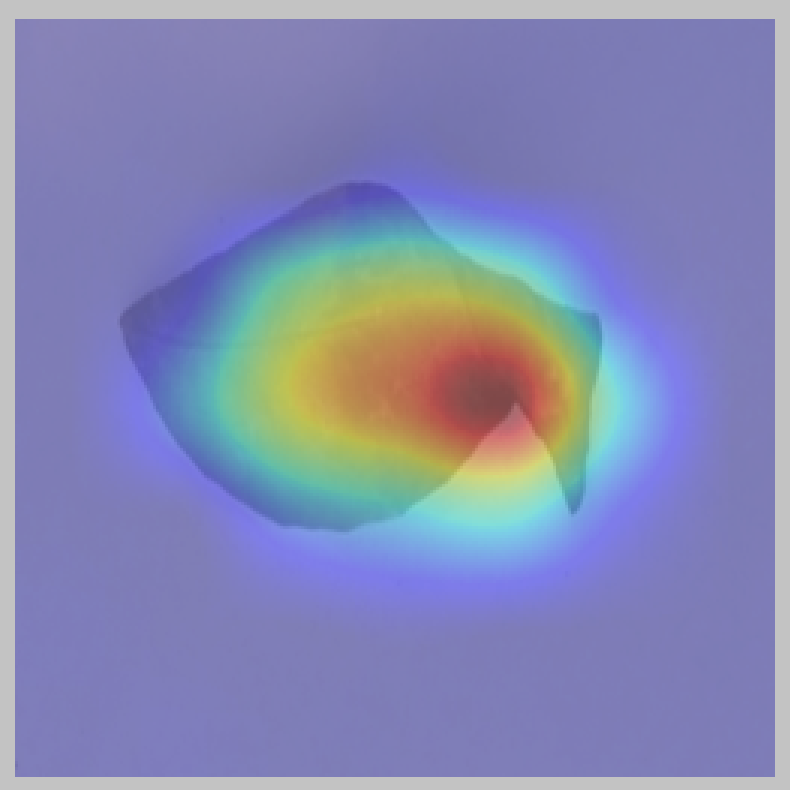} &
        \includegraphics[width=0.12\textwidth]{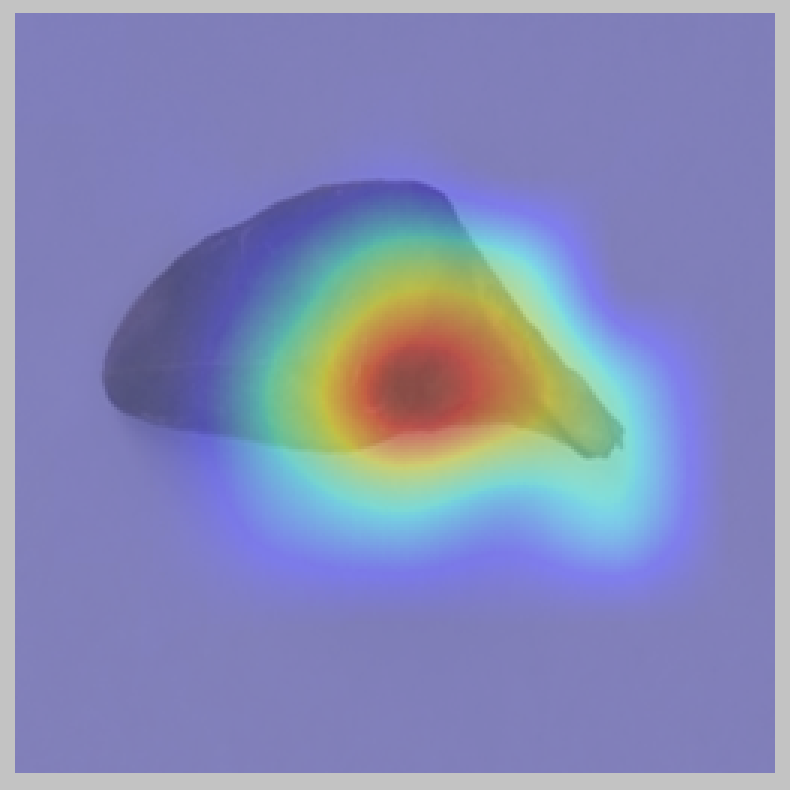} &
        \includegraphics[width=0.12\textwidth]{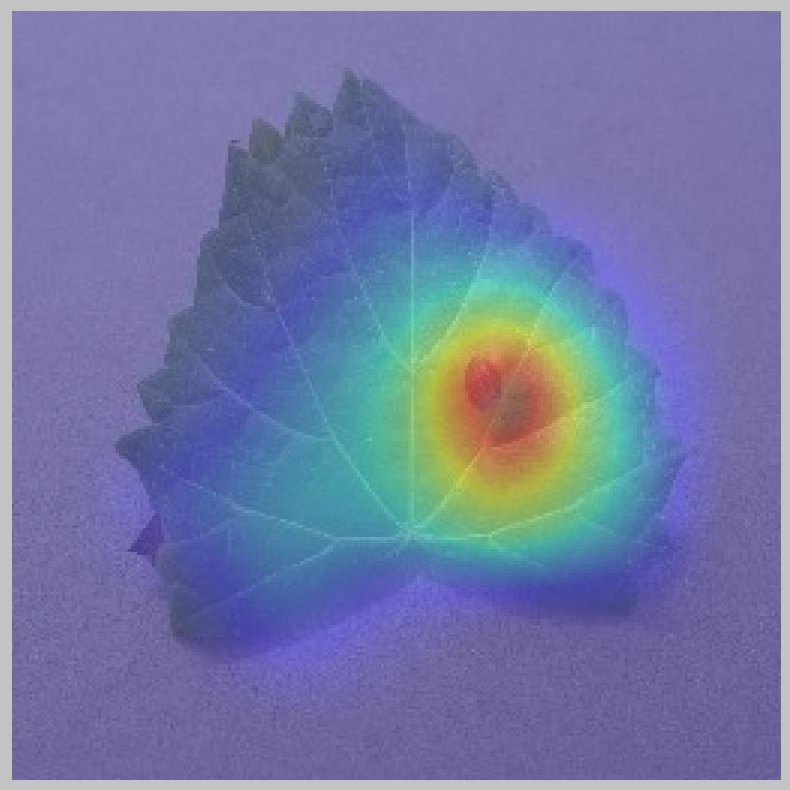} &
        \includegraphics[width=0.12\textwidth]{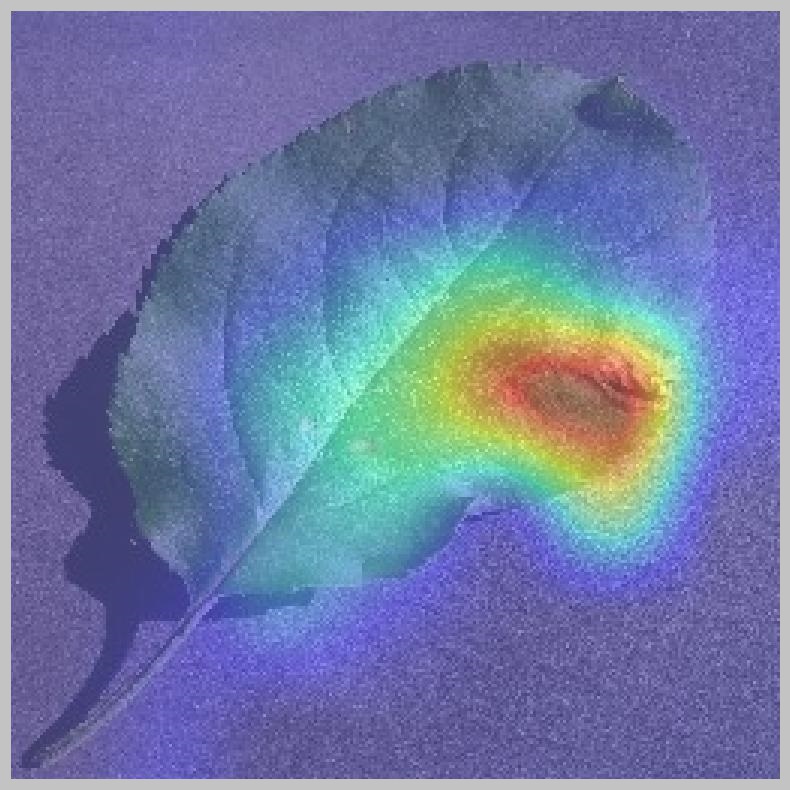} \\ 
    \end{tabular}
  };

  \draw[dash pattern=on 10pt off 4pt, line width=1pt] ([yshift=-2pt]img.north east) ++(-9.4cm,0) -- ++(0,-6.1cm);
  \draw[dash pattern=on 10pt off 4pt, line width=1pt] ([yshift=-2pt]img.north east) ++(-4.7cm,0) -- ++(0,-6.1cm);

\end{tikzpicture}

\vspace{0.2em}
\begin{tabular}{m{0.33\linewidth} m{0.33\linewidth} m{0.33\linewidth}} 
    \centering \scriptsize ~~~~~~~~~~~~~~~~~~~PlantDoc & \centering \scriptsize Citrus~~~~~~ & \centering \scriptsize PlantVillage~~~~~~~~~~~~~~~~~~~~~~~~~~~
\end{tabular}
\vspace{-1.5em}
\caption{Grad-CAM maps of PSMamba and the second-best method, ConMamba \cite{mamun2025conmamba} on test datasets. 
First Row: randomly selected test images. 
Second row: Grad-CAM maps from the ConMamba \cite{mamun2025conmamba}. Third row: Grad-CAM maps from the proposed PSMamba framework.}
\label{fig:gradcam}

\end{figure}

\paragraph{Results on Citrus dataset (Table \ref{tab:citrus})}
This dataset evaluates fine-grained discrimination among visually similar diseases. PSMamba achieves 92.26\% accuracy, outperforming ConMamba by 0.88\% and exceeding transformer-based baselines, including Mix-ViT \cite{yu2023mix}, IEM-ViT \cite{zhang2023information}, and CAE \cite{bedi2021plant}. The Citrus dataset requires distinguishing between highly similar lesion shapes and micro-level texture variations that occupy small spatial regions. Many self-supervised models, including MaskCOV \cite{yu2021maskcov}, Mix-ViT \cite{yu2023mix}, and ConMamba \cite{mamun2025conmamba}, struggle with these tasks because global-alignment strategies often suppress the fine-grained details needed to separate closely related symptom classes. PSMamba addresses these challenges through the local-student branch, which preserves high-resolution micro-patterns, and the mid-student branch, which reinforces spatial continuity cues. The integration of fine-grained detail with broader contextual cues makes PSMamba particularly effective for fine-grained disease classification.

Taken together, these results demonstrate that PSMamba delivers strong and consistent improvements across test datasets. By combining bidirectional VM encoding, a hierarchical multi-granular crop structure, alignment-aware positional adaptation, and dual-student distillation, PSMamba captures the full spectrum of plant disease symptoms, ranging from micro-level lesion textures to macro-level structural patterns. This integrated formulation yields robust, generalisable, and disease-focused representations across all evaluated plant disease datasets.

\subsection{Ablation studies}

\subsubsection{Spatial Attention in PSMamba}
To further understand how PSMamba achieves its performance gains across diverse datasets, we analyse the spatial attention behaviour of the model using Grad-CAM visualisations, as shown in Figure 3. These heatmaps provide qualitative evidence of how effectively the framework localises disease-relevant regions across test datasets. The comparison focuses on three rows: the original test images, the Grad-CAM responses from the second-best method, ConMamba \cite{mamun2025conmamba}, and the corresponding activation maps from PSMamba.

\begin{table}[!t]
\centering
\scriptsize
\caption{Performance of different multi-granular view configurations in the proposed PSMamba framework across different datasets. Accuracy (\%) is reported as the evaluation metric.}
\label{tab:mmdp_ablation}
\begin{tblr}{cells = {c},vline{5} = {-}{},hline{1-2,6} = {-}{},}
\textbf{Configuration} & \textbf{Global} & \textbf{Mid} & \textbf{Local} & \textbf{PlantVillage} & \textbf{PlantDoc} & \textbf{Citrus} \\
Global view          & \cmark & \xmark & \xmark & 95.80 & 91.45 & 86.70 \\
Global + Local view  & \cmark & \xmark & \cmark & 97.82 & 93.30 & 90.10 \\
Global + Mid view    & \cmark & \cmark & \xmark & 96.90 & 92.95 & 88.50 \\
\textbf{PSMamba}     & \cmark & \cmark & \cmark & \textbf{98.46} & \textbf{95.58} & \textbf{92.15} \\
\end{tblr}
\end{table}

Across all datasets, PSMamba consistently demonstrates sharper, more spatially coherent, and lesion-focused attention compared with ConMamba \cite{mamun2025conmamba}. In the PlantDoc samples, ConMamba \cite{mamun2025conmamba} often exhibits diffuse activations that extend into non-diagnostic background regions. This behaviour is expected, as single-scale SSL frameworks tend to overfit high-contrast edges or textured backgrounds when field images exhibit strong environmental variance. In contrast, PSMamba produces compact and well-defined activation regions that align closely with the annotated lesion boundaries. The model focuses primarily on symptomatic structures such as necrotic patches, vein distortions, and colour-modulated disease zones, reflecting the benefit of multi-granular supervision and MPA-guided positional consistency. These coherent activation patterns indicate that PSMamba is more resilient to domain shifts and less influenced by scene-level distractors.

A similar trend is observed in the Citrus dataset, which emphasises fine-grained discrimination among diseases with highly similar lesion appearances. ConMamba’s \cite{mamun2025conmamba} activations tend to be broader, often covering large portions of the leaf surface without isolating the subtle boundary differences that differentiate one citrus disease from another. PSMamba, on the other hand, consistently highlights the core symptomatic regions, including micro-level texture variations, lesion edges, and finely localised discolouration patterns. The Local-student branch plays a key role here by learning high-resolution cues, while the Mid-student branch reinforces context-sensitive relationships along vein structures. These complementary pathways allow the model to focus on diagnostically meaningful structures even when symptoms occupy only a small portion of the leaf.

The PlantVillage examples further emphasise PSMamba’s ability to exploit multi-scale cues even under controlled imaging conditions. While both ConMamba \cite{mamun2025conmamba} and PSMamba identify lesion regions, ConMamba’s activations often remain broad and sometimes overly centralised, indicating difficulty in distinguishing between subtle texture gradients and broader leaf colouration. By contrast, PSMamba produces precise and strongly peaked activations that narrow down to the actual symptomatic areas without spillover into healthy areas. This behaviour reflects the benefits of progressive distillation and the alignment-aware strategy, which helps maintain representational consistency across the local, mid, and global scales. 

Taken together, the Grad-CAM visualisations show that PSMamba not only improves quantitative accuracy but also enhances the interpretability and reliability of spatial localisation. The model consistently focuses on disease-relevant structures across diverse resolutions and environmental conditions, producing attention maps that are sharper, more discriminative, and more coherent than those of ConMamba \cite{mamun2025conmamba}. 


\begin{figure}[!t]
\centering
\begin{tikzpicture}
  \node[inner sep=0pt] (img) {
    \begin{tabular}{ccc ccc c}
        \rotatebox{90}{\scriptsize ~~~~~Original} &  
        \includegraphics[width=0.13\textwidth]{Figures/visual/gradcam/pd_o_2.png} &
        \includegraphics[width=0.13\textwidth]{Figures/visual/gradcam/pd_o_1.png} & 
        \includegraphics[width=0.13\textwidth]{Figures/visual/gradcam/citrus_o_1.png} &
        \includegraphics[width=0.13\textwidth]{Figures/visual/gradcam/citrus_o_2.png} & 
        \includegraphics[width=0.13\textwidth]{Figures/visual/gradcam/pv_o_1.jpeg} &
        \includegraphics[width=0.13\textwidth]{Figures/visual/gradcam/pv_o_2.jpeg} \\ 

        \rotatebox{90}{\scriptsize ~~~~~~Global } &  
        \includegraphics[width=0.13\textwidth]{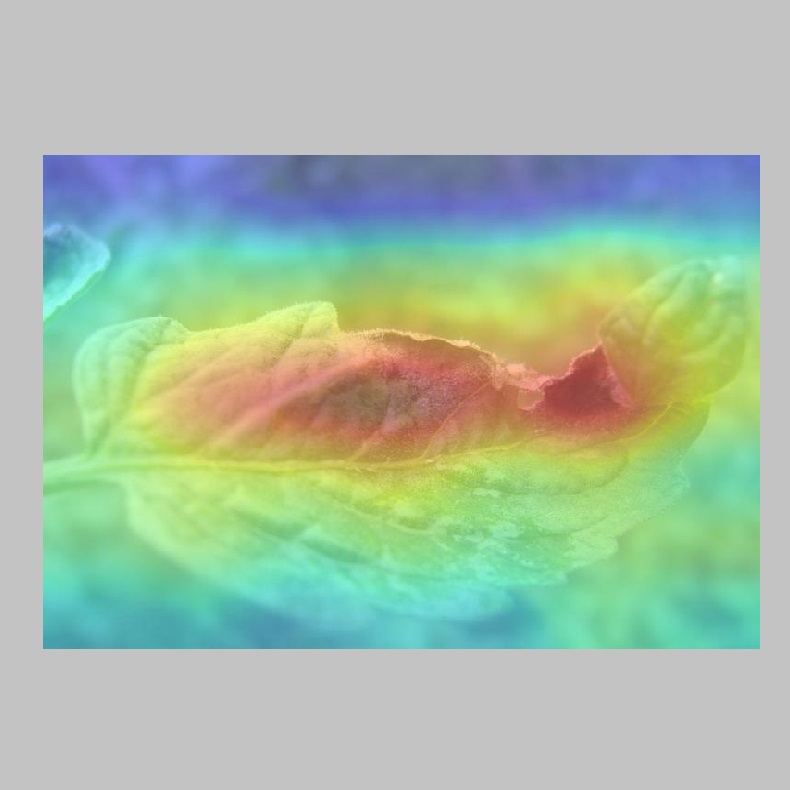} &
        \includegraphics[width=0.13\textwidth]{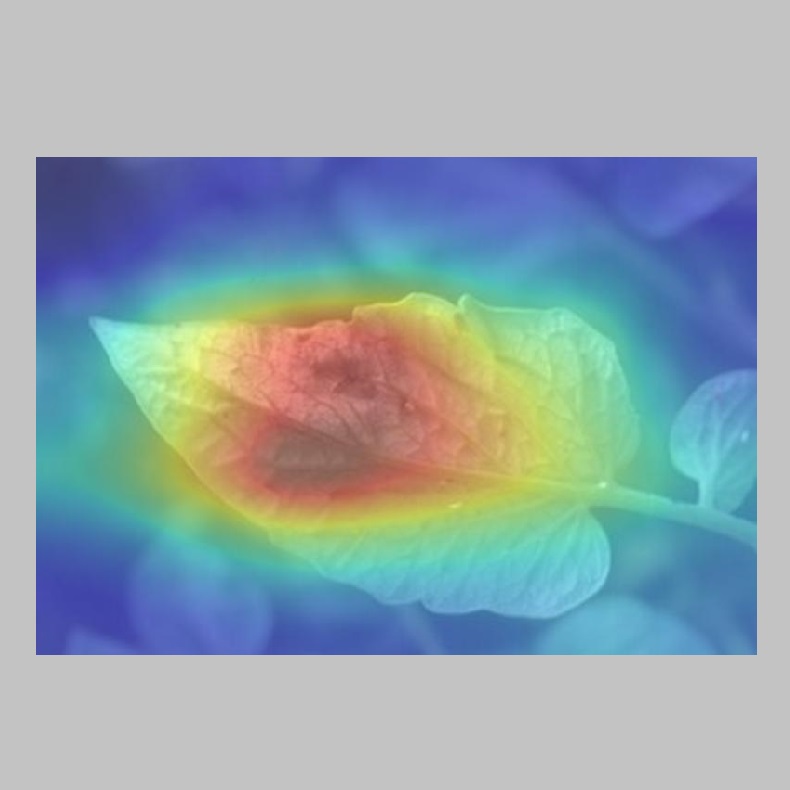} & 
        \includegraphics[width=0.13\textwidth]{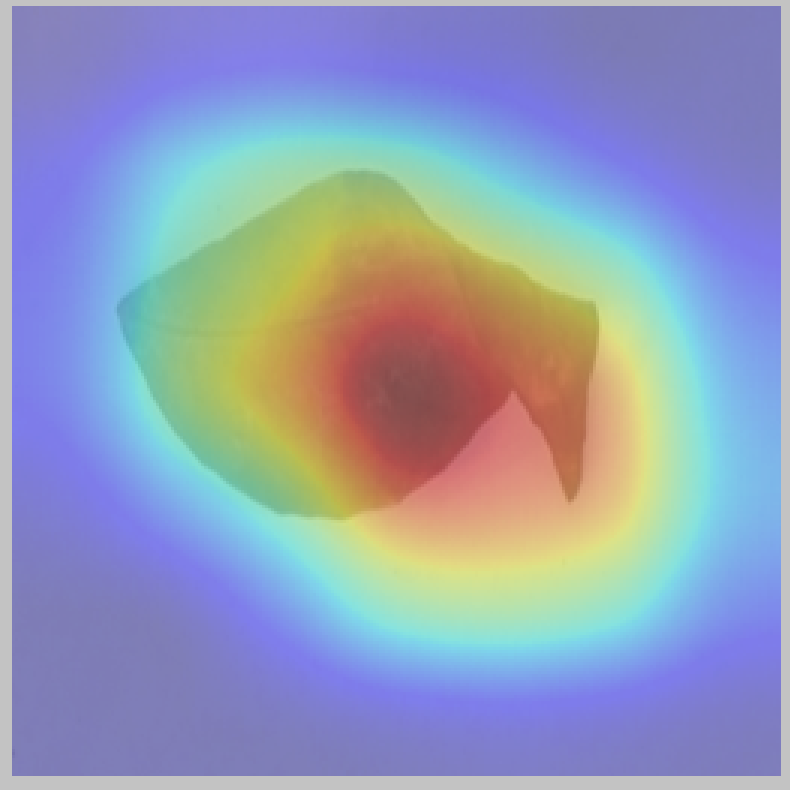} &
        \includegraphics[width=0.13\textwidth]{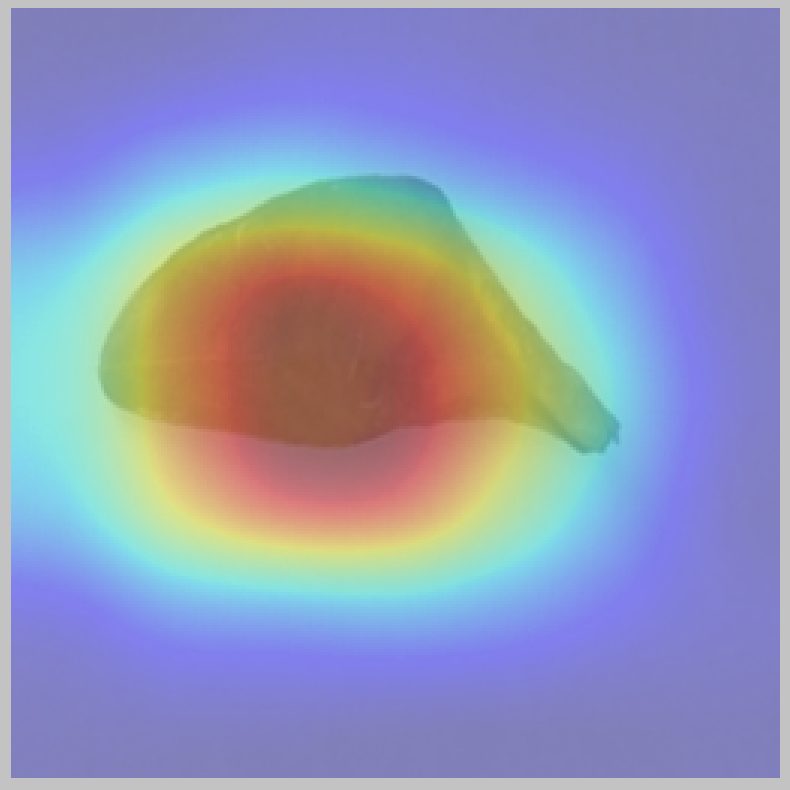} & 
        \includegraphics[width=0.13\textwidth]{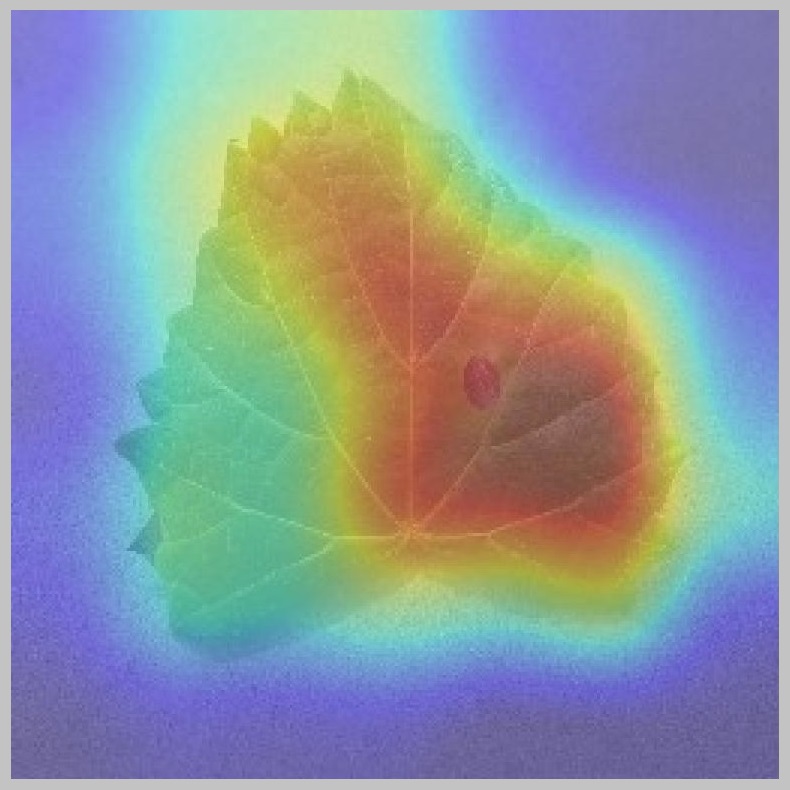} &
        \includegraphics[width=0.13\textwidth]{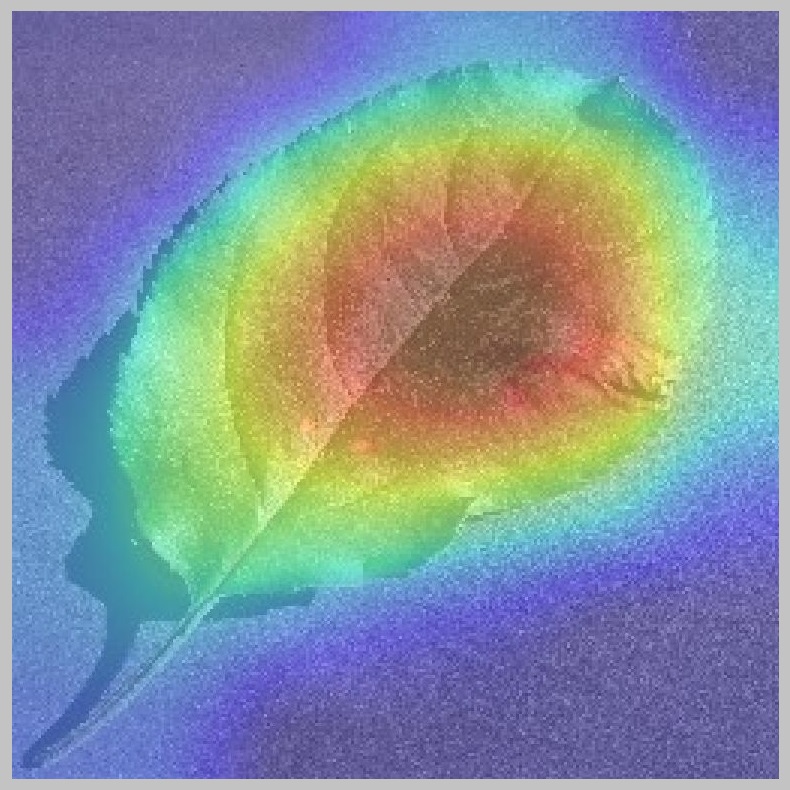} \\ 

        \rotatebox{90}{\scriptsize ~~Global+Mid} &  
        \includegraphics[width=0.13\textwidth]{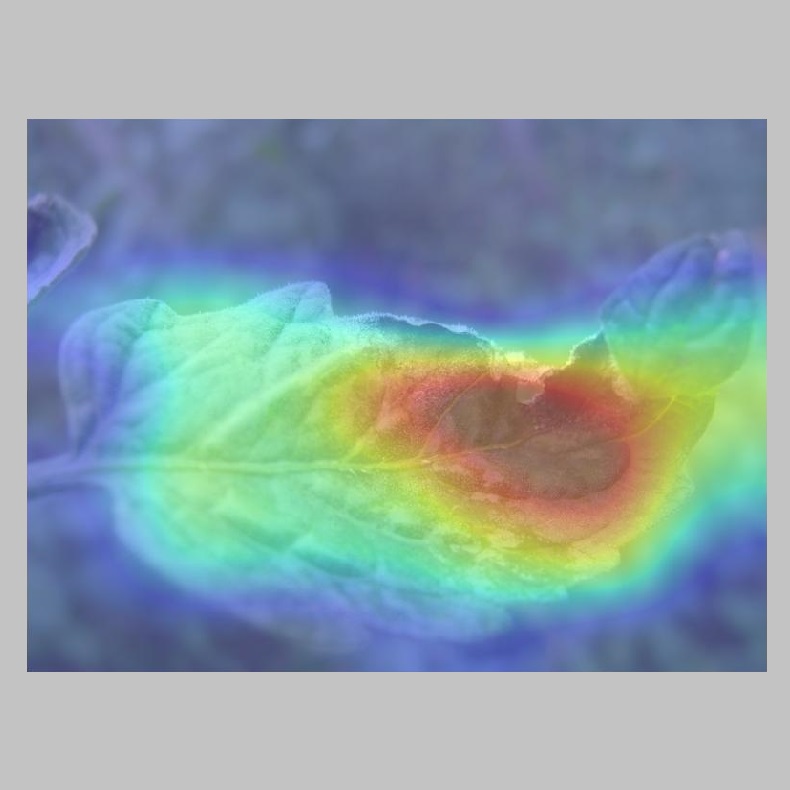} &
        \includegraphics[width=0.13\textwidth]{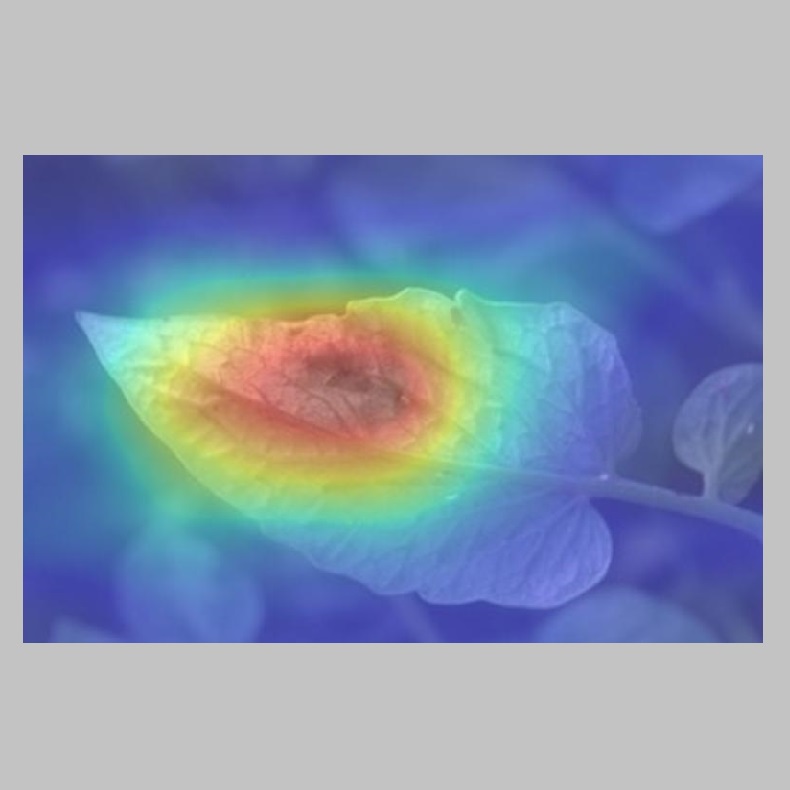} & 
        \includegraphics[width=0.13\textwidth]{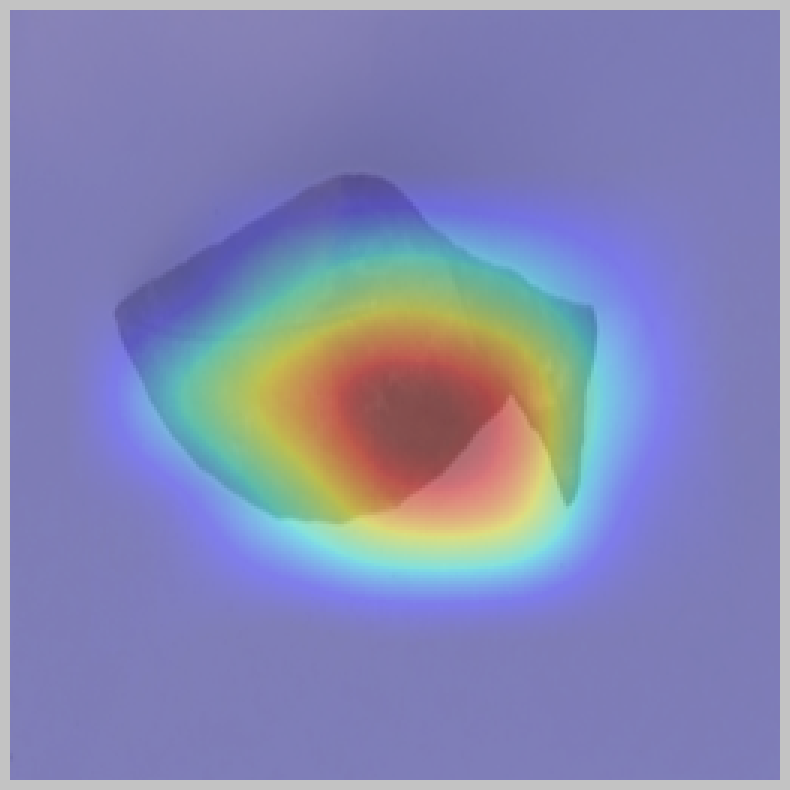} &
        \includegraphics[width=0.13\textwidth]{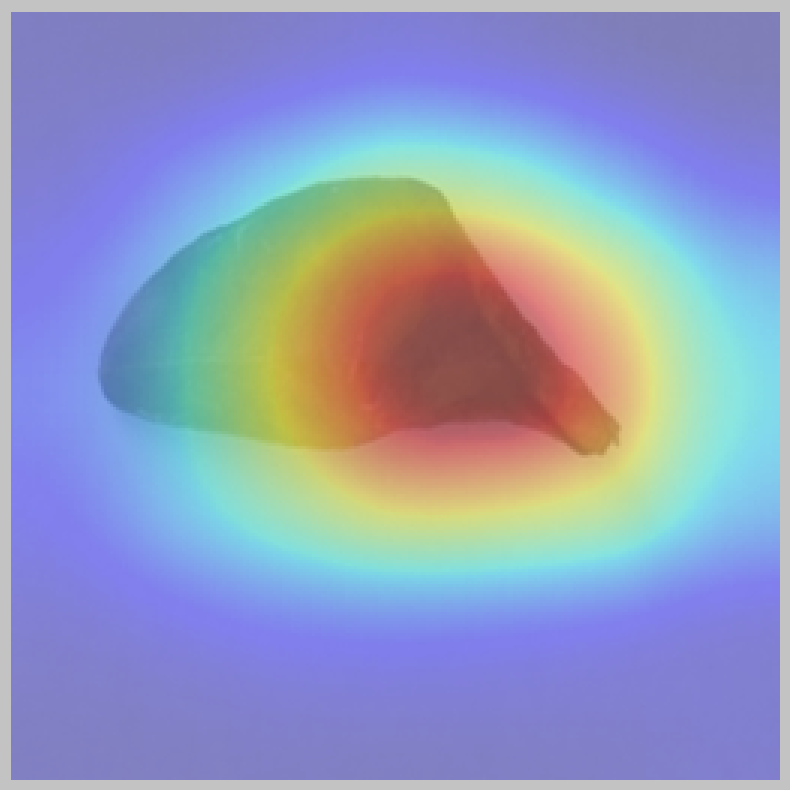} & 
        \includegraphics[width=0.13\textwidth]{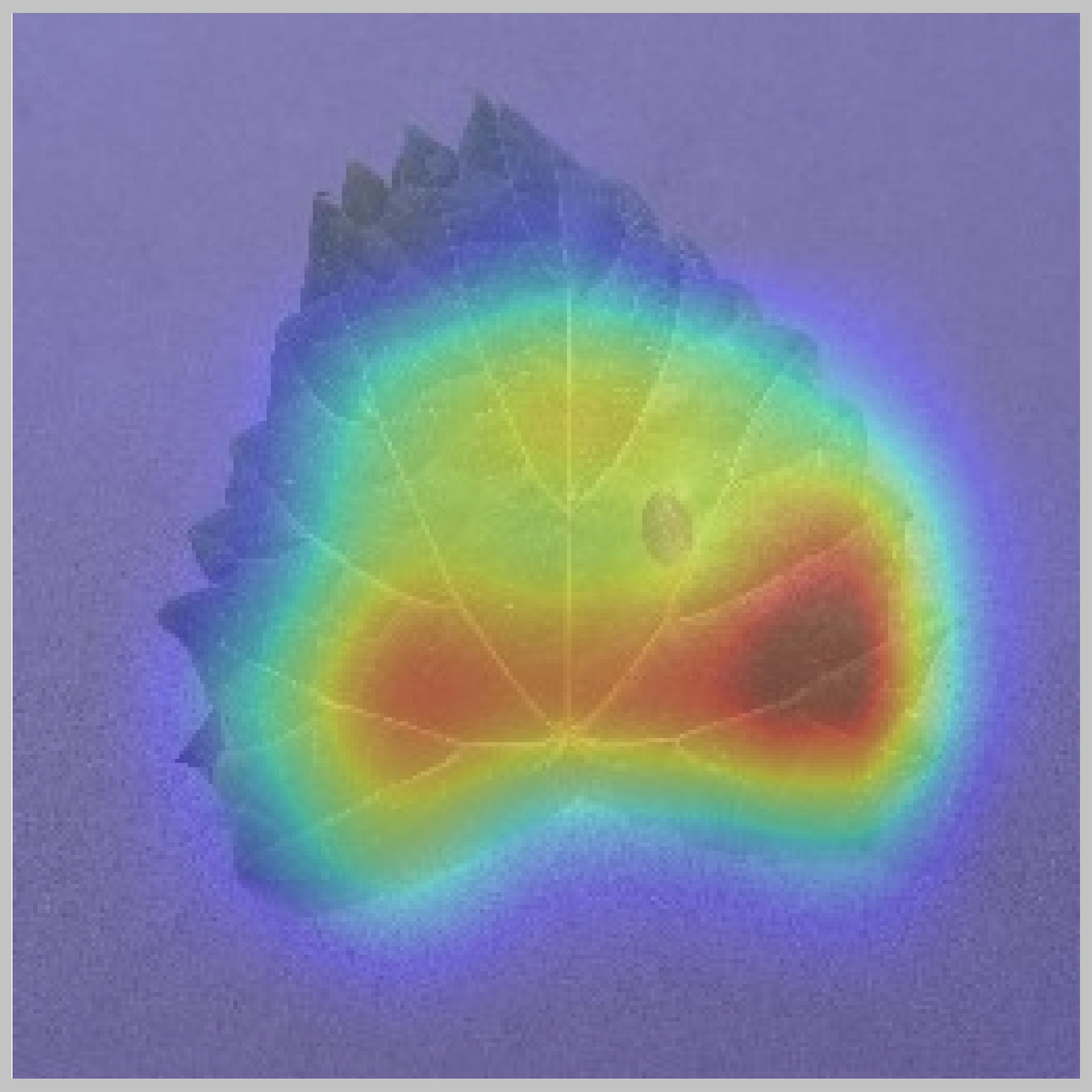} &
        \includegraphics[width=0.13\textwidth]{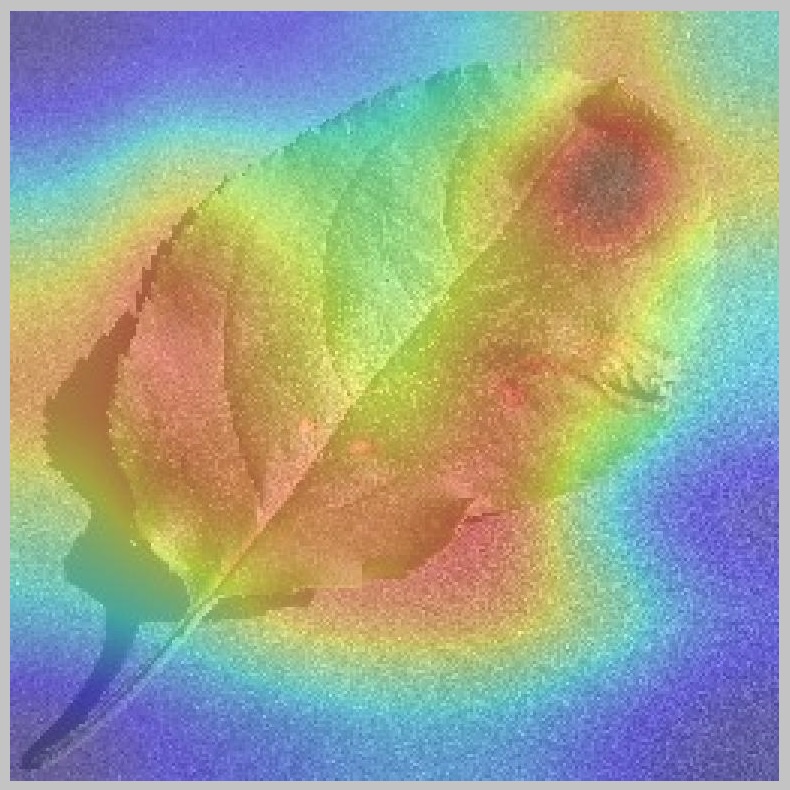} \\ 

        \rotatebox{90}{\scriptsize ~~Global+Local} &  
        \includegraphics[width=0.13\textwidth]{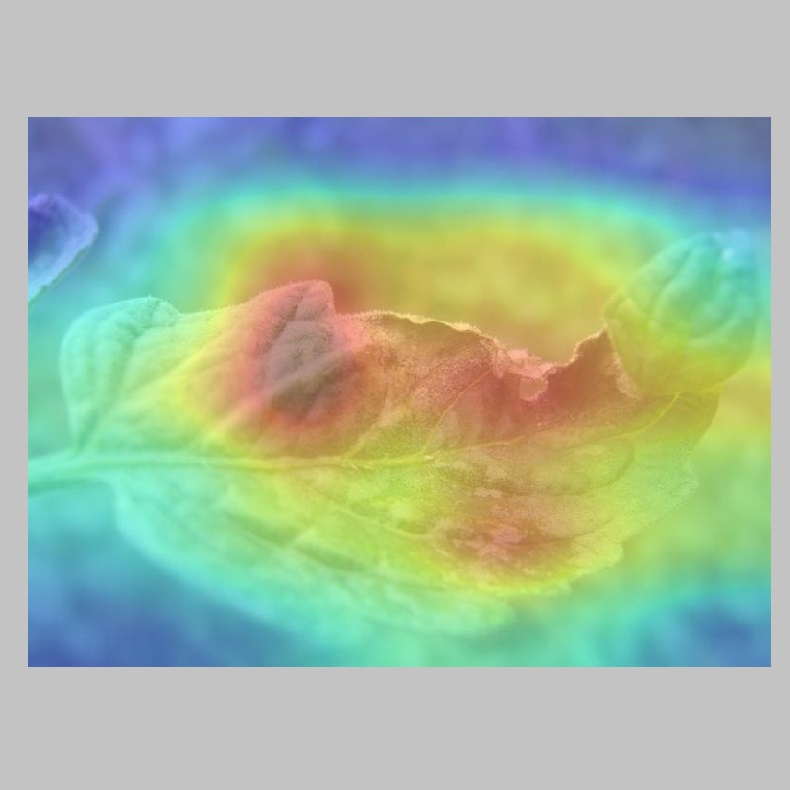} &
        \includegraphics[width=0.13\textwidth]{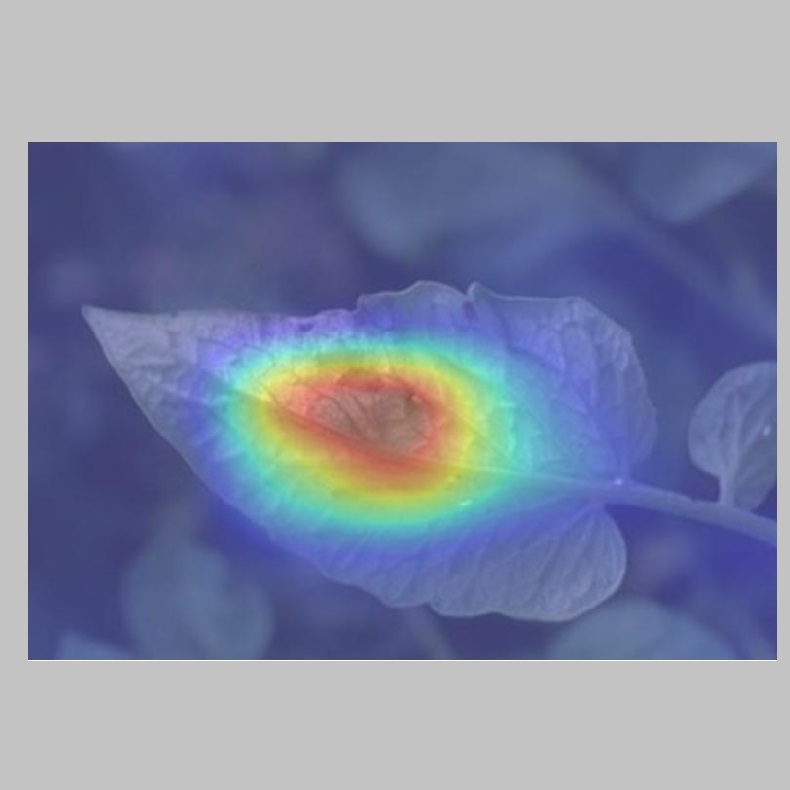} & 
        \includegraphics[width=0.13\textwidth]{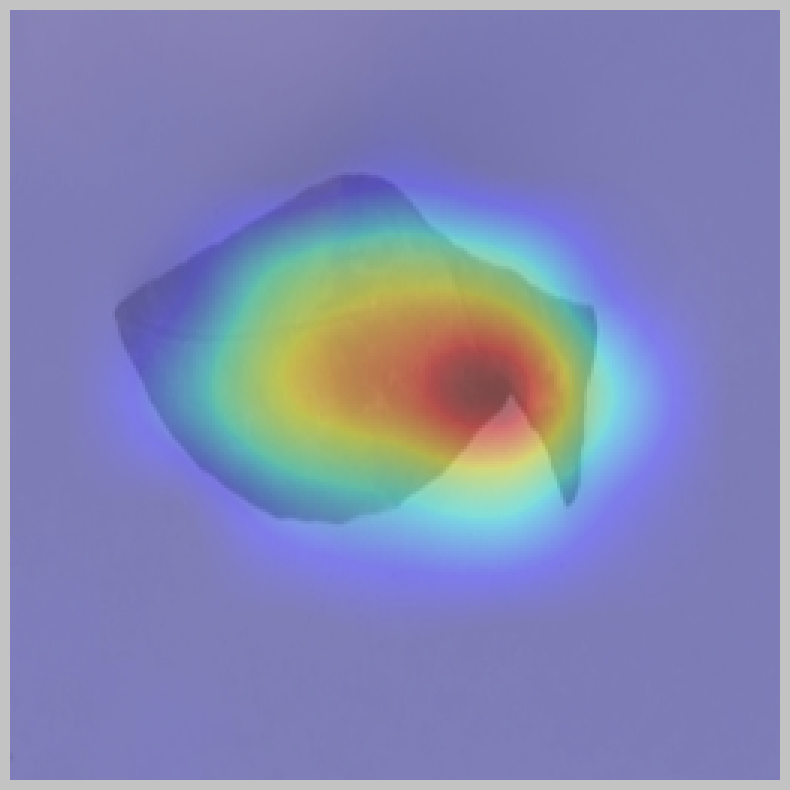} &
        \includegraphics[width=0.13\textwidth]{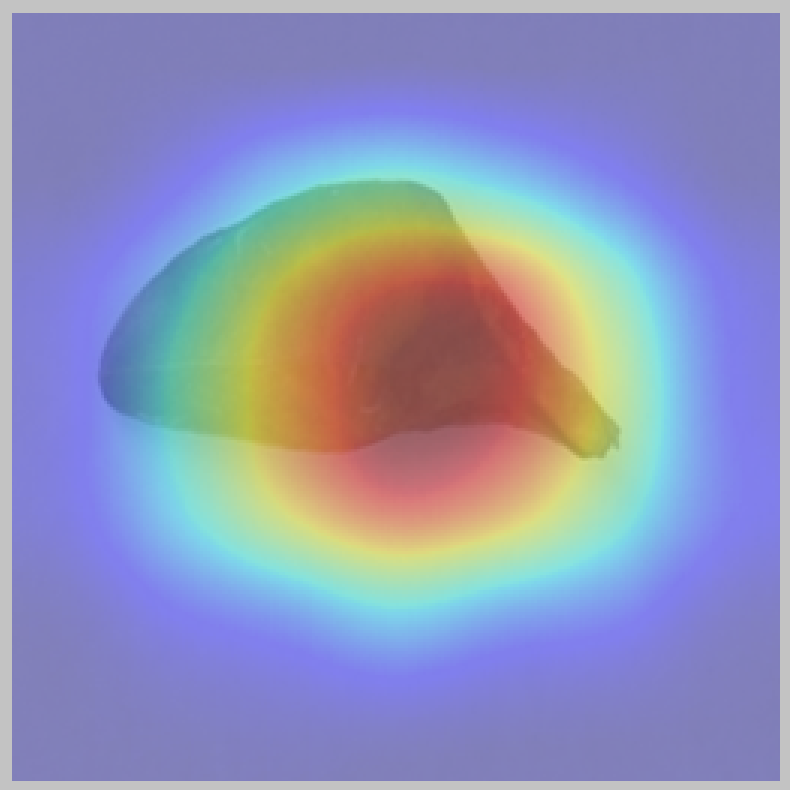} & 
        \includegraphics[width=0.13\textwidth]{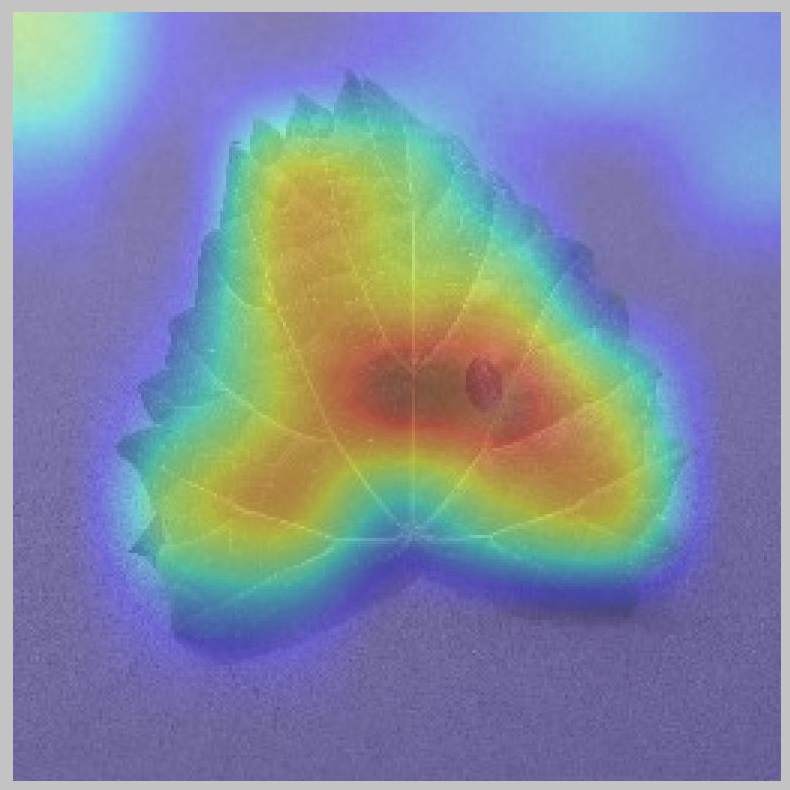} &
        \includegraphics[width=0.13\textwidth]{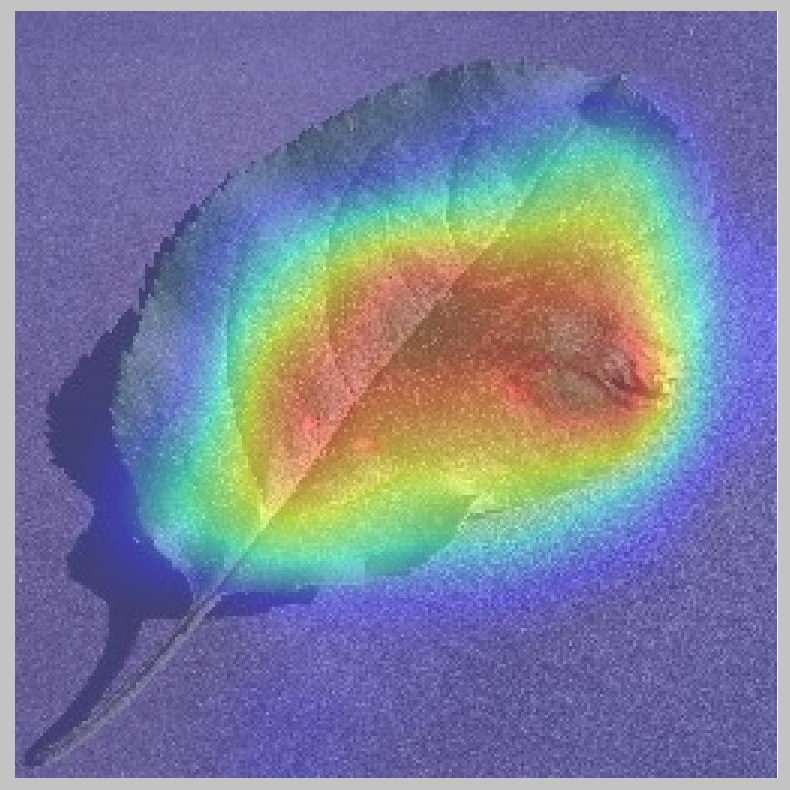} \\ 

        \rotatebox{90}{\scriptsize ~~~~PSMamba} &  
        \includegraphics[width=0.13\textwidth]{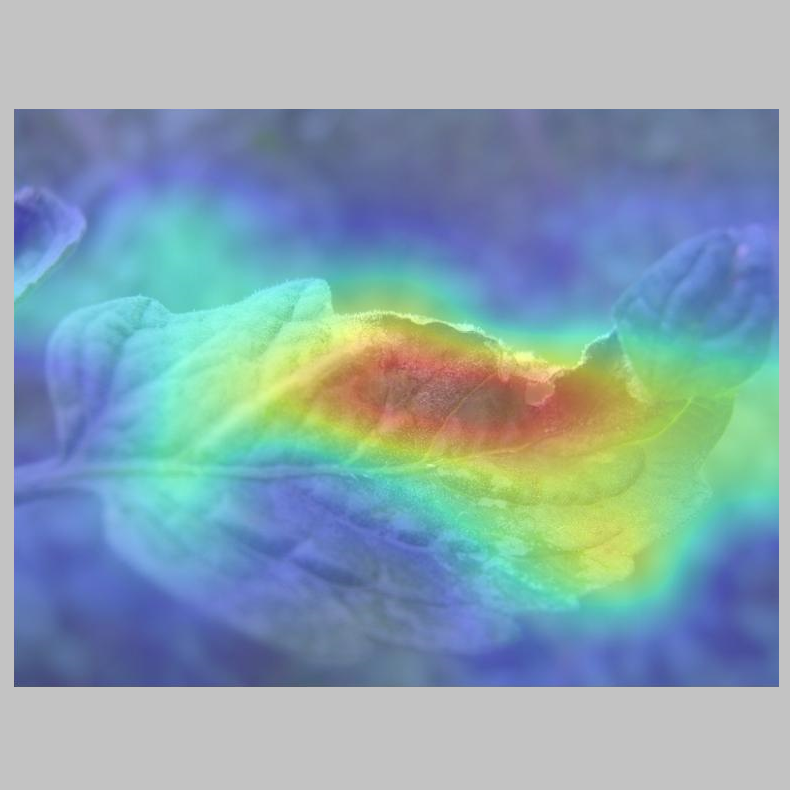} &
        \includegraphics[width=0.13\textwidth]{Figures/visual/gradcam/pd_mmdp_1.png} & 
        \includegraphics[width=0.13\textwidth]{Figures/visual/gradcam/citrus_mmdp_1.png} &
        \includegraphics[width=0.13\textwidth]{Figures/visual/gradcam/citrus_mmdp_2.png} & 
        \includegraphics[width=0.13\textwidth]{Figures/visual/gradcam/pv_mmdp_1.JPG} &
        \includegraphics[width=0.13\textwidth]{Figures/visual/gradcam/pv_mmdp_2.JPG} \\ 
    \end{tabular}
  };

  \draw[dash pattern=on 12pt off 5pt, line width=1pt] ([yshift=-2pt]img.north east) ++(-10cm,0) -- ++(0,-11.0cm);
  \draw[dash pattern=on 12pt off 5pt, line width=1pt] ([yshift=-2pt]img.north east) ++(-5cm,0) -- ++(0,-11.0cm);

\end{tikzpicture}

\vspace{0.2em}
\begin{tabular}{m{0.33\linewidth} m{0.33\linewidth} m{0.33\linewidth}} 
    \centering \scriptsize ~~~~~~~~~~~~~PlantDoc & \centering \scriptsize Citrus~~~~ & \centering \scriptsize PlantVillage~~~~~~~~~~~~~~~~~~~~
\end{tabular}
\vspace{-1.5em}
\caption{Grad-CAM visualisations of PSMamba under different multi-granular view settings (Global-only, Global+Mid, Global+Local, and Global+Mid+Local) on test datasets.}
\label{fig:gradcam_multi}
\vspace{-4mm}
\end{figure}

\subsubsection{Effectiveness of Multi-Granular Views}

To examine the effectiveness of hierarchical spatial supervision in PSMamba, we evaluate four view configurations: (i) Global only, (ii) Global + Local, (iii) Global + Mid, and (iv) the full multi-granular setting combining Global, Mid, and Local views (PSMamba). Table~\ref{tab:mmdp_ablation} summarises the resulting performance across test datasets.

The Global-only configuration provides the weakest performance across all datasets, indicating that single-scale learning fails to capture the multi-level lesion patterns characteristic of plant diseases. Adding Local views yields a substantial accuracy gain, particularly on PlantVillage and Citrus, highlighting the importance of fine-scale texture details and sharply defined lesion boundaries. In contrast, the Global + Mid setting shows the greatest relative improvement on PlantDoc. The complete multi-granular configuration achieves the highest accuracies on all benchmarks, with absolute improvements of +2.66\%, +4.28\%, and +5.45\% over the Global-only baseline on PlantVillage, PlantDoc, and Citrus, respectively. These results demonstrate that hierarchical view integration is essential for capturing both global structural context and localised lesion patterns.

Figure~\ref{fig:gradcam_multi} provides complementary interpretability evidence using Grad-CAM visualisations. Models trained with only Global views produce diffuse and often misplaced activations, reflecting a limited ability to isolate disease-relevant indicators. Introducing Mid-level views sharpens the activation patterns but still leads to unintended activation spread into healthy areas, especially in field images. Including Local views yields more precise alignment with annotated lesions, as the model learns discriminative texture and colour variations. The full multi-granular configuration produces the most coherent and stable localisation across datasets, capturing broader structural patterns and fine-grained lesion details. This consistency between quantitative outcomes and visual interpretability confirms that multi-scale supervision substantially enhances the discriminative capacity and reliability of PSMamba under both controlled and real-field conditions.

\begin{table}[!t]
\centering
\scriptsize
\caption{Effect of different positional embedding strategies on the performance of PSMamba. Accuracy (\%) is reported as the evaluation metric.}
\label{tab:mpa_ablation}
\begin{tblr}{
  cells = {c},
  hline{1,5} = {-}{0.08em},
  hline{2} = {-}{0.05em},
}
\textbf{Method}     & \textbf{PlantVillage} & \textbf{PlantDoc} & \textbf{Citrus} \\
Shared Positional Embedding           & 97.91                 & 94.25             & 90.74           \\
Per-scale Positional Embedding        & 98.12                 & 95.12             & 91.46           \\
\textbf{Multi-scale Positional Alignment} & \textbf{98.46}        & \textbf{95.58}    & \textbf{92.15}  
\end{tblr}
\end{table}

\subsubsection{Effect of Multi-scale Positional Alignment (MPA)}
The impact of positional encoding strategies on representation quality is evaluated in Table \ref{tab:mpa_ablation}. The results reveal clear and consistent performance differences among shared embeddings, per-scale embeddings, and the proposed MPA module. Across all test datasets, MPA delivers the strongest accuracy, highlighting the importance of aligning tokens across heterogeneous spatial resolutions within the multi-granular PSMamba framework.

Shared positional embeddings yield the weakest performance, as forcing all views to share a single embedding table suppresses scale-specific spatial cues that are essential for recognising lesion patterns. This limitation becomes particularly evident on PlantDoc and Citrus, where variations in illumination, viewpoint, and lesion subtlety demand more precise spatial correspondence across crops. Per-scale embeddings mitigate this issue by assigning separate positional tables to each view and thus better preserving resolution-specific priors. However, because these embeddings are learned independently, Global, Mid, and Local tokens still lack an explicit mechanism for cross-scale alignment, producing inconsistencies during hierarchical distillation.

The proposed MPA achieves the highest accuracy on all benchmarks (98.46 \%, 95.58 \%, and 92.15 \% on PlantVillage, PlantDoc, and Citrus, respectively), demonstrating the benefit of explicitly aligning tokens between crop resolutions. By combining geometry-aware landing positions with learnable positional adjustments and neighbourhood projection, MPA ensures that Mid and Local tokens land coherently on the Global grid. This stabilises cross-scale distillation, reduces positional mismatch, and enables the model to propagate lesion-relevant structure across hierarchical views. The performance gains over per-scale embeddings are especially prominent on PlantDoc (+0.46 \%) and Citrus (+0.69 \%), confirming that MPA is particularly valuable in challenging, variable, and fine-grained environments.

Overall, these results show that effective positional alignment is critical for multi-granular representation learning. MPA allows PSMamba to maintain semantic consistency across Global, Mid, and Local views, directly contributing to its improvements in accuracy, robustness, and lesion localisation.

\begin{table}[!t]
\centering
\scriptsize
\caption{Performance of PSMamba and second best methods, ConMamba \cite{mamun2025conmamba} on test datasets under data imbalance settings.}
\label{tab:Class Imbalance}
\begin{tblr}{
  cells = {c},
  cell{2}{1} = {c=5}{},
  cell{3}{1} = {r=2}{},
  cell{5}{1} = {r=2}{},
  cell{7}{1} = {c=5}{},
  cell{8}{1} = {r=2}{},
  cell{10}{1} = {r=2}{},
  cell{12}{1} = {c=5}{},
  cell{13}{1} = {r=2}{},
  cell{15}{1} = {r=2}{},
  hline{1-3,7-8,12-13,17} = {-}{},
  hline{5,10,15} = {2-5}{},
}
\textbf{Model}   & \textbf{Sampling} & \textbf{Accuracy} & {\textbf{Macro}\\\textbf{F1}} & {\textbf{Minority}\\\textbf{F1}} \\
PlantVillage     &                   &                   &                               &                                  \\
ConMamba \cite{mamun2025conmamba} & Random            & 98.63            & 97.38             & 96.12               \\
                 & Balanced          & 98.71             & 97.91             & 97.23                \\
\textbf{PSMamba} & Random            & 98.87             & 97.69            & 96.43               \\
                 & Balanced          & \textbf{98.97}    & \textbf{98.21}    & \textbf{97.54}       \\
PlantDoc         &                   &                   &                               &                                  \\
ConMamba \cite{mamun2025conmamba} & Random            & 94.29             & 91.23             & 89.15                \\
                 & Balanced          & 94.56             & 92.84             & 91.71                \\
\textbf{PSMamba} & Random            & 95.64             &  94.11            & 91.14                \\
                 & Balanced          & \textbf{95.91}    & \textbf{94.73}    & \textbf{93.69}       \\
Citrus           &                   &                   &                               &                                  \\
ConMamba \cite{mamun2025conmamba} & Random            & 91.38             & 91.66             & 89.42                \\
                 & Balanced          & 91.80             & 92.17             & 91.01                \\
\textbf{PSMamba} & Random            & 92.26             &  91.71            & 90.55                \\
                 & Balanced          & \textbf{92.68}    & \textbf{92.22}    & \textbf{92.14}       
\end{tblr}

\end{table}

\subsubsection{PSMamba for Class Imbalance}
Class imbalance is a persistent challenge in plant disease datasets, where certain diseases naturally occur less frequently and exhibit subtle symptoms that are easily overshadowed by dominant categories. Imbalanced data can distort representation learning by biasing models toward majority classes, reducing inter-class margins, and causing minority-class features to collapse into the embedding subspace occupied by dominant diseases. Table \ref{tab:Class Imbalance} evaluates the sensitivity of PSMamba to these issues by comparing its performance with the second-best method, ConMamba \cite{mamun2025conmamba} under both random (imbalanced) and balanced sampling strategies across the test datasets.

On the PlantVillage dataset, PSMamba achieves strong performance under both sampling strategies, but the improvements under balanced sampling reflect its superior ability to lift minority-class performance without sacrificing overall accuracy. When trained with randomly sampled data, PSMamba already surpasses ConMamba \cite{mamun2025conmamba} in accuracy, macro F1, and minority-class F1, demonstrating inherent robustness to imbalance. Under balanced sampling, PSMamba further improves to an accuracy of 98.97\%, a macro F1 of 98.21\%, and a minority-class F1 of 97.54\%. These gains indicate that multi-scale representation learning helps preserve discriminative details for underrepresented diseases. The Local-student pathway, which captures fine-grained lesion textures, and the Mid-student pathway, which models vein–lesion interactions, play complementary roles in counteracting the collapse of minority-class features that typically occurs in single-stream SSL models.

The benefits of PSMamba become more pronounced in the PlantDoc dataset, where class imbalance coincides with real-field variability. Under random sampling, PSMamba yields higher accuracy and macro F1 than ConMamba \cite{mamun2025conmamba} and achieves a minority-class F1 of 91.14\%, showing stronger resilience to distortion caused by uneven class frequencies. When balanced sampling is applied, PSMamba reaches a minority-class F1 of 93.69\%, outperforming ConMamba \cite{mamun2025conmamba} by a significant margin. These results highlight the value of the MPA module, which ensures that student representations are aligned to a consistent global token grid. Minority-class samples, which often contain incomplete or noisy lesion patterns in PlantDoc, benefit from this alignment because supervision becomes geometrically stable across views, reducing the risk of misclassification into visually dominant classes. 
The Citrus dataset provides the most stringent test of imbalance robustness due to its highly fine-grained nature. Under random sampling, PSMamba achieves higher accuracy and minority-class performance than ConMamba \cite{mamun2025conmamba}, despite the challenging symptom similarities. When balanced sampling is applied, PSMamba reaches a minority-class F1 of 92.14 \%, demonstrating the highest minority sensitivity in all experiments. This improvement reflects the complementarity between local-scale lesion encoding and mid-level structural modelling. Fine-grained diseases require precise characterisation of micro-patterns, and PSMamba’s hierarchical supervision ensures that minority-class examples contribute meaningfully to the learned feature space rather than being absorbed by majority-class attractor regions.

Taken together, these results illustrate that PSMamba not only improves overall recognition accuracy but also significantly strengthens representation quality for underrepresented disease categories. Its multi-granular architectural design mitigates imbalance-induced feature collapse by preserving fine-grained lesion cues, stabilising mid-scale spatial relationships, and aligning multi-resolution views through consistent geometric mapping. Unlike ConMamba, which relies primarily on a single-scale recurrence backbone, PSMamba’s hierarchical distillation distributes supervisory signals across scale-specific pathways, enabling minority classes to maintain distinct and well-formed embedding clusters.

\begin{figure}[!t]
    \centering
    \begin{subfigure}{0.25\textwidth}
        \centering
        \includegraphics[width=\linewidth]{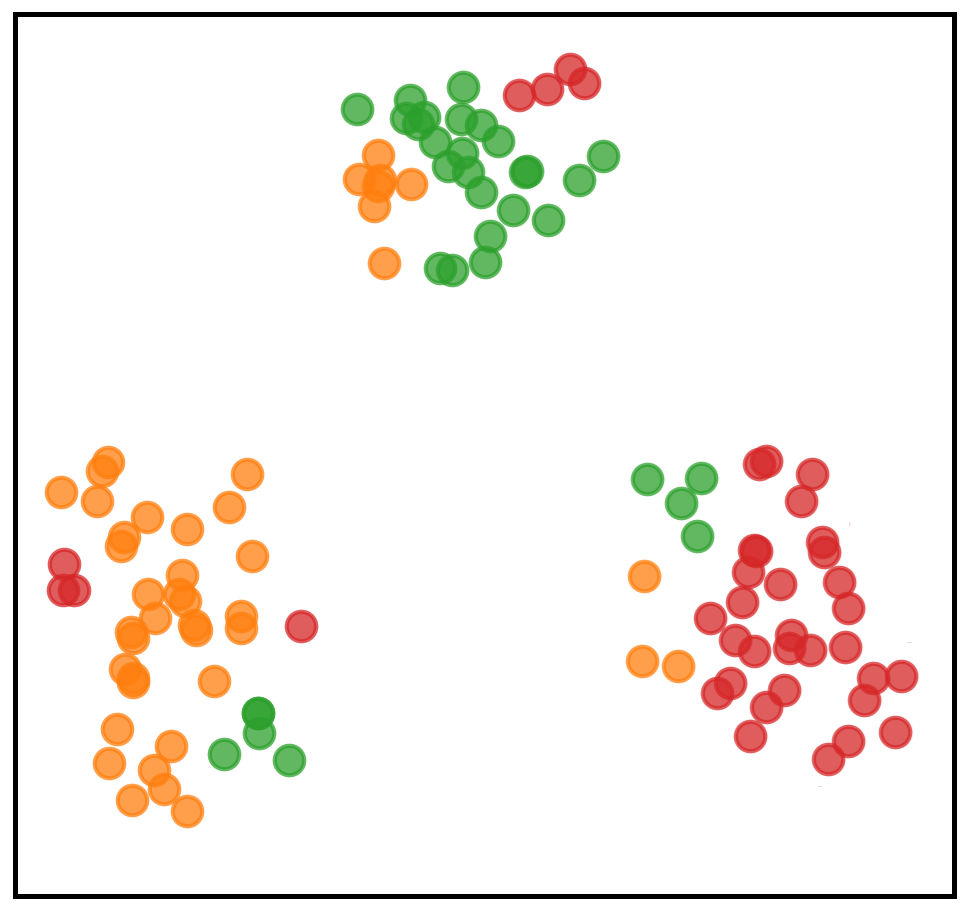}
        \caption{ConMamba \cite{mamun2025conmamba}}
        \label{fig:a}
    \end{subfigure}
    \hspace{1.5em} 
    \begin{subfigure}{0.25\textwidth}
        \centering
        \includegraphics[width=\linewidth]{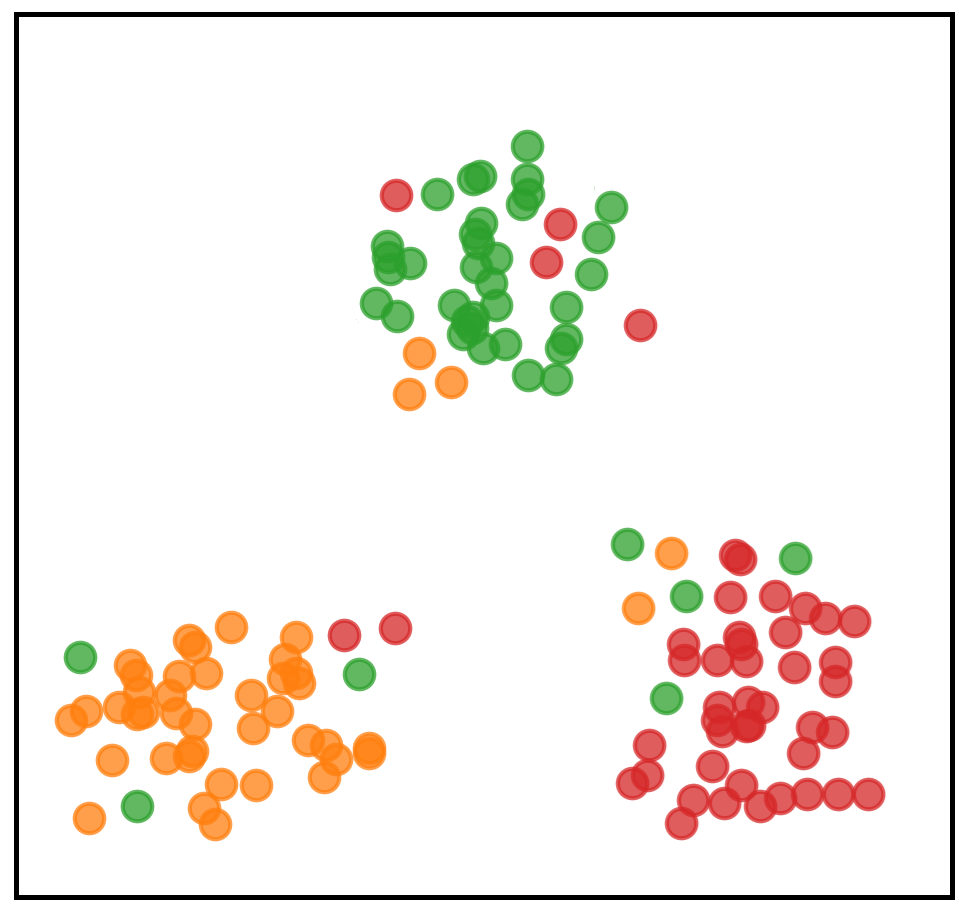}
        \caption{Global + Mid view}
        \label{fig:b}
    \end{subfigure}

    \vspace{0.1cm}

    \begin{subfigure}{0.25\textwidth}
        \centering
        \includegraphics[width=\linewidth]{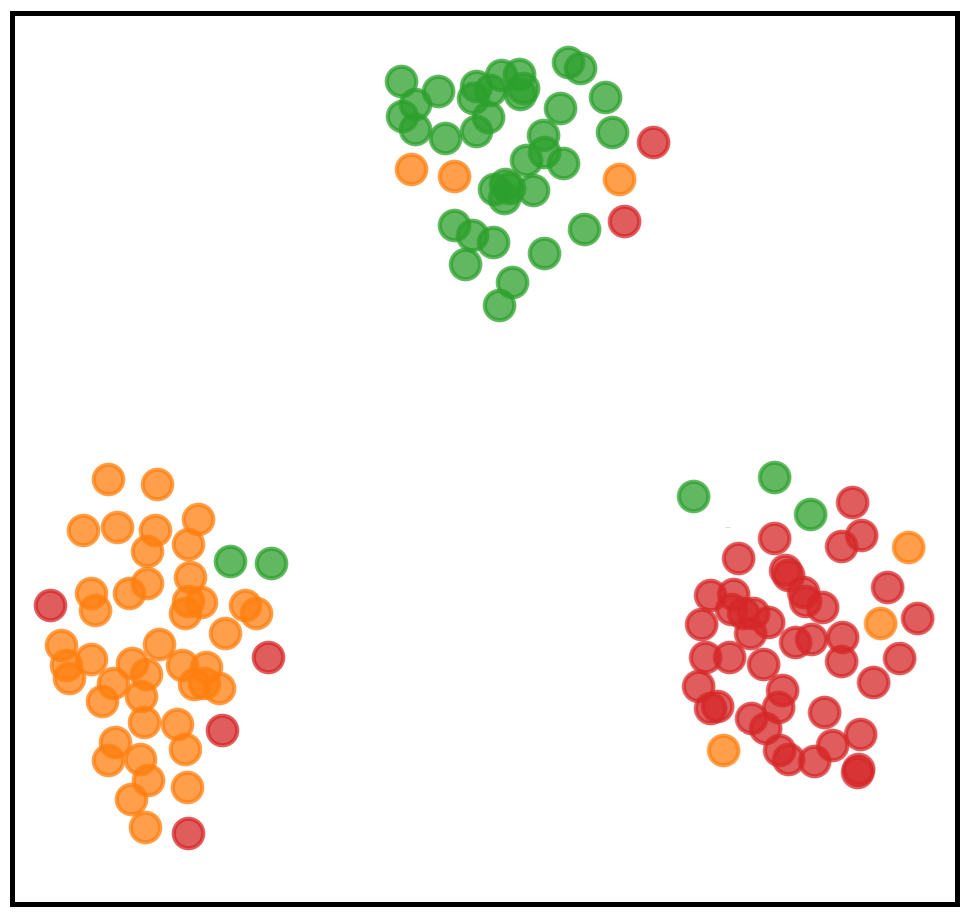}
        \caption{Global + Local view}
        \label{fig:c}
    \end{subfigure}
    \hspace{1.5em}
    \begin{subfigure}{0.25\textwidth}
        \centering
        \includegraphics[width=\linewidth]{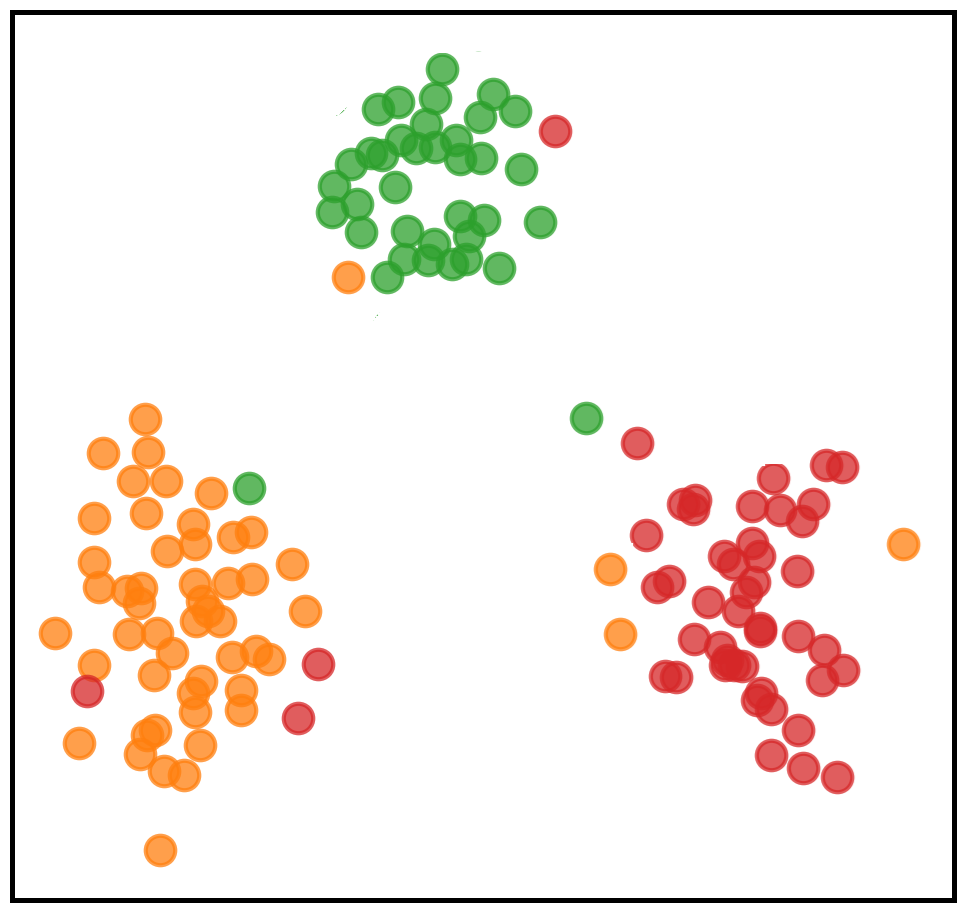}
        \caption{PSMamba}
        \label{fig:d}
    \end{subfigure}

    \caption{t-SNE visualisations of learned feature representations for the Apple plant of the \textbf{PlantDoc} dataset across four configurations: (a) Second best method, ConMamba \cite{mamun2025conmamba}, (b) Global + Mid view, (c) Global + Local view, and (d) the proposed PSMamba.}
    \label{fig:tsne_pd}
\end{figure}

\subsubsection{Feature Visualisations using t-SNE}

\begin{figure}[!t]
    \centering
    \begin{subfigure}{0.25\textwidth}
        \centering
        \includegraphics[width=\linewidth]{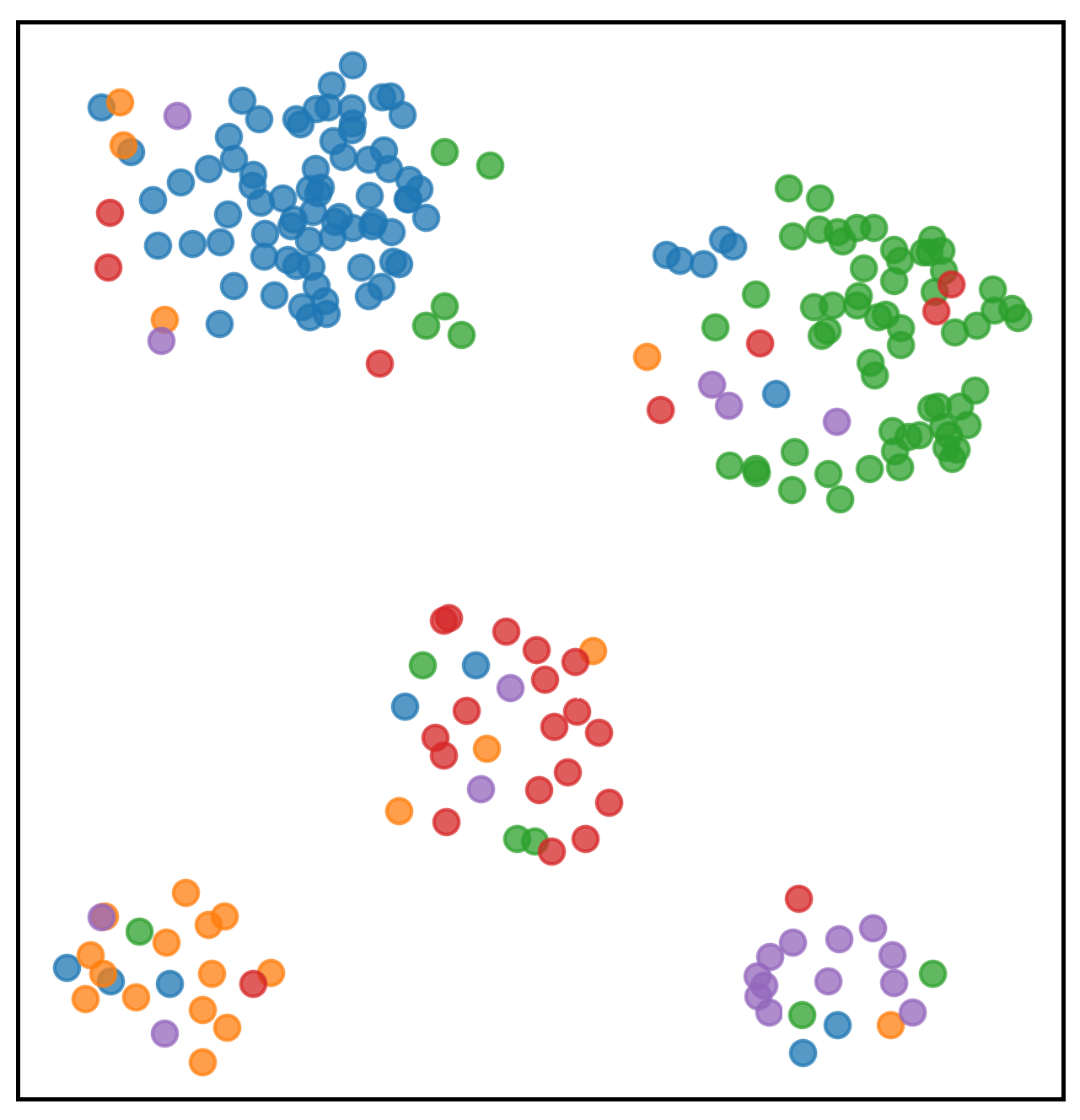}
        \caption{ConMamba \cite{mamun2025conmamba}}
        \label{fig:a}
    \end{subfigure}
    \hspace{1.5em} 
    \begin{subfigure}{0.25\textwidth}
        \centering
        \includegraphics[width=\linewidth]{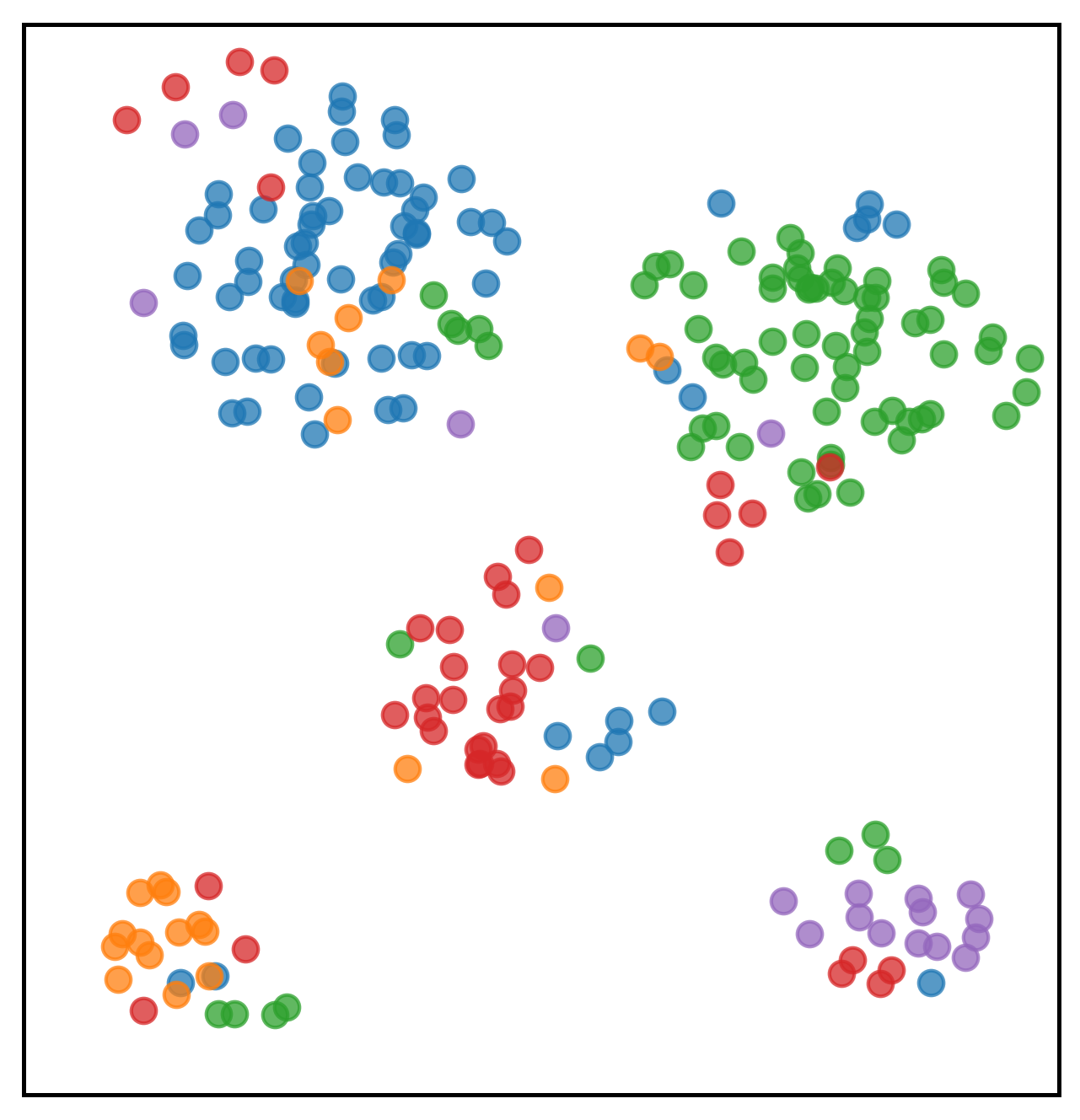}
        \caption{Global + Mid view}
        \label{fig:b}
    \end{subfigure}

    \vspace{0.1cm}

    \begin{subfigure}{0.25\textwidth}
        \centering
        \includegraphics[width=\linewidth]{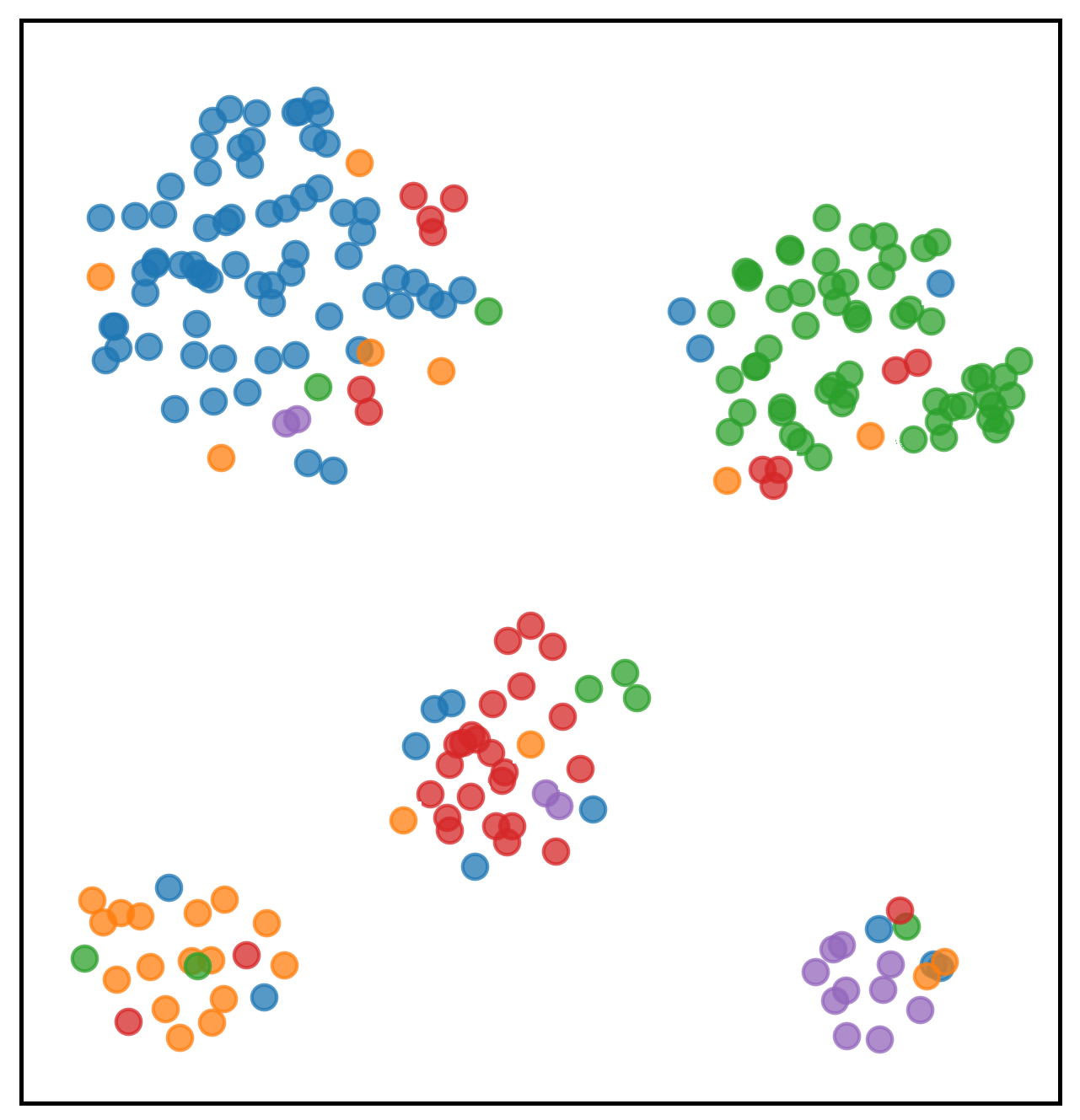}
        \caption{Global + Local view}
        \label{fig:c}
    \end{subfigure}
    \hspace{1.5em} 
    \begin{subfigure}{0.25\textwidth}
        \centering
        \includegraphics[width=\linewidth]{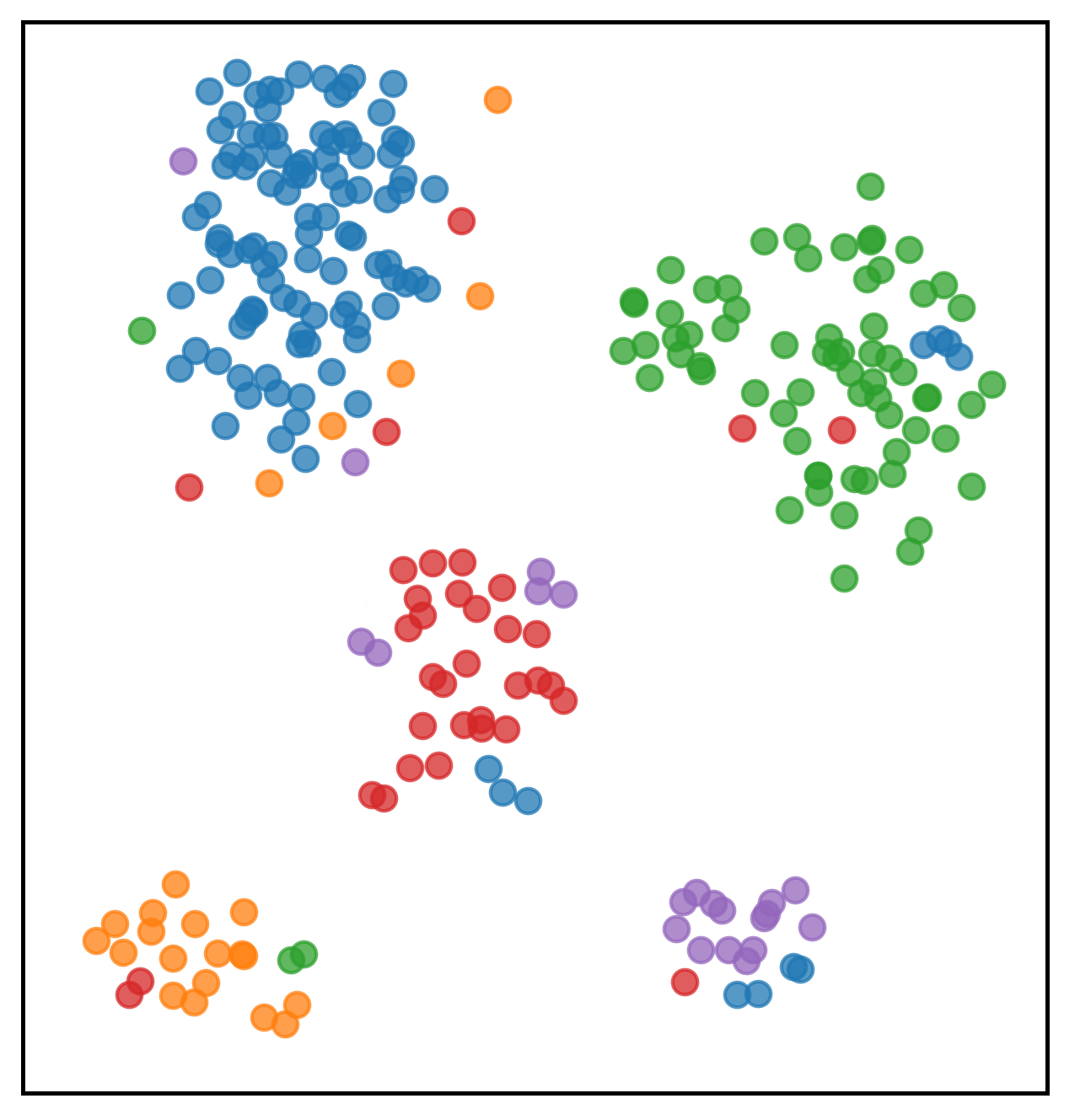}
        \caption{PSMamba}
        \label{fig:d}
    \end{subfigure}

    \caption{t-SNE visualisations of learned feature representations for the \textbf{Citrus dataset} across four configurations: (a) Second best method, ConMamba \cite{mamun2025conmamba}, (b) Global + Mid view, (c) Global + Local view, and (d) the proposed PSMamba.}
    \label{fig:tsne_citrus}

\end{figure}

\begin{figure}[t]
    \centering
    \begin{subfigure}{0.25\textwidth}
        \centering
        \includegraphics[width=\linewidth]{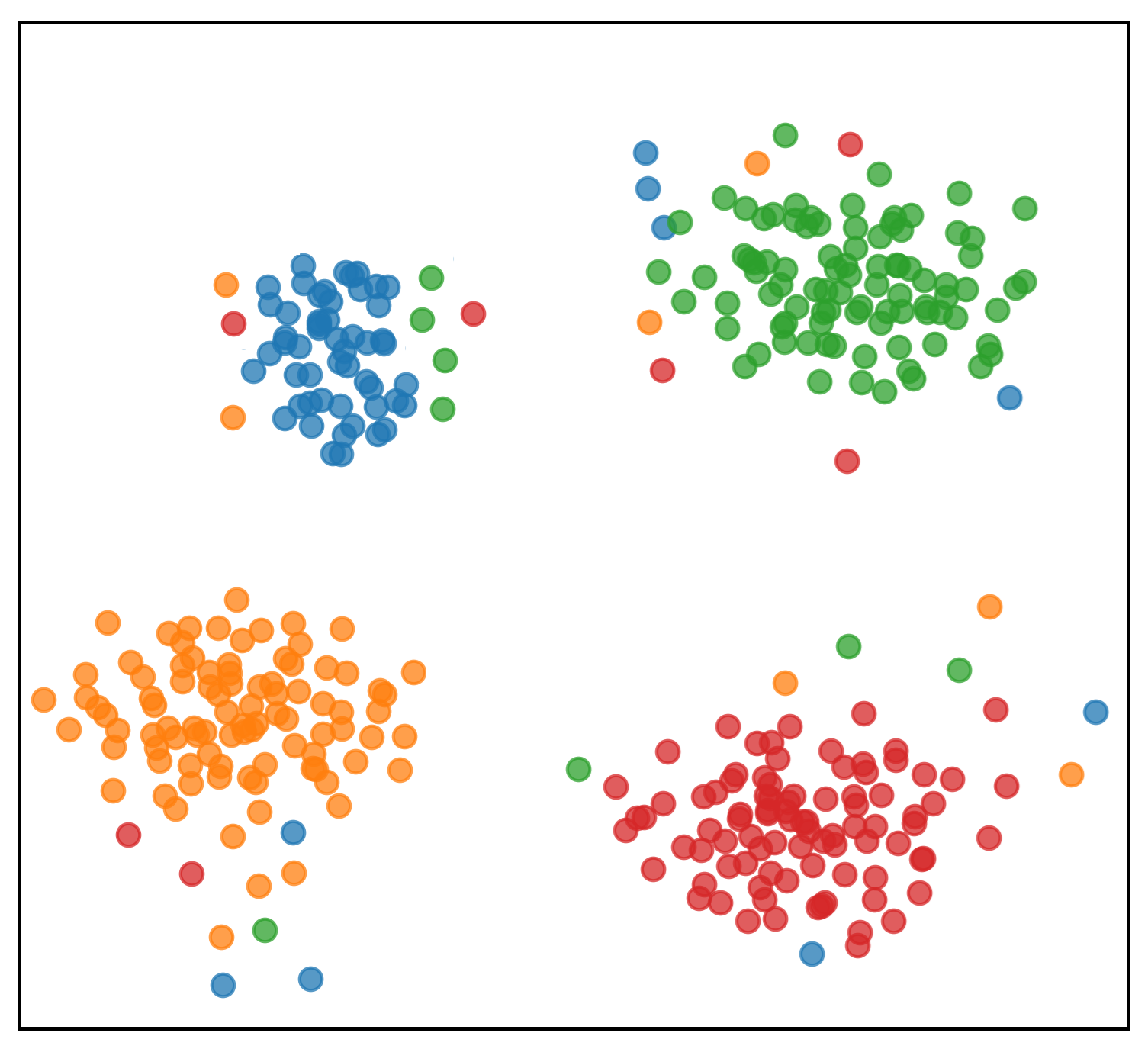}
        \caption{ConMamba \cite{mamun2025conmamba}}
        \label{fig:a}
    \end{subfigure}
    \hspace{1.5em} 
    \begin{subfigure}{0.25\textwidth}
        \centering
        \includegraphics[width=\linewidth]{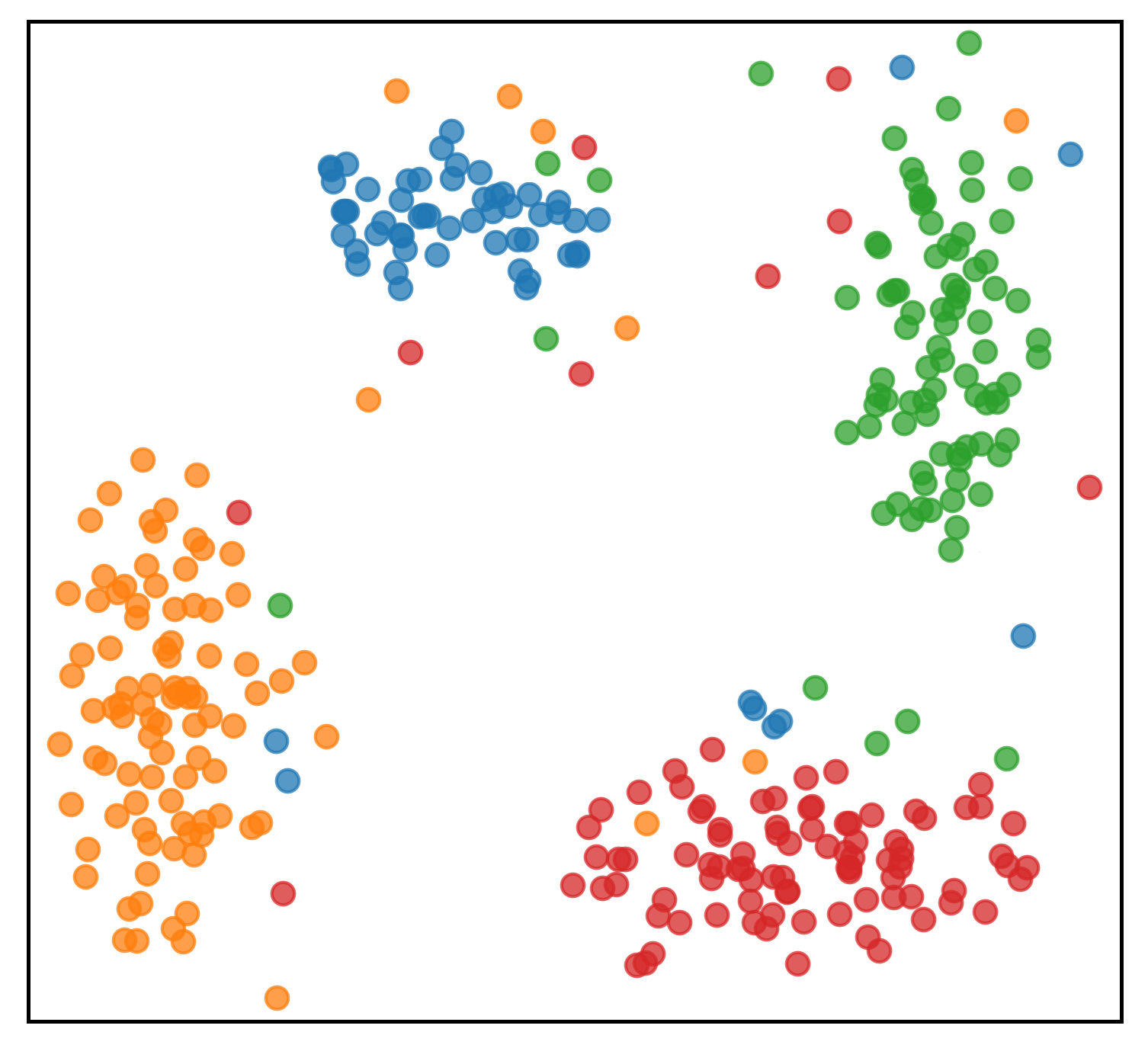}
        \caption{Global + Mid view}
        \label{fig:b}
    \end{subfigure}

    \vspace{0.1cm}

    \begin{subfigure}{0.25\textwidth}
        \centering
        \includegraphics[width=\linewidth]{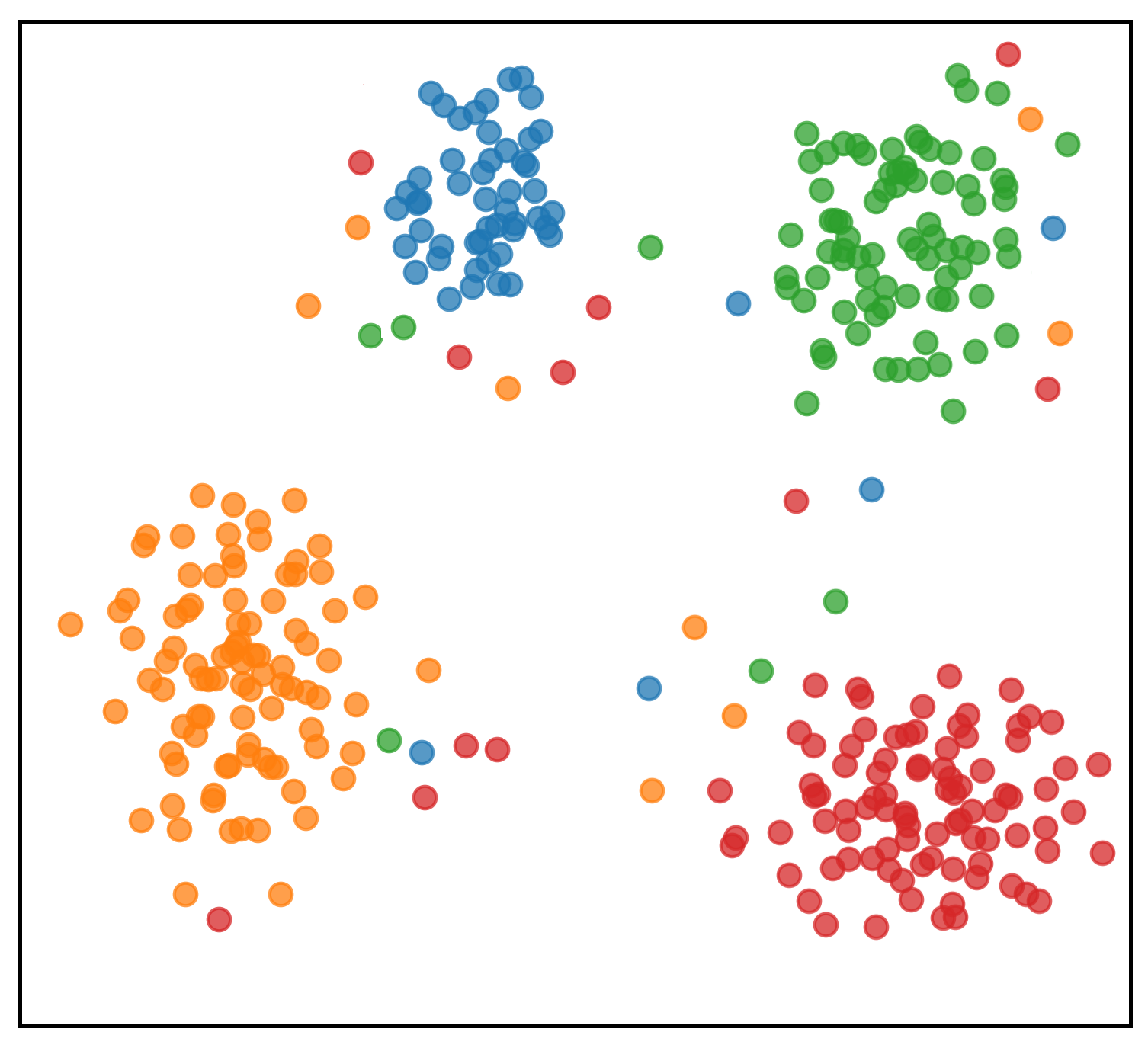}
        \caption{Global + Local view}
        \label{fig:c}
    \end{subfigure}
    \hspace{1.5em} 
    \begin{subfigure}{0.25\textwidth}
        \centering
        \includegraphics[width=\linewidth]{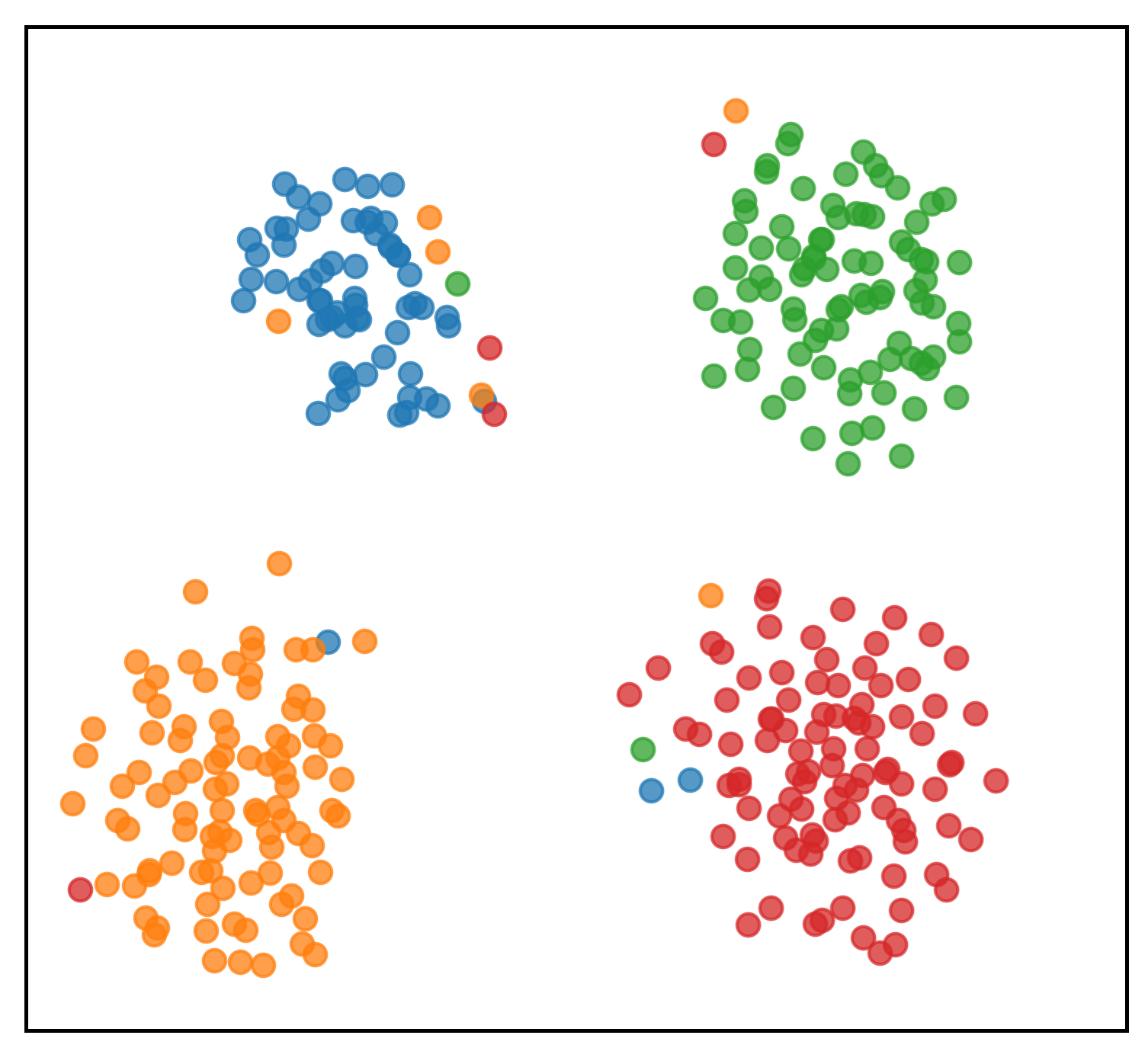}
        \caption{PSMamba}
        \label{fig:d}
    \end{subfigure}

    \caption{t-SNE visualisations of learned feature representations for the Apple plant of the \textbf{PlantVillage} dataset across four configurations: (a) Second best method, ConMamba \cite{mamun2025conmamba}, (b) Global + Mid view, (c) Global + Local view, and (d) the proposed PSMamba.}
    \label{fig:tsne_pv}
\end{figure}

To further examine the discriminative structure of the learned representation space, we employ t-SNE to project high-dimensional token embeddings onto a two-dimensional manifold. Figures \ref{fig:tsne_pd}--\ref{fig:tsne_pv} compare feature distributions from the second-best methods, ConMamba \cite{mamun2025conmamba}, the Global + Mid configuration, the Global + Local configuration, and the proposed PSMamba across the test datasets. These visualisations allow us to assess compactness, separation, and inter-class boundaries produced by each approach and provide qualitative evidence for the effectiveness of multi-granular supervision in shaping more structured latent spaces.

Across the PlantDoc dataset (Figure \ref{fig:tsne_pd}), PSMamba (Figure \ref{fig:tsne_pd} (d)) generates clusters that are noticeably more compact and clearly separated than those produced by ConMamba \cite{mamun2025conmamba} and the partial-view configurations. The Global + Mid (Figure \ref{fig:tsne_pd} (b)) configuration shows moderate improvements, with classes beginning to form more coherent groups but still exhibiting noticeable intra-class spread. The Global + Local (Figure \ref{fig:tsne_pd} (c)) configuration improves further by emphasising fine-grained features; however, it remains insufficient to resolve the domain-shift challenges inherent to field imagery. 

A similar progression is observed for the Citrus dataset, where class distinctions are subtle and rely heavily on micro-level texture variations. The ConMamba \cite{mamun2025conmamba} (Figure \ref{fig:tsne_citrus} (a)) embedding space shows drift and partial mixing among classes. The Global + Mid (Figure \ref{fig:tsne_citrus} (b)) configuration provides slightly improved grouping, reflecting the contribution of vein-level context. The Global + Local (Figure \ref{fig:tsne_citrus} (c)) configuration captures richer micro-patterns, producing clusters that are more coherent but still partially entangled. PSMamba (Figure \ref{fig:tsne_citrus} (d)), however, demonstrates marked improvement, with clearly separated clusters for all citrus disease categories. 

The results on the PlantVillage dataset further validate the robustness of the proposed PSMamba framework. ConMamba \cite{mamun2025conmamba} (Figure \ref{fig:tsne_pv} (a)) produces reasonably separated clusters but still displays mild spreading within each class. Both the Global + Mid (Figure \ref{fig:tsne_pv} (b)) and Global + Local (Figure \ref{fig:tsne_pv} (c)) configurations improve this behaviour, demonstrating the individual contributions of intermediate and high-resolution cues. Yet neither configuration alone completely resolves subtle cross-class ambiguity. PSMamba (Figure \ref{fig:tsne_pv} (d)) achieves the clearest and most compact cluster formations across all classes, characterised by dense intra-class regions and large inter-cluster margins.

Collectively, the t-SNE analyses reveal that PSMamba consistently produces the most structured and discriminative feature distributions across all datasets. The compact clusters and clearly defined boundaries reflect the effectiveness of aligning multi-granular views and distilling complementary information across local, mid-scale, and global perspectives. These visual insights reinforce the quantitative performance improvements and emphasise that PSMamba not only achieves higher accuracy but also learns a more interpretable, stable, and disease-aware latent space. 

\subsection{Limitations}

While PSMamba is evaluated comprehensively across three complementary datasets, the scope of this study remains focused on image-level disease classification under standardised experimental settings. The framework has not yet been assessed on additional plant health analysis tasks such as lesion localisation or segmentation, which may require task-specific adaptations of the framework. Furthermore, the multi-granular crop configuration used in this work is fixed, and exploring adaptive or dataset-aware scale choices could offer additional flexibility. These considerations define the boundaries of the current study and highlight opportunities for extending PSMamba to a broader range of plant-disease analysis scenarios.

\section{Conclusions}


This work presented PSMamba, a progressive SSL framework designed to address the hierarchical and multi-scale nature of plant disease symptoms. Unlike prior SSL approaches that emphasise global alignment or rely on single-view supervision, PSMamba integrates a bidirectional VM encoder for efficient long-range and fine-grained modelling, a dual-student distillation strategy that specialises in Mid- and Local-scale lesion cues, and an MPA module that ensures coherent token correspondence across heterogeneous spatial resolutions. Through this combination, the framework jointly captures broad structural patterns, vein-level interactions, and subtle texture irregularities that support fine-grained disease recognition. Comprehensive experiments on PlantVillage, PlantDoc, and Citrus demonstrate that PSMamba consistently surpasses state-of-the-art SSL baselines in cross-domain robustness and minority-class sensitivity under class imbalance. Qualitative evaluations further show that PSMamba yields more compact and visually coherent feature distributions, as well as sharper lesion-focused activation responses, validating the effectiveness of its multi-granular learning strategy.
Future work will explore extending the progressive alignment and distillation mechanisms to other agricultural modalities, improving scalability for large-scale field deployments, and integrating temporal or multispectral information to further enhance real-world plant disease monitoring.

{
    \small
    \bibliographystyle{elsarticle-num}
    \bibliography{reference}

@String(ICASSP=	{ICASSP})

@String(AAAI = {AAAI})

@inproceedings{caron2021emerging,
  title={Emerging properties in self-supervised vision transformers},
  author={Caron, Mathilde and Touvron, Hugo and Misra, Ishan and J{\'e}gou, Herv{\'e} and Mairal, Julien and Bojanowski, Piotr and Joulin, Armand},
  booktitle={Proceedings of the IEEE/CVF international conference on computer vision},
  pages={9650--9660},
  year={2021}
}

@article{hughes2015open,
  title={An open access repository of images on plant health to enable the development of mobile disease diagnostics},
  author={Hughes, David and Salath{\'e}, Marcel and others},
  journal={arXiv preprint arXiv:1511.08060},
  year={2015}
}

@incollection{singh2020plantdoc,
  title={PlantDoc: A dataset for visual plant disease detection},
  author={Singh, Davinder and Jain, Naman and Jain, Pranjali and Kayal, Pratik and Kumawat, Sudhakar and Batra, Nipun},
  booktitle={Proceedings of the 7th ACM IKDD CoDS and 25th COMAD},
  pages={249--253},
  year={2020}
}

@article{rauf2019citrus,
  title={A citrus fruits and leaves dataset for detection and classification of citrus diseases through machine learning},
  author={Rauf, Hafiz Tayyab and Saleem, Basharat Ali and Lali, M Ikram Ullah and Khan, Muhammad Attique and Sharif, Muhammad and Bukhari, Syed Ahmad Chan},
  journal={Data in brief},
  volume={26},
  pages={104340},
  year={2019},
  publisher={Elsevier}
}

@article{fang2021self,
  title={Self-supervised cross-iterative clustering for unlabeled plant disease images},
  author={Fang, Uno and Li, Jianxin and Lu, Xuequan and Gao, Longxiang and Ali, Mumtaz and Xiang, Yong},
  journal={Neurocomputing},
  volume={456},
  pages={36--48},
  year={2021},
  publisher={Elsevier}
}

@inproceedings{chen2021exploring,
  title={Exploring simple siamese representation learning},
  author={Chen, Xinlei and He, Kaiming},
  booktitle={Proceedings of the IEEE/CVF conference on computer vision and pattern recognition},
  pages={15750--15758},
  year={2021}
}

@article{grill2020bootstrap,
  title={Bootstrap your own latent-a new approach to self-supervised learning},
  author={Grill, Jean-Bastien and Strub, Florian and Altch{\'e}, Florent and Tallec, Corentin and Richemond, Pierre and Buchatskaya, Elena and Doersch, Carl and Avila Pires, Bernardo and Guo, Zhaohan and Gheshlaghi Azar, Mohammad and others},
  journal={Advances in neural information processing systems},
  volume={33},
  pages={21271--21284},
  year={2020}
}

@article{monowar2022self,
  title={Self-supervised clustering for leaf disease identification},
  author={Monowar, Muhammad Mostafa and Hamid, Md Abdul and Kateb, Faris A and Ohi, Abu Quwsar and Mridha, MF},
  journal={Agriculture},
  volume={12},
  number={6},
  pages={814},
  year={2022},
  publisher={MDPI}
}

@article{mamun2025conmamba,
  title={ConMamba: Contrastive Vision Mamba for Plant Disease Detection},
  author={Mamun, Abdullah Al and Zhang, Miaohua and Ahmedt-Aristizabal, David and Hayder, Zeeshan and Awrangjeb, Mohammad},
  journal={arXiv preprint arXiv:2506.03213},
  year={2025}
}

@article{Zhao2023CLA,
  title={{CLA:} A self-supervised contrastive learning method for leaf disease identification with domain adaptation},
  author={Zhao, Ruzhun and Zhu, Yuchang and Li, Yuanhong},
  journal={Computers and Electronics in Agriculture},
  volume={211},
  pages={107967},
  year={2023},
  publisher={Elsevier}
}

@article{bedi2021plant,
  title={Plant disease detection using hybrid model based on convolutional autoencoder and convolutional neural network},
  author={Bedi, Punam and Gole, Pushkar},
  journal={Artificial Intelligence in Agriculture},
  volume={5},
  pages={90--101},
  year={2021},
  publisher={Elsevier}
}

@inproceedings{chung2024addressing,
  title={Addressing Data Imbalance in Plant Disease Recognition through Contrastive Learning},
  author={Chung, Bryan},
  booktitle={2024 IEEE 3rd International Conference on AI in Cybersecurity},
  pages={1--6},
  year={2024},
  organization={IEEE}
}

@inproceedings{he2022masked,
  title={Masked autoencoders are scalable vision learners},
  author={He, Kaiming and Chen, Xinlei and Xie, Saining and Li, Yanghao and Doll{\'a}r, Piotr and Girshick, Ross},
  booktitle={Proceedings of the IEEE/CVF conference on computer vision and pattern recognition},
  pages={16000--16009},
  year={2022}
}

@article{dosovitskiy2020image,
  title={An image is worth 16x16 words: Transformers for image recognition at scale},
  author={Dosovitskiy, Alexey and Beyer, Lucas and Kolesnikov, Alexander and Weissenborn, Dirk and Zhai, Xiaohua and Unterthiner, Thomas and Dehghani, Mostafa and Minderer, Matthias and Heigold, Georg and Gelly, Sylvain and others},
  journal={arXiv preprint arXiv:2010.11929},
  year={2020}
}

@article{gu2023mamba,
  title={Mamba: Linear-time sequence modeling with selective state spaces},
  author={Gu, Albert and Dao, Tri},
  journal={arXiv preprint arXiv:2312.00752},
  year={2023}
}

@inproceedings{chen2020simple,
  title={A simple framework for contrastive learning of visual representations},
  author={Chen, Ting and Kornblith, Simon and Norouzi, Mohammad and Hinton, Geoffrey},
  booktitle={International conference on machine learning},
  pages={1597--1607},
  year={2020},
  organization={PmLR}
}

@inproceedings{he2020momentum,
  title={Momentum contrast for unsupervised visual representation learning},
  author={He, Kaiming and Fan, Haoqi and Wu, Yuxin and Xie, Saining and Girshick, Ross},
  booktitle={Proceedings of the IEEE/CVF conference on computer vision and pattern recognition},
  pages={9729--9738},
  year={2020}
}

@article{barbedo2019plant,
  title={Plant disease identification from individual lesions and spots using deep learning},
  author={Barbedo, Jayme Garcia Arnal},
  journal={Biosystems engineering},
  volume={180},
  pages={96--107},
  year={2019},
  publisher={Elsevier}
}

@article{kumar2023soybean,
  title={Soybean disease detection and segmentation based on Mask-RCNN algorithm},
  author={Kumar, M and Chandel, NS and Singh, D and Rajput, LS},
  journal={J Exp Agric Int},
  volume={45},
  number={5},
  pages={63--72},
  year={2023}
}

@article{abid2024bangladeshi,
  title={Bangladeshi crops leaf disease detection using YOLOv8},
  author={Abid, Md Shahriar Zaman and Jahan, Busrat and Al Mamun, Abdullah and Hossen, Md Jakir and Mazumder, Shazzad Hossain},
  journal={Heliyon},
  volume={10},
  number={18},
  year={2024},
  publisher={Elsevier}
}

@article{gidaris2018unsupervised,
  title={Unsupervised representation learning by predicting image rotations},
  author={Gidaris, Spyros and Singh, Praveer and Komodakis, Nikos},
  journal={arXiv preprint arXiv:1803.07728},
  year={2018}
}

@inproceedings{zhang2016colorful,
  title={Colorful image colorization},
  author={Zhang, Richard and Isola, Phillip and Efros, Alexei A},
  booktitle={European conference on computer vision},
  pages={649--666},
  year={2016},
  organization={Springer}
}

@article{bardes2021vicreg,
  title={Vicreg: Variance-invariance-covariance regularization for self-supervised learning},
  author={Bardes, Adrien and Ponce, Jean and LeCun, Yann},
  journal={arXiv preprint arXiv:2105.04906},
  year={2021}
}

@article{bao2021beit,
  title={Beit: Bert pre-training of image transformers},
  author={Bao, Hangbo and Dong, Li and Piao, Songhao and Wei, Furu},
  journal={arXiv preprint arXiv:2106.08254},
  year={2021}
}

@article{zhou2021ibot,
  title={ibot: Image bert pre-training with online tokenizer},
  author={Zhou, Jinghao and Wei, Chen and Wang, Huiyu and Shen, Wei and Xie, Cihang and Yuille, Alan and Kong, Tao},
  journal={arXiv preprint arXiv:2111.07832},
  year={2021}
}

@inproceedings{li2024hierarchical,
  title={Hierarchical multi-scale feature fusion network based for apple leaf disease recognition},
  author={Li, Daxiang and Zhang, Wenkai},
  booktitle={Proceedings of the 2024 7th International Conference on Artificial Intelligence and Pattern Recognition},
  pages={379--385},
  year={2024}
}

@inproceedings{fan2021multiscale,
  title={Multiscale vision transformers},
  author={Fan, Haoqi and Xiong, Bo and Mangalam, Karttikeya and Li, Yanghao and Yan, Zhicheng and Malik, Jitendra and Feichtenhofer, Christoph},
  booktitle={Proceedings of the IEEE/CVF international conference on computer vision},
  pages={6824--6835},
  year={2021}
}

@inproceedings{chen2021crossvit,
  title={Crossvit: Cross-attention multi-scale vision transformer for image classification},
  author={Chen, Chun-Fu Richard and Fan, Quanfu and Panda, Rameswar},
  booktitle={Proceedings of the IEEE/CVF international conference on computer vision},
  pages={357--366},
  year={2021}
}

@inproceedings{touvron2021training,
  title={Training data-efficient image transformers \& distillation through attention},
  author={Touvron, Hugo and Cord, Matthieu and Douze, Matthijs and Massa, Francisco and Sablayrolles, Alexandre and J{\'e}gou, Herv{\'e}},
  booktitle={International conference on machine learning},
  pages={10347--10357},
  year={2021},
  organization={PMLR}
}

@article{xu2021bag,
  title={Bag of instances aggregation boosts self-supervised distillation},
  author={Xu, Haohang and Fang, Jiemin and Zhang, Xiaopeng and Xie, Lingxi and Wang, Xinggang and Dai, Wenrui and Xiong, Hongkai and Tian, Qi},
  journal={arXiv preprint arXiv:2107.01691},
  year={2021}
}

@article{abbasi2020compress,
  title={Compress: Self-supervised learning by compressing representations},
  author={Abbasi Koohpayegani, Soroush and Tejankar, Ajinkya and Pirsiavash, Hamed},
  journal={Advances in neural information processing systems},
  volume={33},
  pages={12980--12992},
  year={2020}
}

@inproceedings{tejankar2021isd,
  title={Isd: Self-supervised learning by iterative similarity distillation},
  author={Tejankar, Ajinkya and Koohpayegani, Soroush Abbasi and Pillai, Vipin and Favaro, Paolo and Pirsiavash, Hamed},
  booktitle={Proceedings of the IEEE/CVF international conference on computer vision},
  pages={9609--9618},
  year={2021}
}

@article{navaneet2022simreg,
  title={Simreg: Regression as a simple yet effective tool for self-supervised knowledge distillation},
  author={Navaneet, KL and Koohpayegani, Soroush Abbasi and Tejankar, Ajinkya and Pirsiavash, Hamed},
  journal={arXiv preprint arXiv:2201.05131},
  year={2022}
}

@article{fang2021seed,
  title={Seed: Self-supervised distillation for visual representation},
  author={Fang, Zhiyuan and Wang, Jianfeng and Wang, Lijuan and Zhang, Lei and Yang, Yezhou and Liu, Zicheng},
  journal={arXiv preprint arXiv:2101.04731},
  year={2021}
}

@inproceedings{dadashzadeh2022auxiliary,
  title={Auxiliary learning for self-supervised video representation via similarity-based knowledge distillation},
  author={Dadashzadeh, Amirhossein and Whone, Alan and Mirmehdi, Majid},
  booktitle={Proceedings of the IEEE/CVF Conference on Computer Vision and Pattern Recognition},
  pages={4231--4240},
  year={2022}
}

@article{gu2021efficiently,
  title={Efficiently modeling long sequences with structured state spaces},
  author={Gu, Albert and Goel, Karan and R{\'e}, Christopher},
  journal={arXiv preprint arXiv:2111.00396},
  year={2021}
}

@article{gu2020hippo,
  title={Hippo: Recurrent memory with optimal polynomial projections},
  author={Gu, Albert and Dao, Tri and Ermon, Stefano and Rudra, Atri and R{\'e}, Christopher},
  journal={Advances in neural information processing systems},
  volume={33},
  pages={1474--1487},
  year={2020}
}

@inproceedings{masuyama2024mamba,
  title={Mamba-Based Decoder-Only Approach with Bidirectional Speech Modeling for Speech Recognition},
  author={Masuyama, Yoshiki and Miyazaki, Koichi and Murata, Masato},
  booktitle={2024 IEEE Spoken Language Technology Workshop},
  pages={1--6},
  year={2024},
  organization={IEEE}
}

@article{erol2024audio,
  title={Audio mamba: Bidirectional state space model for audio representation learning},
  author={Erol, Mehmet Hamza and Senocak, Arda and Feng, Jiu and Chung, Joon Son},
  journal={IEEE Signal Processing Letters},
  year={2024},
  publisher={IEEE}
}

@inproceedings{zhang2024motion,
  title={Motion mamba: Efficient and long sequence motion generation},
  author={Zhang, Zeyu and Liu, Akide and Reid, Ian and Hartley, Richard and Zhuang, Bohan and Tang, Hao},
  booktitle={European Conference on Computer Vision},
  pages={265--282},
  year={2024},
  organization={Springer}
}

@article{zhou2024bit,
  title={Bit-mamsleep: Bidirectional temporal mamba for eeg sleep staging},
  author={Zhou, Xinliang and Han, Yuzhe and Chen, Zhisheng and Liu, Chenyu and Ding, Yi and Jia, Ziyu and Liu, Yang},
  journal={arXiv preprint arXiv:2411.01589},
  year={2024}
}

@article{li2024harmamba,
  title={HARMamba: Efficient and lightweight wearable sensor human activity recognition based on bidirectional mamba},
  author={Li, Shuangjian and Zhu, Tao and Duan, Furong and Chen, Liming and Ning, Huansheng and Nugent, Christopher and Wan, Yaping},
  journal={IEEE Internet of Things Journal},
  year={2024},
  publisher={IEEE}
}

@article{liang2024bi,
  title={Bi-mamba+: Bidirectional mamba for time series forecasting},
  author={Liang, Aobo and Jiang, Xingguo and Sun, Yan and Shi, Xiaohou and Li, Ke},
  journal={arXiv preprint arXiv:2404.15772},
  year={2024}
}

@inproceedings{gao2025ssd,
  title={SSD-TS: Exploring the potential of linear state space models for diffusion models in time series imputation},
  author={Gao, Hongfan and Shen, Wangmeng and Qiu, Xiangfei and Xu, Ronghui and Yang, Bin and Hu, Jilin},
  booktitle={Proceedings of the 31st ACM SIGKDD Conference on Knowledge Discovery and Data Mining V. 2},
  pages={649--660},
  year={2025}
}

@inproceedings{singh2025best,
  title={BEST-STD: Bidirectional Mamba-Enhanced Speech Tokenization for Spoken Term Detection},
  author={Singh, Anup and Demuynck, Kris and Arora, Vipul},
  booktitle={ICASSP 2025-2025 IEEE International Conference on Acoustics, Speech and Signal Processing (ICASSP)},
  pages={1--5},
  year={2025},
  organization={IEEE}
}

@article{vaswani2017attention,
  title={Attention is all you need},
  author={Vaswani, Ashish and Shazeer, Noam and Parmar, Niki and Uszkoreit, Jakob and Jones, Llion and Gomez, Aidan N and Kaiser, {\L}ukasz and Polosukhin, Illia},
  journal={Advances in neural information processing systems},
  volume={30},
  year={2017}
}

@inproceedings{xie2022simmim,
  title={Simmim: A simple framework for masked image modeling},
  author={Xie, Zhenda and Zhang, Zheng and Cao, Yue and Lin, Yutong and Bao, Jianmin and Yao, Zhuliang and Dai, Qi and Hu, Han},
  booktitle={Proceedings of the IEEE/CVF conference on computer vision and pattern recognition},
  pages={9653--9663},
  year={2022}
}

@article{yu2021maskcov,
  title={Maskcov: A random mask covariance network for ultra-fine-grained visual categorization},
  author={Yu, Xiaohan and Zhao, Yang and Gao, Yongsheng and Xiong, Shengwu},
  journal={Pattern Recognition},
  volume={119},
  pages={108067},
  year={2021},
  publisher={Elsevier}
}

@inproceedings{he2022transfg,
  title={Transfg: A transformer architecture for fine-grained recognition},
  author={He, Ju and Chen, Jie-Neng and Liu, Shuai and Kortylewski, Adam and Yang, Cheng and Bai, Yutong and Wang, Changhu},
  booktitle={Proceedings of the AAAI conference on artificial intelligence},
  volume={36},
  pages={852--860},
  year={2022}
}

@article{yu2023mix,
  title={Mix-ViT: Mixing attentive vision transformer for ultra-fine-grained visual categorization},
  author={Yu, X. and Wang, J. and Zhao, Y. and Gao, Y.},
  journal={Pattern Recognition},
  volume={135},
  pages={109131},
  year={2023}
}

@article{zhang2023information,
  title={An information entropy masked vision transformer (IEM-ViT) model for recognition of tea diseases},
  author={Zhang, J. and Guo, H. and Guo, J. and Zhang, J.},
  journal={Agronomy},
  volume={13},
  number={4},
  pages={1156},
  year={2023}
}

@article{huan2025unified,
  title={A unified self-supervised framework for plant disease detection on laboratory and in-field images},
  author={Huan, Xiaoli and Chen, Bernard and Zhou, Hong},
  journal={Electronics},
  volume={14},
  number={17},
  pages={3410},
  year={2025},
  publisher={MDPI}
}

@article{caron2020unsupervised,
  title={Unsupervised learning of visual features by contrasting cluster assignments},
  author={Caron, Mathilde and Misra, Ishan and Mairal, Julien and Goyal, Priya and Bojanowski, Piotr and Joulin, Armand},
  journal={Advances in neural information processing systems},
  volume={33},
  pages={9912--9924},
  year={2020}
}

@article{zhu2024vision,
  title={Vision mamba: Efficient visual representation learning with bidirectional state space model},
  author={Zhu, Lianghui and Liao, Bencheng and Zhang, Qian and Wang, Xinlong and Liu, Wenyu and Wang, Xinggang},
  journal={arXiv preprint arXiv:2401.09417},
  year={2024}
}

@article{al2024plant,
  title={Plant disease detection using self-supervised learning: A systematic review},
  author={Al Mamun, Abdullah and Ahmedt-Aristizabal, David and Zhang, Miaohua and Hossen, Md Ismail and Hayder, Zeeshan and Awrangjeb, Mohammad},
  journal={IEEE Access},
  year={2024},
  publisher={IEEE}
}
}

\end{document}